\documentclass[10pt]{article}

\usepackage[margin=1in]{geometry}

\usepackage{amsmath,amssymb,amsthm}

\usepackage{newtxtext,newtxmath}   

\usepackage{natbib}
\usepackage{xcolor}
\usepackage{subcaption}

\usepackage{hyperref}

\usepackage{mathrsfs}
\usepackage{graphicx}
\usepackage{booktabs}
\usepackage{bbm}

\newtheorem{theorem}{Theorem}
\newtheorem*{theorem1}{Theorem 1}
\newtheorem*{theorem2}{Theorem 2}
\newtheorem*{theorem3}{Theorem 3}
\newtheorem*{theorem4}{Theorem 4}

\newtheorem*{lemma1}{Lemma 1}
\newtheorem*{corollary1}{Corollary 1}
\newtheorem{corollary}{Corollary}
\newtheorem{lemma}{Lemma}
\newtheorem{definition}{Definition}

\title{Score Approximation for Diffusion Models on \\ Arbitrary Low-Dimensional Structures}
\author{%
  \begin{minipage}{\textwidth}\centering
    \begin{minipage}[t]{0.24\textwidth}\centering\small
      Xinhe Mu \\
      Academy of Mathematics and \\
      Systems Sciences \\
      Chinese Academy of Sciences \\
      Beijing, China \\
      \texttt{muxinhe22@mails.ucas.ac.cn}
    \end{minipage}\hfill
    \begin{minipage}[t]{0.24\textwidth}\centering\small
      Zaijiu Shang \\
      Shanghai Institute for \\
      Mathematics and \\
      Interdisciplinary Sciences \\
      Shanghai, China \\
      \texttt{zaijiu@simis.cn;} \\
      \texttt{zaijiu@amss.ac.cn}
    \end{minipage}\hfill
    \begin{minipage}[t]{0.24\textwidth}\centering\small
      Zhaoqi Zhou \\
      Huawei Technologies Co., Ltd. \\
      Beijing, China \\
      \texttt{zhouzhaoqi1@huawei.com}
    \end{minipage}\hfill
    \begin{minipage}[t]{0.24\textwidth}\centering\small
      Chuan Zhou \\
      Academy of Mathematics and \\
      Systems Sciences \\
      Chinese Academy of Sciences \\
      Beijing, China \\
      \texttt{zhouchuan@amss.ac.cn}
    \end{minipage}
  \end{minipage}
  \\[3em]
  \begin{minipage}{\textwidth}\centering
    \begin{minipage}[t]{0.32\textwidth}\centering\small
      Qi Meng \\
      Academy of Mathematics and \\
      Systems Sciences \\
      Chinese Academy of Sciences \\
      Beijing, China \\
      \texttt{meq@amss.ac.cn}
    \end{minipage}\hfill
    \begin{minipage}[t]{0.32\textwidth}\centering\small
      Guiying Yan \\
      Academy of Mathematics and \\
      Systems Sciences \\
      Chinese Academy of Sciences \\
      Beijing, China \\
      \texttt{yangy@amt.ac.cn}
    \end{minipage}\hfill
    \begin{minipage}[t]{0.32\textwidth}\centering\small
      Zhiming Ma \\
      Academy of Mathematics and \\
      Systems Sciences \\
      Chinese Academy of Sciences \\
      Beijing, China \\
      \texttt{mazm@amt.ac.cn}
    \end{minipage}
  \end{minipage}
}
\date{{\today}}
\begin{document}
\maketitle
\begin{abstract}
Score-based diffusion models have achieved remarkable empirical success, motivating extensive theoretical work to establish their foundations. However, existing complexity bounds for score approximation, a vital step in diffusion modeling, rely on rigid constraints such as Lipschitz continuous scores or lower bounded densities. This severely limits their applicability to real-world perceptual data, where singularities, sharp boundaries, and disjoint clusters routinely violate such restrictive assumptions. We bridge this gap between theory and practice, presenting the first universal score approximation theorem applicable to any compactly supported distribution in $\mathbb{R}^n$. Using a novel discretization technique that directly models the underlying distribution, we prove that the neural network complexity is governed by the support’s upper Minkowski dimension $d$ rather than the ambient dimension $n$. Furthermore, by leveraging the inherent smoothing of Gaussian kernels, we show that even for irregular, fractal distributions, an $\epsilon$ approximation error can be achieved with  $\mathcal{O}(\epsilon^{-{d/(M+1)}})$ local Taylor modules each at size of $\mathcal{O}(n^M)$, an $\epsilon$-scaling rate previously achieved only for distributions with $(M+1)$-H\"older smoothness. Thus, we reveal a possible mechanism by which score-based diffusion models represent non-smooth data distributions.
\end{abstract}

\section{Introduction}
In machine learning, researchers are always enthusiastic to unravel the theory behind an architecture that achieves groundbreaking success. The community have witnessed this surge of theory papers \citep{resnet1,transformer1,transformer2} after the success of Resnet \citep{resnetorig} and transformers \citep{transformerorig}, and the field of score-based diffusion models is no exception: ever since its introduction \citep{DDPM,SGM,SGMSDE,Beatgan} around 2020, solid research has established error bounds for using diffusion models to reconstruct a distribution from finite samples \citep{Minimax,MinimaxWithLBAssuption,samplegeneralization,lowdimgeneralization}, discretizing the backward diffusion process using numerical SDE solvers \citep{exponentialcomplexityconvergence,Polyconverge,polyconvergeuserfriendly,backwardsdelowadapt}, and using ReLU networks to approximate the score function \citep{wangsubspace,wangmanifold}. 

A common assumption made by theoretical papers is that their target of approximation exhibits some degree of smoothness, and the field of diffusion models is also no exception: \citet{Sampaseasy,Polyconverge,wangsubspace,fastpfode} assume the underlying distribution $p_0$ to have Lipschitz score or on-support score (when $p_0$ is supported on a subspace of $\mathbb{R}^n$), while \citet{samplegeneralization,MinimaxWithLBAssuption,Minimax,wangmanifold}, respectively, require $p_0$ to be smooth in the $C^1$, Besov, Sobolev, or H\"older sense; \citet{aegradone,aegradtwo} relax the smoothness condition, but still require $p_0$ to be differentiable almost everywhere. Even the rare exceptions \citet{dlinear,nonasymptot,Polyconvergegeneral} assume the availability of an accurate score approximation network, whose existence is currently proven only for smooth distributions. However, these common assumptions can be limiting---in both practice and theory---for the reasons below.

Practically, real-world image distributions are characterized by singularities \citep{singularityone,singularitytwo} and sharp boundaries \citep{sharpboundary} near which the density abruptly changes. Indeed, the simple fact that RGB pixel intensities and often clip abruptly at 0 or 255 alone creates step-like discontinuities, breaking standard continuity assumptions. Theoretically, mitigating the problems posed by discontinuous densities and spiky gradients in the data distribution was a key reason why diffusion models introduce Gaussian noise---and in many different magnitudes \citep{SGM}. While small noise scales still result in score blowup around discontinuities \citep{SGM,wangsubspace}, such problems were taken into consideration when most works introduce early stopping \citep{nonasymptot,dlinear}, set error tolerance
\begin{align}
\label{MSE}
\mathrm{E}_{p_t}\left\|\nabla\log p_t(x)-\nabla\log p_{t,appr}(x)\right\|_2^2
\end{align}
to ${\epsilon^2}/{\sigma^4(t)}$ or ${\epsilon^2}/{\sigma^2(t)}$ (rather than a fixed $\epsilon^2$) at time $t$ \citep{wangsubspace,wangmanifold,Polyconvergegeneral}, and require convergence in Wasserstein rather than $\mathrm{TV}$ distance for distributions supported on manifolds \citep{Polyconvergegeneral,lowdimgeneralization,wangsubspace}. Thus, convergence guarantees under smoothness assumptions fail to explain how score-based diffusion models achieve one of their fundamental purposes.
\begin{figure*}[t]
  \centering
  \includegraphics[width=\linewidth]{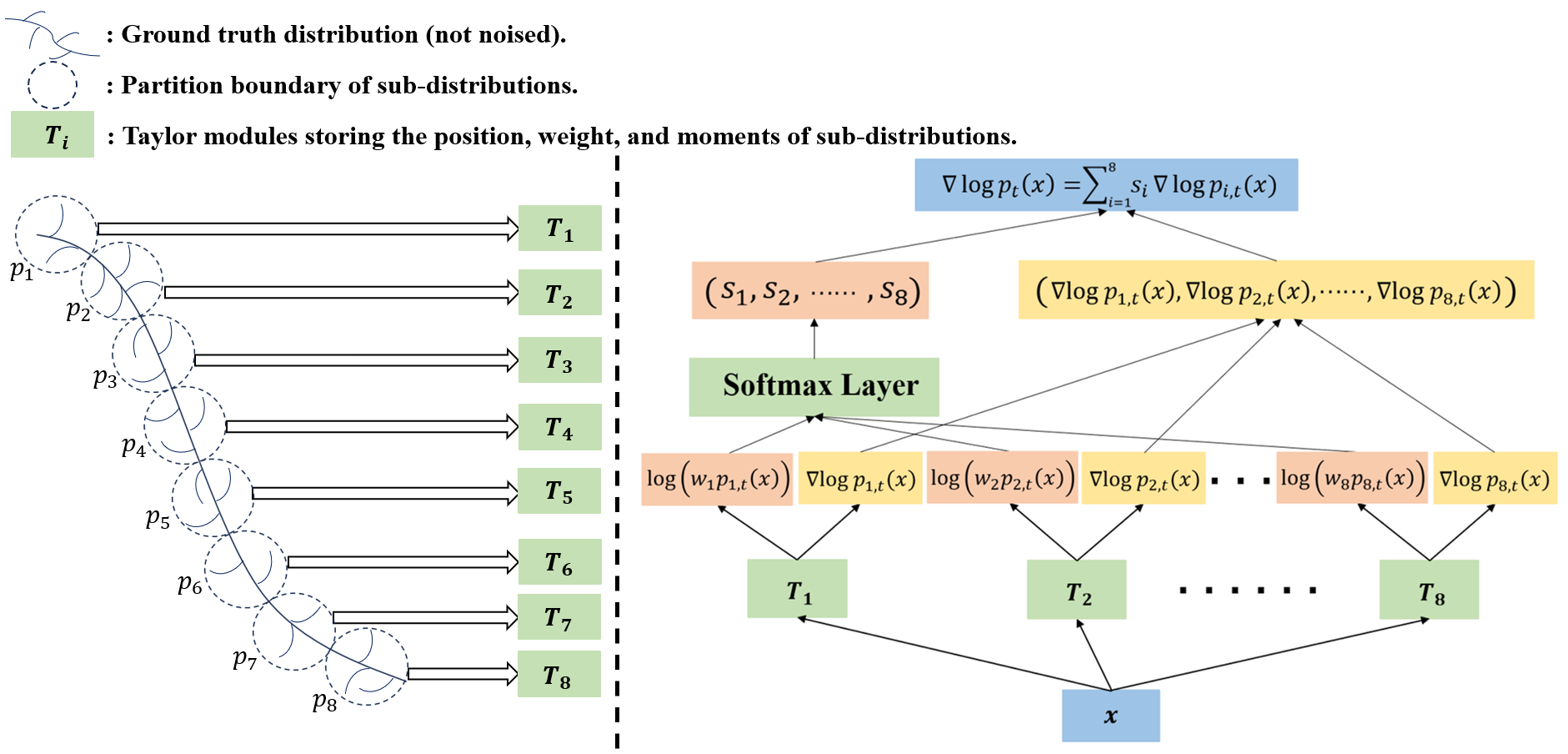}
  \caption{Overview of our discretization-based network architecture (illustrated for $P=8$ local covers). Theorem \ref{Thm:Discretizationerror} analyzes its accuracy under EA regularity, Theorem \ref{Thm:Universalregularity} proves EA regularity to be universally satisfied, and Theorem \ref{Thm:ReLU} realizes it using a ReLU neural network.}
  \label{ScoreApprox}
\end{figure*}

To address this gap, using the architecture outlined in Figure \ref{ScoreApprox} and proven effective by Theorems \ref{thm:main} to \ref{Thm:ReLU}, we establish a truly universal score approximation theory applicable to \textit{any} compactly supported distribution with no smoothness assumptions. Our main contributions are: 
\begin{itemize}
\item We construct a ReLU network that approximates the score of any compact distribution $p$ by decomposing it into sub-distributions $\{p_j\}$ of small diameter, using Taylor expansion to approximate each $p_j$'s convolution with a Gaussian kernel, then calculating $\nabla\log p_t(x)$ as the average of $\{\nabla\log p_{j,t}(x)\}$ weighed by a Softmax function. The size of our network is governed by the support's upper Minkowski dimensionality, and substantially smaller than those of existing works \citep{wangsubspace,wangmanifold} for nonsmooth (which renders their theory inapplicable) or moderately smooth data.

\item We define EA regularity (Definition \ref{def:reg}), a ``relaxed" Ahlfors regularity \citep{ahlfors} satisfied by any compact distribution. By ensuring that irregular density spikes are overpowered by the super-exponential decay of Gaussian kernels, EA regularity allows us to bound the error of our network and, as a bonus, bound $j$th order moments of $\nabla^k\log p_t(x)$ for all $j,k\in\mathbb{N^+}$, a fundamental problem in diffusion theory\citep{existingscorebound} potentially helpful for enhancing works including
\citet{Polyconvergegeneral,polyconvergeuserfriendly}.

\item We use $x$-prediction to approximate the score of two jagged, discontinuous fractal distributions with Minkowski dimension $\log 4/\log 5$, respectively embedded in a one-dimensional and a two-dimensional submanifold of $\mathbb{R}^{32}$. Despite differences in manifold dimensionality, approximation error for the two distributions exhibits similar scaling behavior with respect to network width $W$ and early stopping time $t_0$, verifying our prediction that network complexity is governed by the support's upper Minkowski dimension rather than manifold dimension.   
\end{itemize}
\section{Background and Related Work}
Since the introduction of Score-based Generative Modeling \citep{SGM} and DDPM \citep{DDPM}, pioneering works of Song, Sohl-Dickstein, and Kingma \citep{SGMSDE}, Song, Dhariwal, and Chen \citep{Consistency}, Dhariwal and Nichol \citep{Beatgan}, and many others, have established  the \textit{forward-backward process} as an empiracally verified framework for overcoming the challenges posed by sharp gradients, high-curvature manifold supports, and fragmented typical sets \citep{SGM} common in high-dimensional generative tasks.

Given a set of samples from an unknown distribution $p_0$ in $\mathbb{R}^n$, the \textit{forward process} corrupts $p_0$ with Gaussian noise by having samples undergo the \textit{forward Stochastic Differential Equation (SDE)}
\begin{align}
\label{forward}
dx=-\alpha(t)xdt+\beta(t)dw,t\in [0,T]
\end{align}
with $w$ the $n$-dimensional standard Brownian motion, and $T$ large enough such that the marginal distribution $p_T$ at time $T$ effectively becomes a known Gaussian distribution. Under such perturbation, a random sample  $x(t)$ can be expressed as the sum of $k^{-1}(t)x_{sig}$ and $\sigma(t)x_{noise}$, where
$$x_{sig}\sim p_0, x_{noise}\sim \mathcal{N}(0,I_n), k(t)=e^{\int_0^t\alpha(s)ds}, \sigma^2(0)=0, \mathrm{and}\ \frac{d\sigma^2}{dt}=\beta^2(t)-2\alpha(t)\sigma^2(t)$$
with initial condition $\sigma^2(0)=0$. Thus, defining  $(\tau_k\circ f)(x)=k^nf(kx)$ and denoting the probability density function of $\mathcal{N}(0,\sigma^2(t)I_n)$ as $\varphi_{\sigma^2(t)}$, the marginal distribution $p_t$ admits the expression 
\begin{align}
\label{ptexpression}
p_{t}=(\tau_{k(t)}\circ p_0)*\varphi_{\sigma^2(t)},
\end{align}
showing that all positive choices for $\alpha$ and $\beta$ create distributions equivalent up to a space-based rescaling and time-based reparameterization \citep{Polyconverge,Minimax}. As such, we assume the VE scheme ($\alpha(t)=0,\beta(t)=1,\sigma^2(t)=t,k(t)=1$) unless otherwise specified.

The \textit{backward process} removes noise from corrupted samples, solving the \textit{backward diffusion SDE}
$$d\tilde{x}=-[\alpha(t)x+\beta^2(t)\nabla \log p_t(x)]dt+\beta(t)d\hat{w};\ \ \tilde{x}(T)\sim\mathcal{N}(0,\sigma^2(T)I_n)\approx p_T$$
from time $T$ to time $t_0$, where $\hat{w}$ denotes a standard Wiener process in the reverse-time direction. Here, information lost in the noise injection phase is recovered using the \textit{score function} $\nabla\log p_t(x)$, and the early stopping time $t_0$ is introduced because the score is widely observed to blow up as $t\to 0$, eventually becoming impossible to approximate \citep{SGM,MinimaxWithLBAssuption}.  

As the score function $\nabla\log p_t(x)$ satisfies the property
$$\sigma^2(t)\nabla\log p_t(x)=k^{-1}(t)(\mathrm{E}_{p(x_{sig}|x,t)}{x_{sig})-x},$$
existing models often calculate the score function using the $x$ (data)-prediction model. That is, training a neural network $s_\theta(x,t)$ to approximate $\mathrm{E}_{p(x_{sig}|x,t)}x_{sig}$, and estimating
\begin{align}
\label{dataperdiction}
\nabla\log p_{t,appr}(x)=\sigma^{-2}(t)[k^{-1}(t){s_\theta(x,t)}-{x}],
\end{align}
a technique shown to better capture low-dimensional structures of the underlying distribution \citep{letdenoise} and is thus adopted in our analysis.
\subsection{Prior Theoretical Guarantees}
Existing theoretical analysis for diffusion models mainly focus on one of the three topics below:
\begin{enumerate}
\item Given sufficient resource, how well can a score-based diffusion model reconstruct the underlying distribution using a finite number of data points?
\item Given an accurate enough score estimator,  how many numerical steps do score-based diffusion models need in order to converge to the underlying distribution?
\item Given constraints on network size and depth, how well can a neural network approximate the score function $\nabla\log p_t(x)$?
\end{enumerate}
Question 1 is the main focus of \citet{samplegeneralization,improvedcomplexitybound,MinimaxWithLBAssuption}, with highly interesting results derived by \citet{Minimax}, which proves that score-based diffusion models are minimax optimal at reconstructing distributions with sub-Gaussian tails  inside the $\beta$-Sobolev ($\beta \leq 2$) space, and \citet{lowdimgeneralization}, which establishes, in Wasserstein-p distance, error bounds for generalizing from $n$ samples, applicable to any distribution in $\mathbb{R}^n$ with finite $q$-th moment and $(p,q)-$Wasserstein dimension \footnote{See \citet{lowdimgeneralization} for the exact definition of $(p,q)$-Wasserstein dimension} $d$. 

Question 2 is the main focus of \citet{Polyconverge,Polyconvergegeneral,polyconvergeuserfriendly,exponentialcomplexityconvergence,nonasymptot,dlinear} and notably, \citet{backwardsdelowadapt}, which proves that for any underlying distribution $p_0$ in $\mathbb{R}^n$ supported on a set of upper Minkowski dimension $d$, an ideal choice of DDPM \citep{DDPM} coefficients can achieve, in $K$ denoising steps, a TV distance of $\mathcal{O}((d+\log n)^2/\sqrt{K})$ between the real $p_{t_0}$ and reconstructed $p_{t_0,s_\theta}$ given an accurate score estimator $s_\theta$. That is, we can accurately reconstruct any compact distribution in a polynomial number of steps simply by accurately approximating its score function $\nabla\log p_t(x)$.

In comparison, question 3 has received notably less attention and yielded limited results. Traditional results for approximating general functions on compact sets \citep{exponentialrelu} also apply to the score function. However, those bounds on the network size almost always grow exponentially with the ambient dimensionality, becoming impractical for high-dimensional image distributions satisfying the manifold hypothesis. \citet{wangsubspace} presents a score matching network with size independent of the ambient dimensionality, but their assumption--that the data distribution is supported on a $d$-dimensional linear subspace of $\mathbb{R}^n$--is a significant simplification of the manifold hypothesis seldom satisfied for real-world videos or images.

To the best of our knowledge, the most meaningful advancement in this direction is a concurrent work \citet{wangmanifold}, which expands the results of \citet{wangsubspace} to $C^\beta$ smooth, lower and upper bounded distributions concentrated on compact $d-$dimensional manifolds $\mathcal{M}_d\subseteq\mathbb{R}^n$ with positive reach $1/k_1$. Using geometric techniques, \citet{wangmanifold} proves that such distributions can be approximated with mean square error $\mathcal{O}(n^{2\gamma+d+2}\delta^2/t)$ using a neural network of size $\mathcal{O}(1/\delta^{{d}/{\beta}})$, which, to achieve $L^2$ error tolerance $\epsilon$, translates to a network size of order\footnote{For $k_1\leq\epsilon$, $\gamma$ defined by \citet{wangmanifold} satisfies $\gamma=1$ and hence $2\gamma+2=4$.}
$$\mathcal{O}\big(n^{\alpha}(\epsilon)^{-\theta}(t_0)^{-\omega}\big), \alpha={\frac{d^2+4d}{2\beta}},\theta=\frac{d}{\beta},\omega=\frac{d}{2\beta}$$
for manifolds of large reach. Aside from the fact that $n^{{d^2}/{2\beta}}$ may grow super-exponentially for moderately smooth ($\beta\leq2$) distributions, this presents an impressive result for approximating the score function of smooth data distributions lying on smooth, low-dimensional manifolds.

In contrast, we take an analytics-heavy approach utilizing the smoothing effect of Gaussian kernels. For a distribution with compact support $K\subseteq \mathbb{R}^n$ of upper Minkowski dimension $d$, we can derive, for all $M\leq \sqrt{n}$, an $M$th order complexity bound (omitting Polylog factors for simplicity) of order
\begin{align}
\label{complexityboundsimp}
\tilde{\mathcal{O}}\big(n^{\tilde\alpha(M)}(\epsilon)^{-\tilde{\theta}(M)}(t_0)^{-\tilde{\omega}(M)}\big), \tilde{\alpha}=M+\frac{(M+2)d}{2(M+1)},\tilde{\theta}=\frac{d}{(M+1)},\tilde{\omega}=\frac{(M+2)d}{2(M+1)}
\end{align}
which, when $K$ lies in a $d'$ dimensional manifold $\mathcal{M}$ with large reach, can be tightened to order
$$\mathcal{O}(n^M(d')^{\tilde{\omega}(M)}\epsilon^{-\tilde{\theta}(M)}t_0^{-\tilde{\omega}(M)}).$$
removing key constraints of \citet{wangmanifold} and reducing the exponent in $n$ (for all $\beta\leq2;d\geq2$) and $1/\epsilon$ (when choosing $M>\beta-1$) at a possibly inevitable cost (to be discussed in Section 5) of raising the exponent in $1/t_0$.
\begin{table}[h]
    \centering
    \caption{Comparison with \citet{wangmanifold}}
    \label{comparison}
    \begin{tabular}{c|c|c}
        \toprule
    &Our Work&\citet{wangmanifold}\\
        \midrule
    Exponent in $1/t_0$ & $\frac{M+2}{2(M+1)}d$ & $d/2\beta$\\
    Exponent in $1/\epsilon$ & $d/(M+1)$ & $d/\beta$ \\
    Exponent in $n$ & $M+\frac{M+2}{2(M+1)}d$ \textbf{(linear in $d$)} & $({d^2}+4d)/{2\beta}$ (quadratic in $d$)\\
    Requires $p_0$ & \textbf{No mandatory requirements} & $\beta$-H\"older, lower bounded\\
    Requires $\mathrm{supp}(p_0)$& \textbf{No mandatory requirements}& $C^2$, positive reach manifold\\
    Size of $M$ and $\beta$ & \textbf{Any} $M\leq\sqrt{n}$ ($n^M$ scaling tradeoff) & Usually $\beta\leq2$ (Lipschitz score)    \\
    Removing $n^d$ term & \textbf{Require $C^2$ positive reach manifold}& Not currently achievable\\
    \bottomrule
    \end{tabular}
\end{table}
\section{Main Results}
In this section, we present our four main theorems as well as the intuition behind their proof and formulation. Rigorous theorem proofs, as well as more detailed theorem formulations explicitly recording Polylog factors and constants omitted by big O-notation, will be provided in the appendix. 

We first provide our central theorem, establishing complexity bounds for approximating the score of arbitrary compact distributions.
\begin{theorem}
\label{thm:main}
Let $p_0$ be a probability distribution in $\mathbb{R}^n$, $n>25$, such that $\mathrm{supp}({p_0})$ is compact with diameter $R>3$, and that for all $r<1$, $\mathrm{supp}({p_0})$ can be covered by $C_Pr^{-d}$ balls of radius $r$. Under the VE perturbation scheme ($k(t)=1; \sigma^2(t)=t$), for target error $\epsilon<1/4$, early stopping time $t_0<1/4$ (thresholds for $n, R, t_0, \epsilon$ only simplify constants, see the remark in Appendix A), and order $1\leq M\leq\sqrt{n}$, set the concentration parameter $\lambda$ and radius parameter $r$ to be
$$\lambda\sim\mathcal{O}\big(\sqrt{n+\log ({C_PR}/{\epsilon t_0})+d\log{1}/{t_0}}\big), r\sim\mathcal{O}\left((\sqrt{t_0}/\lambda)^{\frac{M+2}{M+1}}\epsilon^{\frac{1}{M+1}}\right),$$
then the true score function $\nabla\log p_t(x)$ can then be approximated via the data-prediction formulation (\ref{dataperdiction})
with mean squared error (\ref{MSE}) bounded by $6\epsilon^2$ for all $t\in[t_0,T]$, using a fully connected ReLU network $s_\theta(x,t)$ with depth $D$ and width $W$ of order (omitting Polylog factors)
$$D\sim\tilde{\mathcal{O}}\big((M^2+d)\big);W\sim \tilde{\mathcal{O}}\big(C_Pr^{-d}(n^M+d^2)\big).$$ 

If, in addition, $\mathrm{supp}(p_0)$ is concentrated on a $C^2$, $d'$-dimensional manifold $\mathcal{M}$ ($d'\geq d$) with positive reach $1/k_1$, then for $t_0\leq\mathcal{O}(\frac{d'}{n^2k_1^2})$, $\lambda$ can be further shrunk to $\lambda_{\mathcal{M}}$ of magnitude
$$\lambda_{\mathcal{M}}\sim\mathcal{O}\big(\sqrt{d'+\log ({C_PC_\mathcal{M}R}/{\epsilon t_0})+d\log{1}/{t_0}}\big),$$
while $r$ can also be reduced accordingly, after relaxing the bound on (\ref{MSE}) to $7\epsilon^2$. 

Here, the constant $C_\mathcal{M}$ ensures that for all $r>0$, $\mathcal{M}$ admits a covering $\{B(x_i,r)\}_{i=1}^N$ such that $\forall y_i\in\mathcal{M}$, $B(y_i,\rho)$ contains the centers of at most $C_\mathcal{M}(1+\rho/r)^{d'}$ balls in $\{B(x_i,r)\}_{i=1}^N$. Such a constant exists (see Appendix F for proof) for all compact $C^2$ manifolds with positive reach.
\end{theorem}

The proof of Theorem \ref{thm:main} is organized as follows: Theorem \ref{Thm:Discretizationerror} analyzes the approximation algorithm in Figure \ref{ScoreApprox}, proving it to be accurate for any EA regular point. Theorem \ref{Thm:Universalregularity} establishes EA regularity as a property of all compact distributions, proving Theorem \ref{Thm:Discretizationerror}'s bound as universally applicable. Finally, Theorem \ref{Thm:ReLU} explicitly realizes Figure \ref{ScoreApprox}'s algorithm with a ReLU network. 

\subsection{Theorem 2: Analysis of Discretization Formula}
Our key insight is that corruption by Gaussian noise makes any distribution $C^\infty$ smooth, and thus can be handled by Taylor expansion on a sufficiently local scale. For example, consider a distribution $p_i$, $\mathrm{supp}(p_i)\subseteq B(y_i,r)$. For $r<<\sqrt{t}$, $p_{i,t}(x)$ as defined of (\ref{ptexpression}) admits the Taylor expansion\footnote{See proof of Theorem \ref{Thm:Discretizationerror} in the appendix for detailed derivation}
$$
e^{-\frac{\left\|x-y_i\right\|^2}{2t}}(1+{R}_{1,i}+\Delta_{M+1})=e^{-\frac{\left\|x-y_i\right\|^2}{2t}}\Big[1+\sum_{j=1}^M \frac{1}{j!t^{(j/2)}} H_j(\frac{x-y_i}{\sqrt{t}}):\int_{B(y_i,r)}(y-y_i)^{\otimes j}dp_i(y)+\Delta_{M+1}\Big]
$$
where $H_j=(-1)^je^{\frac{\|x\|^2}{2}}\nabla_j(e^{-\frac{\left\|x\right\|^2}{2}})$ is the Hermite tensor polynomial, ``$:$'' denotes tensor contraction, and the remainder $\Delta_{M+1}$ is given by
\begin{align}
\label{residue}
\Delta_{M+1}(x,t)=\frac{1}{(M+1)!}\int_{B(y_i,r)}e^{\frac{\langle 2x-y'-y_i,y'-y_i\rangle}{2t}}H_{M+1}(\frac{x-y'}{\sqrt{t}}):(\frac{y-y_i}{\sqrt{t}})^{\otimes M+1}dp_i(y).
\end{align}
The size of $\Delta_{M+1}$ can be estimated by the formula \citep{hermite}
\begin{align}
\label{hermite}
H_j(u):v^{\otimes j}=He_j(\langle u,v/\|v\|\rangle)\cdot\|v\|^j;He_j(r)=(-1)^je^{\frac{r^2}{2}}\frac{d^j}{dr^j}(e^{-\frac{r^2}{2}})
\end{align}
which implies that when $p_i$ lies on a smooth manifold $\mathcal{M}$, the component of $x-y_i$ normal to $T_{y_i}\mathcal{M}$ can grow much larger than the tangent component without notably enlarging $\Delta_{M+1}$. Similarly, 
$$t\nabla p_{i,t}(x)+(x-y_i)p_{i,t}(x)=p_i*[(y-y_i)\varphi_{t}]$$
admits the Taylor expansion\footnote{This expansion is deliberately one order lower. We allow ourselves a slight abuse of notation to define the contraction operation $:$ between a $j$-covariant tensor $P$ and a $(j+1)$-contravariant tensor $Q$ as producing a $1-$contravariant tensor $v$ such that $f(v)=(P\otimes f):Q$ for any 1-covariant tensor $f$.} $e^{-\frac{\|x-y_i\|^2}{2t}}(R_{2,i}+\tilde{\Delta}_M)$, where
$$R_{2,i}=\int_{B(y_i,r)}(y-y_i)dp_i(y)+\sum_{j=1}^{M-1}\frac{1}{j!}\cdot t^{-\frac{j}{2}}\cdot H_j(\frac{x-y_i}{\sqrt{t}}):\Big(\int_{B(y_i,r)}({y-y_i})^{\otimes j+1}dp_i(y)\Big)$$
while $\tilde{\Delta}_M$ shares similar properties with $\Delta_{M+1}$. As $j$th order moments of $(y-y_i)$ can be stored as network weights, while polynomials can be approximated by ReLU networks using the method of \citet{squarerelu}, this provides a blueprint for approximating $p_{i,t}(x)$ and $\nabla\log p_{i,t}(x)$.

Furthermore, for any decomposition $p_0=\sum_{i=1}^Pw_ip_i(x)$, $\nabla\log p_t(x)$ can be proven to equal 
\begin{align}
\label{scoredecomposition}
\mathrm{Softmax}\big(\log \big(w_1 p_{1,t}(x)\big),...,\log \big(w_Pp_{P,t}(x)\big)\big)\cdot(\nabla\log p_{1,t},\nabla\log p_{2,t},...,\nabla\log p_{P,t}),
\end{align}
which is exactly the architecture of Figure \ref{ScoreApprox}: cover $\mathrm{supp}(p_0)$ by balls $\{B(y_i,r)\}$, compute the log-likelihood and score of each $p_i$ with a Taylor module $T_i$, and average with Softmax weights. The danger is that the relative remainder ${\Delta_{M+1}}/(1+{R_{1,i}+\Delta_{M+1}})$ risks growing with $\|x-y_i\|$ at rate $\exp({\frac{\|x-y_i\|r}{t}})$, making distant kernels with large mass contaminate Softmax weights. The regularity notion below rules this out.
\begin{definition}
\label{def:reg}
Given a measure $\mu$ on $\mathbb{R}^n$, a point $x$ is EA (Exponentially Ahlfors) regular with constants $C_1,C_2$ at scale $r$, or $(C_1, C_2, r)$-EA regular, if there exists a center $z_0$, $x\in B(z_0,r)$, and
\begin{align}
\label{regularity}
C_1\cdot e^{C_2\frac{\left\|z_0-y\right\|}{r}}\mu(B(z_0,r))>\mu( B(y,r)), \quad \forall y \in \mathbb{R}^n.
\end{align}
\end{definition}
The concept of EA regularity ensures that super-exponential decay of Gaussian kernels overwhelms mass fluctuations of $p_0$, such that mass contributed by kernels near $x$ dominate the Softmax weights of (\ref{scoredecomposition}). With such, we derive the following theorem for the discretization error.
\begin{theorem}
\label{Thm:Discretizationerror}
Let $p_0$ be a compact distribution supported on a $C^2$, $d'$-dimensional manifold $\mathcal{M}\subseteq \mathbb{R}^n$ with positive reach ${1}/{k_1}$ (for truly arbitrary distributions, $\mathbb{R}^n$ can be viewed as its own submanifold with infinite reach), such that for all $r<1$, $\mathrm{supp}(p_0)$ can be covered by $C_Pr^{-d}$ balls of radius $r$ ($d\leq d'$). Define the function $\mathrm{Clamp_B}$, acting pointwise on vectors, as
$$\mathrm{Clamp}_B(x)=\max\{-B,\min\{B,x\}\}.$$
Under the VE scheme and the assumption that $\epsilon<1/4, t< 1/k_1^2,$ for $ M\geq 1$, take $\lambda(\epsilon),r$ of order
$$\lambda(\epsilon)\sim \mathcal{O}\big(\sqrt{C_2+\log ({C_1C_P}/{\epsilon})+d\log{1}/{t_c}}\big),r\sim\mathcal{O}((\sqrt{t_c}/\lambda)^{\frac{M+2}{M+1}}\epsilon^{\frac{1}{M+1}})$$
with $t_c=\min\{t,1/4\}$, and take $K$ such that $k_1K\lambda(\epsilon)\sqrt{t}\leq1/6$. Then, for any covering $\{B(y_i,r)\}_{i=1}^P$ of $\mathrm{supp}(p_0)$ such that all $y_i$ lie in $\mathcal{M}$, and for any decomposition $p_0=\sum_{i=1}^Pw_ip_i$ such that $\mathrm{supp}(p_i)\subseteq B(y_i,r)$, the discrete approximation
\begin{align}
\label{discreteapprox}
\nabla\log p_t(x)\approx\mathrm{Softmax}\big(\log \big(w_1\tilde{p}_{1,t}(x)\big),...,\log \big(w_P\tilde{p}_{P,t}(x)\big)\big)\cdot (\frac{\tilde{q}_{1,t}(x)}{t},...,\frac{\tilde{q}_{P,t}(x)}{t})-\frac{x}{t}
\end{align}
with 
$$\tilde{p}_{i,t}(x)=e^{-\frac{\left\|x-y_i\right\|^2}{2t}}\cdot \big(1+\mathrm{Clamp_{\frac{1}{3}}({R}_{1,i}})\big);\tilde{q}_{i,t}(x)=y_i+\frac{\mathrm{Clamp}_{2r}({R}_{2,i})}{1+\mathrm{Clamp_{\frac{1}{3}}({R}_{1,i}})}$$
has error bounded by $\epsilon$, if $x\in B(z, K\lambda(\epsilon)\sqrt{t}-r)$
holds for some $(C_1, C_2, \lambda(\epsilon)\sqrt{t})$-EA regular point $z$, and the projection of $x-z$ onto $z$'s tangent space $T_z\mathcal{M}$ has norm at most $\lambda(\epsilon)\sqrt{t}-r$.
\end{theorem}

\subsection{Theorem 3: Universal EA regularity}

To make the local error bound of Theorem \ref{Thm:Discretizationerror} hold with near certainty, we prove EA regularity to be a universal property of compact distributions $p_0\in\mathbb{R}^n$, with $C_1, C_2$ independent of $r$ or $p_0$.

\begin{theorem}
\label{Thm:Universalregularity}
(1) Let $\mu$ be an arbitrary probability distribution in $\mathbb{R}^n$ with compact support. $\forall$ radius $r$ fixed, define $U_\delta$ to be the set of points that are $(e^{3n}\frac{1}{\delta},\frac{n}{2},r)$-EA regular under $\mu$, then
$\mu(U_\delta)>1-\delta.$

(2) Let $\mu$ be a distribution concentrated on a $C^2$, $d'$-dimensional manifold $\mathcal{M}\subseteq \mathbb{R}^n$. Assume that for all $r>0$, $\mathcal{M}$ admits a covering $\{B(x_i,r)\}_{i=1}^N$ such that $\forall y_i\in\mathcal{M}$, $B(y_i,\rho)$ contains the centers of at most $C_\mathcal{M}(1+\rho/r)^{d'}$ balls inside $\{B(x_i,r)\}_{i=1}^N$. $\forall$ radius $r$ fixed, let $U_\delta$ be the set of points that are $ (C_{\mathcal{M}}e^{2.5d'}\frac{6}{\delta},\frac{1}{2}d',r)$-EA regular under $\mu$, then $\mu(U_\delta)>1-\delta$.
\end{theorem}
For proof, we define a ball $B(x,r)$ as "bad" if there exists a witness $B(y,r)$ violating (\ref{regularity}). That is, a point is $(C_1,C_2,r)$-EA regular if it is contained in any good ball, so for any covering $\{B(z_i,r)\}$ of $\mathrm{supp}(\mu)$, the total measure of bad balls control the total measure of irregular points.

The proof of Theorem \ref{Thm:Universalregularity} is relatively straightforward if all "bad" balls can be made to share the same witness $B(y_0,r)$. For example, the $\mathbb{R}^n$ case is proven by defining the covering $\{B(z_i,r)\}_{i=1}^\infty$ as 
$$\{z_i|i\in\mathbb{Z}\}=\{({2j_1r},...,{2j_nr})\cdot n^{-1/2}|j_1,...,j_n\in\mathbb{Z}\},$$ and utilizing the fact that $\int_{\mathbb{R}^n}e^{-\|x\|}dx$ converges for all $n\in\mathbb{N}$. This proof is also easily generalizable to the case where "bad" balls are mapped to witnesses balls that do not overlap. 

With this in mind, the general case is reduced to the non-overlapping witness case via a Vitali-style argument. Given a function $F:\{z_i|i\in\mathbb{Z}\}\to \mathcal{J}$ mapping each bad ball to its witness, we modify $F$ by iteratively selecting a witness $y_k$ such that 
$$\mu(B(y_k,r))\geq (1-\delta)\sup\{\mu(B(y_j,r))|y_j\in\mathcal{J}\},$$
finding all points such that $\left\|F(z_i)-y_k\right\|\leq 2r$, changing their image to $y_k$, and repeating the process. As $\mathrm{supp}(\mu)$ is compact, this process terminates after a finite iterations to a function $\tilde{F}$ such that $\mathrm{Im}(\tilde{F})$ becomes a discrete set of points with minimum pairwise distance $2r$, and that
\begin{align}
\label{minortermshift}
\|\tilde{F}(z_i)-F(z_i)\|\leq2r, \mu(B(\tilde{F}(z_i),r))\geq (1-\delta)\mu (B(F(z_i),r)), \forall z_i,
\end{align}
reducing $\tilde{F}$ to the non-overlapping witness case after enlarging $C_1$ to account for (\ref{minortermshift})'s influence.
\subsubsection{Extension: Bounding  $\nabla^2\log p_t(x)$ with EA regularity}
EA regularity also proves useful for another fundamental problem in diffusion theory: deriving bounds for the Hessian $\nabla^2\log p_t(x)$, on which the result of \citet{Sampaseasy}, \citet{Polyconverge}, and many other works depend.

This problem has been tackled by \citet{existingscorebound}, however, their bound $L(x,t)$ (see Lemma 3.3) on $\nabla^2\log p_t(x)$ contains a ${1}/{(p_t(x))^2}$ factor, making $\mathrm{E}_{p_t(x)}L(x,t)$ become infinity and forcing \citet{existingscorebound} to truncate their domain of interest to $\{x|p_t(x)\geq\gamma\}$. In contrast, EA regularity allows us to establish  $j$th order moment bounds on $t\nabla^2\log p_t(x)$ for all $1\leq j<\infty.$

\begin{lemma}
Let $h(x,t)$ be a function\footnote{possibly vector or tensor-valued} such that $\|h(x,t)\|\leq (\max\{1,{\|x\|}/{\sqrt{t}}\})^M$ for some $M\in \mathbb{N}$, and define the operator $H_h$ operating on probability measures $\mu$, such that
$$(H_h\circ \mu)(x,t)=\frac{\int_{\mathbb{R}^n}e^{-\frac{\left\|x-y\right\|^2}{2t}}h(y-x,t)d\mu}{\int_{\mathbb{R}^n}e^{-\frac{\left\|x-y\right\|^2}{2t}}d\mu}$$
$\forall x \in B(z,\lambda\sqrt{t})$ for some $(e^{3n}\frac{1}{\delta}, \frac{n}{2}, \lambda\sqrt{t})$-EA regular point $z$, $\lambda\geq \max\{\sqrt{n},M,\sqrt{\log ({1}/{\delta})}\}$, 
$$\|(H_h\circ\mu)(x,t)\|\leq(15\lambda)^M+1.$$
\end{lemma}
\begin{corollary}
Let $\{f_i\}_{i=1}^S$ satisfy $\|f_i(x,t)\|\leq C_i\max\{1,\|x\|/\sqrt{t}\}^{M_i}$, $\forall i\leq S$, and multiplication in $\mathrm{Im}(f_i)$ satisfy $\|f(x)\cdot g(y)\|\leq\|f(x)\|\|g(y)\|$. Under the VE scheme, for any compact distribution $p_0$ with probability measure $\mu$,  $\mathrm{E}_{p_t(x)}\|\prod_{i=1}^S(H_{f_i}\circ\mu)(x,t)\|^j$ is bounded by a constant depending on $j, n,$ and $\{f_i\}$, but independent of $p_0$ or $t$. 

Most notably, $\{t^{M/2}\big(\nabla^M\log p_t(x)\big)|M\in\mathbb{N}^+\}$ lies in the vector space spanned by such functions. For example, defining $\mu$ to be the probability measure associated with $p_0$, 
$$\sqrt{t}\nabla\log p_t(x)=H_\varphi\circ \mu,t\nabla^2\log p_t(x)=H_\psi\circ \mu-(H_\varphi\circ \mu)^{\otimes 2}; \varphi(x,t)=\frac{x}{\sqrt{t}},\psi(x,t)=\frac{x^Tx}{t}-I_n.$$
\end{corollary}
While an $L^\infty$ bound remains infeasible, our results guarantee that $\mu(\{x|\|t\nabla^2\log p_t(x)\|\geq T\}$ decays exponentially with $T$. As methods to handle problematic sets of vanishing measure have been extensively developed by \citet{Polyconvergegeneral}, EA regularity thus provides a promising foundation for future works analyzing the backward SDE to widen their scope of applicability.
\subsection{Theorem 4: Neural Network Approximation}
Finally, we prove that the construction in (\ref{discreteapprox}) can be efficiently realized using a  ReLU network.
\begin{theorem}
\label{Thm:ReLU}
Set $V_J=\mathrm{supp}(p_0)+B(0,J\sqrt{T})$. Assuming that $M\leq \min\{\sqrt{n},{1}/{r}\},$ $r\leq\min\{\sqrt{t_0},1/25\}$, the expression (\ref{discreteapprox}) can be realized  with error $\epsilon$ for all $(t,x)\in[t_0,T]\times V_J$ using the data prediction formulation (\ref{dataperdiction}), with $s_\theta(x,t)$ a ReLU network such that $\mathrm{Im}(s_\theta)\subseteq\mathrm{Conv(\mathrm{supp}(p_0))}+B(0,t_0\epsilon+3\sqrt{n}r).$
The depth $D$ and width $W$ of $s_\theta(x,t)$ are of order (omitting Polylog factors)
$$D\sim \tilde{\mathcal{O}}\big((M^2+d)\big); W\sim \tilde{\mathcal{O}}\big(C_Pr^{-d}(n^M+d^2)\big).$$
\end{theorem}
The proof boils down to two things: approximating $R_{1,i}$ and $R_{2,i}$, for $i\leq P$ and inputs $x\in V_J$; and $\mathrm{Softmax}$, over an unbounded domain. For the first, if the exact values of $\int_{B(y_i,r)}({y-y_i})^{\otimes j}dp_i(y)$ are baked into linear layers, we only need to calculate, for all $j\leq M$, the value of all elements in
$$Q_{n,j}:=\Big\{\prod_{k=1}^jP_{\alpha_k}(\frac{x-y_i}{\sqrt{t}})|\alpha_k \in \{1,...,n\}\Big\}=\Big\{P_{j}(\frac{x-y_i}{\sqrt{t}})\cdot v|j\in\{1,...,n\};v\in Q_{n,j-1}\Big\},$$
which can be done inductively, using \citet{squarerelu}'s method to approximate multiplication.

For the second, noting that $\max\{x_1,...,x_n\}$ can be exactly produced by repeatedly calculating $\max\{x_1,x_2\}=x_1+\mathrm{ReLU}(x_2-x_1),$ we can deduct $z_{max}$ from all inputs, bounding the domain to $(-\infty,0]$ while ensuring that at least one input equals $0$. Thus, the problem of calculating $\mathrm{Softmax}$ with error $\delta$ is reduced to calculating $\exp(x)$ $P$ times on $[-\log (P/\delta),0]$ (smaller inputs become negligible), calculating $h(x)=1/x$ once on $[1,P]$, and calculating multiplication $P$ times on $[0,1]^2$, each with error no more than $\delta/P^2$, all of which can be done citing \citet{exponentialrelu}.  
\section{Experiments}

To numerically illustrate the behavior of score approximation networks on discontinuous, fractal-like distributions, highlighting the interplay between our representability theorems and real-world network training bottlenecks, we use the algorithm in Appendix G to construct an idealized testbed. We define random variables $x, y_1, y_2$, uniformly distributed over irregular, Cantor-type fractals in $[0,1]$, with Minkowski dimension $d_x={\log 4}/{\log 5}$ and $d_{y_1}=d_{y_2}={\log 2}/{\log 5}$. We then embed them into $\mathbb{R}^{32}$ via random matrices $V_1\in \mathbb{R}^{3\times 32}, V_2\in\mathbb{R}^{4\times 32}$ with orthonormal rows, creating
$$X=\big(x,\sin(\pi x)/20,(1-\cos(\pi x))/20\big)\cdot V_1; Y=\big(y_1,y_2,\sin(\pi y_1)/20,\sin(\pi y_2)/20\big)\cdot V_2.$$
whose distributions $p_X$ and $p_Y$ have identical ambient dimension $n=32$ and Minkowski dimension $d_X=d_Y={\log 4}/{\log 5}$, but distinct support manifold dimension ($d_X'=1; d_Y'=2$). By approximating $\nabla\log p_{X,t}$ and $\nabla\log p_{Y,t}$ with $x$-prediction networks of depth $D=8$, varying widths $W$, and varying early stopping times $t_0$, this provides a controlled environment for investigating the following questions. First, is the network's complexity governed by $d$, $d'$, or the ambient dimension $n$? Second, how does the error $\epsilon$ scale with network width $W$ and early stopping time $t_0$?

\begin{figure}[t]
    \centering
    \begin{subfigure}[b]{0.48\textwidth}
        \centering
        \includegraphics[width=\linewidth]{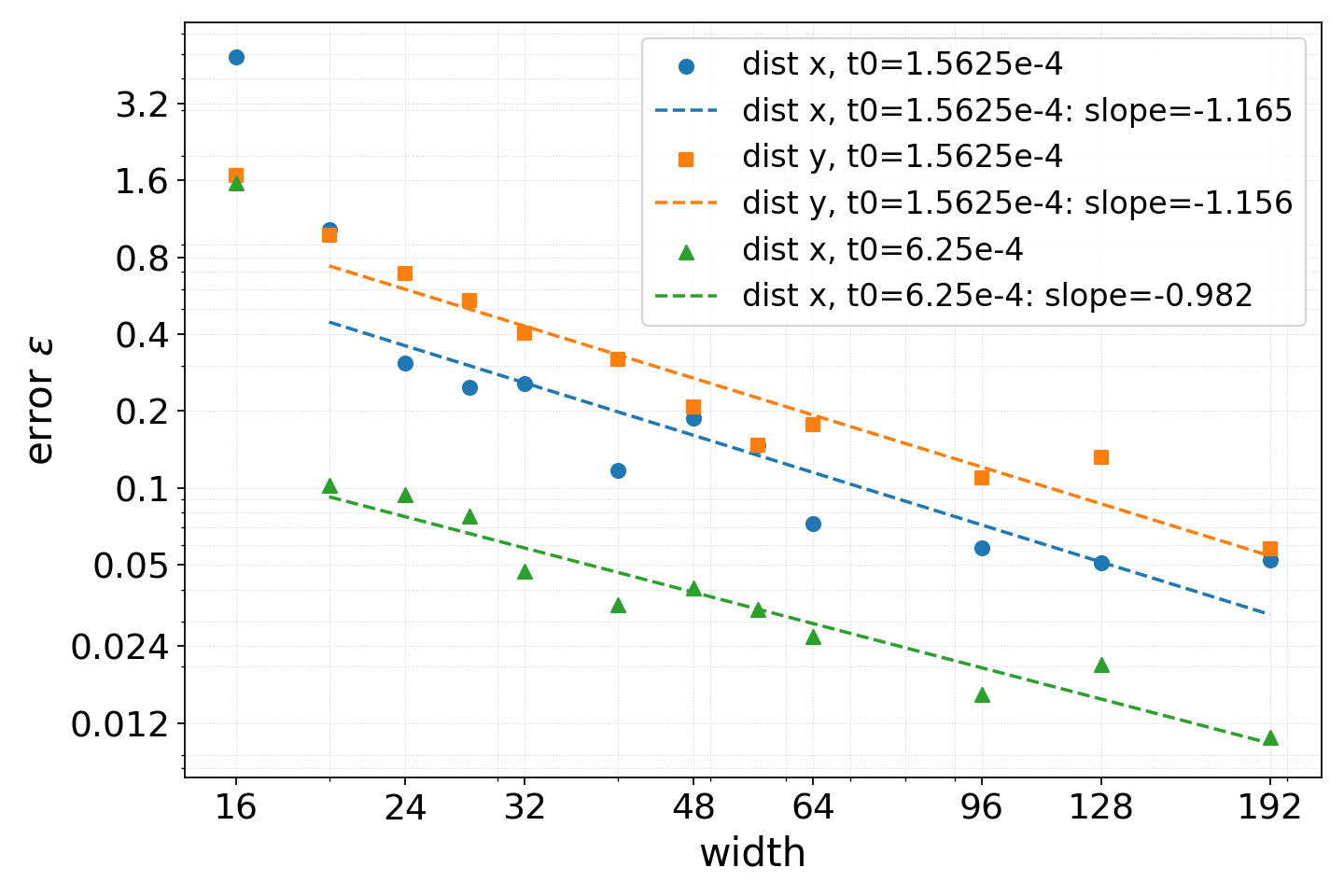}
        \caption{Scaling of $\epsilon$ with $W$ for $p_X$ at $t_0=1.5625e^{-4}$, $p_X$ at $t_0=6.25e^{-4}$, and $p_Y$ at $t_0=1.5625e^{-4}$}
        \label{fig:fractal2d}
    \end{subfigure}
    \hfill
    \begin{subfigure}[b]{0.48\textwidth}
        \centering
        \includegraphics[width=\linewidth]{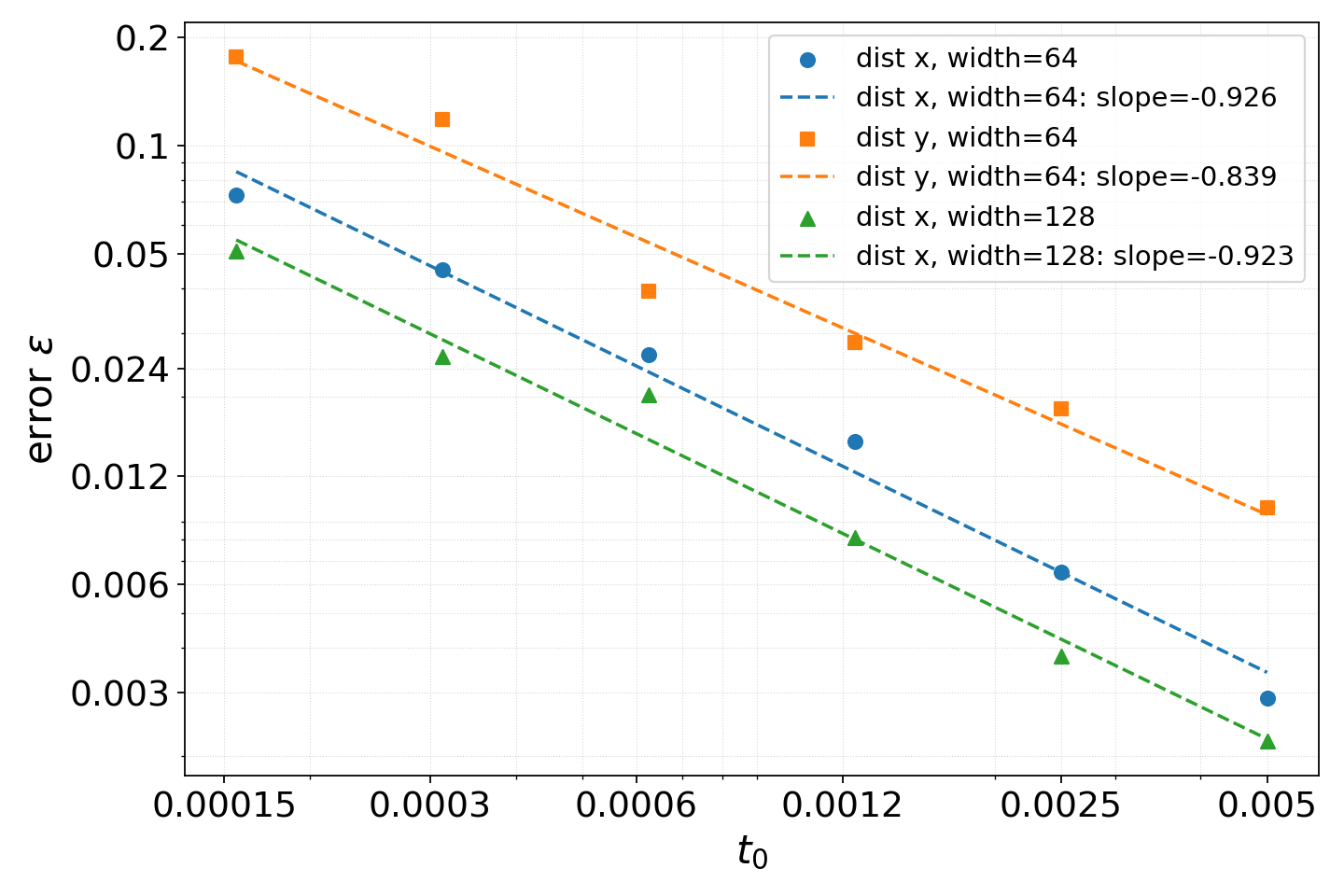}
        \caption{Scaling of $\epsilon$ with $t_0$ for $p_X$ at $W=64$, $p_X$ at $W=128$, and $p_Y$ at $W=64$}
        \label{fig:errwidth}
    \end{subfigure}
    \caption{Experimental results. To match theorem statements, $\epsilon$ is calculated as $\sqrt{(\mathrm{MSE}_{max}/6)}$, $\mathrm{MSE}_{max}$ being the maximum mean squared error observed at a single time frame $t\in [t_0,T]$}
    \label{fig:three}
\end{figure}

Our results \footnote{$W=16$ data points not used for linear regression due to being clear outliers, terminating way too early when encountering an optimization plateau} in Figure 2 provide some evidence. For the first question, ablation studies increasing $n$ to $64$ (Appendix I, Table \ref{tab:raw-n64}) results in a worst-case error increase of only $15\%$. Meanwhile, error of approximating $\nabla\log p_{X,t}$ and $\nabla\log p_{Y,t}$ scale with $W$ and $t_0$ in very similar manners, while complexity bounds attuned to manifold dimension predicts the $W-\epsilon$ slope for $p_X$ to be twice as steep as that for $p_Y$, providing evidence that in our test case, network complexity is dominated by the support's Minkowski dimension $d$ rather than manifold dimension $d'$. 

For the second question, Theorem \ref{thm:main} with $M=1$ results in a scaling of $\epsilon\sim W^{-2/d}\approx W^{-2.3}$, while the trained networks follow
$\epsilon\propto W^{-1.15}$ at $t_0=1.56\times10^{-4}$ and
$W^{-0.98}$ at $t_0=6.25\times10^{-4}$ over $20\le W\le192$. Two effects separate the trained exponent from the existence bound. First, the fit is dominated by a plateau at
large width: at $t_0=1.56\times10^{-4}$ the widths $96$, $128$ and $192$ all reach
$\epsilon\approx0.05$--$0.06$ on $p_X$, and restricting the fit to $W\le64$ steepens the
slopes to $-1.67$ ($p_X$) and $-1.48$ ($p_Y$). Second, while we chose depth $D=8$ due to compute limitations and risks of vanishing gradients, the constructed network of Theorem \ref{Thm:ReLU} require significantly more depth, preventing us from accessing faster rates of convergence. Regarding the dependence on $t_0$, slopes of $\log\epsilon$ against $\log t_0$ are roughly $-0.93$ ($p_X$) and $-0.84$ ($p_Y$), meaning that $\epsilon\,t_0$ only declines marginally as $t_0$ approaches zero. By (\ref{dataperdiction}), this means
that absolute error of $x$-prediction is almost independent of $t_0$. In other words, the $x$-prediction network saturates in capacity
before the finer structure exposed by smaller $t_0$ can be resolved, revealing another training bottleneck. 

While our current work is mainly theoretical, in future studies, we seek to experiment with novel architectural designs that overcome these bottlenecks, rendering the rapid convergence guaranteed by our theoretical bounds truly achievable.

\section{Conclusion}
We build ReLU networks that can approximate the score of any compactly supported distribution. Using a discrete-mixture framework, our network avoids the exponential curse of ambient dimensionality, its complexity governed by the support's Minkowski dimension.

Besides the limitations revealed in numerical experiments, we identify the following limitations warranting future research. First, as with most universal approximation results, our bounds remain highly conservative, a further tightening of which will better guide high-dimensional implementations. Second, our network complexity grows at order at least $\mathcal{O}(t_0^{-d/2})$) as $t_0 \to 0$. We hypothesize that this blowup reflects fundamental information-theoretic lower bounds \citep{tsybakov}, as resolving non-smooth structures at noise scale $\sqrt{t_0}$ encounters a local volume scaling of $\Theta(t_0^{-d/2})$. Establishing a formal minimax bound to verify this remains an exciting open problem.
\section*{Statement of LLM Usage}
Large language models (Claude Fable 5, Gemini 3.1 Pro, and Deepseek v4.0 Flash) were used in this work to aid drafting, proof-checking, and code generation, in the following capacities and no others. (1) Reviewing the essay and providing suggestions for polish and enhancement (2) Verifying the authors' proof line-by-line for gaps, inconsistencies, minor errors, and typos, all of which were manually checked and subsequently repaired by the authors themselves; (3) carrying out, under the authors' explicit instructions, calculation-intensive steps of differentiation, integration, and algebraic manipulation for Theorem 2 and Theorem 4, inside a proof architecture (decomposition and approximation scheme, regularity notion, divide-and-conquer tactic, error analysis strategy, parameter choice) designed entirely by the authors; (4) locating candidate literature
references, each of which was independently retrieved and verified by the authors; and
(5) writing code for the numerical experiment and Appendix H, which gives a detailed description of experiment protocols, under explicit instructions from the authors providing the manifold generation algorithm (see Appendix G), target of experiment (measuring $\epsilon$'s scaling behavior with $t_0$ and $W$), network design ($x$-prediction using fully connected ReLU networks), and list of widths $\{W\}$ and early stopping times $\{t_0\}$ to be tested. All definitions, theorem formulations, proof strategies, and the final form of every argument are the authors' own; every LLM-assisted derivation and citation was verified by at least one author, and the authors take full responsibility for the correctness of all results.

\bibliography{sample}

\appendix

\section{Proof of Theorem 1}
\begin{theorem1}
Let $p_0$ be a probability distribution in $\mathbb{R}^n$, $n\geq 25$, such that $\mathrm{supp}({p_0})$ is compact with diameter $R>3$, and that for all $r<1$, $\mathrm{supp}({p_0})$ can be covered by $C_Pr^{-d}$ balls of radius $r$. Under the Variance Exploding (VE) perturbation scheme ($k(t)=1; \sigma^2(t)=t$), for any target error $\epsilon<\frac{1}{4}$, early stopping time $t_0<\frac{1}{4}$, and order $1\leq M\leq\sqrt{n}$, set the concentration parameter $\lambda$ to be the maximum value in
$$\left\{\sqrt{\frac{15}{14}n+\frac{1}{7}\log C_P+\frac{3}{7}\log \frac{1}{\epsilon}+\frac{2}{7}\log (2R)+\frac{\log n}{14}+\frac{d+6}{14}\log \frac{1}{{t_0}}}+2,\sqrt{n}+2\sqrt{\log \frac{2R}{\epsilon t_0}}+\frac{1}{50}\right\}$$
and set the radius parameter $r$, covering parameter $P$, and network domain parameter $\omega$ to be\footnote{Note that this choice of $r$ naturally satisfies $(3\sqrt{n}+1)r\leq R-\epsilon$ and $r\leq 1/\sqrt{n}$, as is required by Theorem 2 and Theorem 4}
$$r=\min\{\frac{\sqrt{t_0}}{50\lambda},\frac{\sqrt{t_0}}{11\lambda}\cdot (\frac{\epsilon\sqrt{t_0}}{36\lambda})^{\frac{1}{M+1}}\cdot{M}^{\frac{M}{4(M+1)}}\}, P= C_Pr^{-d}, \omega=\frac{R+\sqrt{Tn}+2\sqrt{T\log(2R/\epsilon t_0)}}{t_0}$$
respectively, the true score function $\nabla\log p_t(x)$ can then be approximated via the data-prediction formulation
\begin{align*}
\nabla\log p_{t,appr}(x)=\frac{s_\theta(x,t)-x}{t}
\end{align*}
with mean squared error 
$$\mathrm{E}_{p_t(x)}\left\|\nabla\log p_t(x)-\nabla\log p_{t,appr}(x)\right\|_2^2\leq 6\epsilon^2,\quad \forall t\in[t_0,T],$$
using a fully connected ReLU network $s_\theta(x,t)$ of depth $D$ and width $W$, where\footnote{This matches the claim in the main text, as $\log P=d\log r+\log C_P$, and $\log(M+1)\cdot\frac{1}{(M-1)!}=\mathcal{O}(1)$}
$$D\leq\mathcal{O}\left(M\log \frac{M+1}{\epsilon}+M^2\log n\omega+\log \frac{P\omega}{\epsilon}\log \log \frac{P\omega}{\epsilon}\right); $$
$$W\leq \mathcal{O}\left(P\left(\frac{(n+M)^M}{(M-1)!}\log\frac{n\omega(M+1)}{\epsilon }+\log^2\frac{P\omega}{\epsilon}\right)\right).$$

If, in addition, $\mathrm{supp}(p_0)$ is concentrated on a $C^2$, $d'$-dimensional manifold $\mathcal{M}$ ($d'\geq d$) with positive reach $1/k_1$ such that $$40k_1^2\big(2\sqrt{\log (2R/\epsilon t_0)}+\sqrt{n}\big)^2{n}t_0\leq {d'}, \frac{15nt_0}{14d'}\leq1/4,$$ 
then $\lambda$ can be further shrunk to $\lambda_{\mathcal{M}}$ equaling the maximum value of
$$\left\{\sqrt{d'+\frac{1}{7}\log{(6C_PC_{\mathcal{M}})}+\frac{3}{7}\log\frac{1}{\epsilon}+\frac{2}{7}\log (2R)+\frac{\log n}{14}+\frac{d+6}{14}\log \frac{1}{{t_0}}}+2,\sqrt{d'}+2\sqrt{\log\frac{2R}{\epsilon t_0}}+\frac{1}{50}\right\}$$
while\footnote{This choice of $r$ also satisfies $(3\sqrt{n}+1)r\leq R-\epsilon$ and $r\leq1/\sqrt{n}$, noting that $\sqrt{t_0}/\lambda\leq 1/2\sqrt{n}$} $r$ and $P$ can also be reduced accordingly, at the slight cost of increasing the mean squared error to $7\epsilon^2$. 

Here, $C_\mathcal{M}$ denotes the smallest constant ensuring that for all $r>0$, $\mathcal{M}$ admits a covering $\{B(x_i,r)\}_{i=1}^N$ such that $\forall y_i\in\mathcal{M}$, $B(y_i,\rho)$ contains the centers of at most $C_\mathcal{M}(1+\rho/r)^{d'}$ balls inside $\{B(x_i,r)\}_{i=1}^N$. Such a constant can be proven to exist (see Appendix F) for all compact $C^2$ manifolds with positive reach.
\end{theorem1}
Note: The conditions $R\geq3$, $n\geq25$, $t_0\leq1/4$, $\epsilon\leq1/2$ purely serves the purpose of simplifying long algebraic expressions to avoid piecewise constant tracking. Our theorem also works for $R<3$, $n<25$, etc., though reducing the four parameters below this threshold no longer reduces network complexity accordingly. This can be verified by tracking our usage of the said conditions throughout the theorems, or more conveniently, by (1) adding a point mass at distance 3 from the support and letting its probability approach zero; (2) embedding both $p_0$ and the queried point $x$ into the first $n$ dimensions of $\mathbb{R}^{25}$, and then projecting the approximated $\nabla\log p_t(x)$ back onto $\mathbb{R}^n$ using the method of \cite{wangsubspace} (i.e. keeping the first $n$ coordinates and ignoring the rest); (3) noticing that for $t_0\geq 1/4$, the claim is implied by the case $t_0=1/4$, since $[t_0,T]\subseteq[1/4,T]$; (4) noticing that for $\epsilon\geq1/4$, the claimed is implied by the case $\epsilon=1/4$. 
\subsection{Proof of Part 1 (General Case)}
We prove this result using the theorems 2,3, and 4, whose proofs will be given in the following sections.

$\forall t\in [t_0,T]$ fixed, let $x \sim p_t$. Under the VE scheme, we have 
$$x = x_{sig} + \sqrt{t}z, x_{sig}\sim p_0, z\sim \mathcal{N}(0,I_n).$$
Set a target failure probability $\delta =(\frac{\epsilon t_0}{2R})^2\leq (\frac{\epsilon t}{ R+(3\sqrt{n}+1)r+\epsilon})^2, \forall t\in[t_0,T]$, and let $U_\delta$ denote the set of $(e^{3n}/\delta, n/2, \lambda\sqrt{t})$-EA regular points. By the first statement of Theorem 3, $\mathbb{P}(x_{sig} \in U_\delta) \geq 1 - \delta$.

Next, we bound the magnitude of the Gaussian noise $z$. Here, applying standard Gaussian concentration inequality (and noting $\sqrt{t}\geq 50r$) guarantees that
$$\mathbb{P}\Big(\|z\| \leq \lambda-\frac{r}{\sqrt{t}}\Big)\geq\mathbb{P}\Big(\|z\| \leq \sqrt{n} + \sqrt{2\log(1/\delta)}\Big) \geq 1-\delta.$$ 

Therefore, let $E$ be the intersection of the events $\{x_{sig} \in U_\delta\}$ and $\{\|z\|_2 \leq \sqrt{n}+\sqrt{2\log(1/\delta)}\}$. By the union bound, $\mathbb{P}(E) \geq 1 - 2\delta$. We also know that conditioned on $E$, $x$ must be at most $\lambda\sqrt{t}-r$ distance away from a $(e^{3n}/\delta,n/2,\lambda\sqrt{t})$-EA regular point, to which the results of Theorem 2 (in which we choose $C_1=e^{3n}/\delta$, $C_2=n/2$, view $\mathbb{R}^n$ as its own submanifold with infinite reach, and set $K$ arbitrarily large) apply. 

Now, select a covering $\{B(y_i,r)\}_{i=1}^P$ of $\mathrm{supp}(p_0)$, and construct a score approximation network via the design of Theorem 4 (choosing $J=\sqrt{n}+2\sqrt{\log(2R/\epsilon t_0)}$). The results of Theorem 2 and Theorem 4 ensure that for each $x=x_{sig}+\sqrt{t}z$ satisfying event $E$, it would hold that
$$\left\|\nabla\log p_t(x)-\frac{s_\theta(x,t)-x}{t}\right\|\leq2\epsilon.$$

Conversely, on the complement event $E^c$, Theorem 4 globally confines the output of $s_\theta(x,t)$ to $\mathrm{Conv}(\mathrm{supp}(p_0)) + B(0, 3\sqrt{n}r+t_0\epsilon)$. Therefore, noting that $x+t\nabla\log p_t(x)$ also lies in $\mathrm{Conv(p_0)}$, it would always hold that
$$\left\|\nabla\log p_t(x)-\frac{s_\theta(x,t)-x}{t}\right\|\leq \frac{R+3\sqrt{n}r+t_0\epsilon}{t}\leq\frac{2R}{t}$$
Combining estimates from both regions, the expected mean squared error is bounded by 
\begin{align*}
(2\epsilon)^2\mathbb{P}(E)+\big(\frac{2R}{t}\big)^2\mathbb{P}(E^c)\leq4\epsilon^2+2\delta\big(\frac{2R}{t}\big)^2
\end{align*}
Thus, $\delta\leq(\frac{\epsilon t_0}{2R})^2$ bounds the second term by $2\epsilon^2$. Summing the components, the total expected error is bounded strictly by $6\epsilon^2$. This completes the proof.
\subsection{Proof of Part 2 (Manifold Case)}
We divide the interval $[t_0,T]$ into two parts: $[t_0,t_{mid}]$ and $[t_{mid},T]$, where $t_{mid}=\frac{15nt_0}{14d'}$, and analyze the two intervals separately.

On the interval $[t_0,t_{mid}]$, we set the target failure probability $\delta=(\frac{\epsilon t_0}{2R})^2$, the extension constant $K=\frac{\sqrt{n}+\sqrt{2\log (1/\delta)}+\frac{1}{50}}{\lambda_{\mathcal{M}}}$, and look back at the formula
$$x = x_{sig} + \sqrt{t}z, x_{sig}\sim p_0, z\sim \mathcal{N}(0,I_n).$$
Let $U_\delta$ denote the set of $(C_\mathcal{M}e^{2.5d'}\frac{6}{\delta},\frac{d'}{2},\lambda_{\mathcal{M}}\sqrt{t})$-EA regular points, by Theorem 3, $\mathbb{P}(x_{sig}\in U_\delta)\geq 1-\delta$. 

Also, by the Gaussian concentration inequality, for every $x\in\mathcal{M}$, 
$$\mathbb{P}(\|P_{T_{x}\mathcal{M}}(z)\|\leq \sqrt{d'} + \sqrt{2\log(1/\delta)})\geq 1-\delta; P(\|z\|\leq \sqrt{n} + \sqrt{2\log(1/\delta)})\geq 1-\delta.$$

Therefore, let $E$ denote the event 
$$\{(x_{sig},z)|x_{sig}\in U_\delta, \|P_{T_{x_{sig}}\mathcal{M}}(z)\|\leq\sqrt{d'} + \sqrt{2\log(1/\delta)}, \left\|z\right\|\leq \sqrt{n} + \sqrt{2\log(1/\delta)}\}.$$
We can directly verify (note $\sqrt{n}\geq5$ is at least 250 times as large as $1/50$ and $r/\sqrt{t}$) that
$$6k_1K\lambda_{\mathcal{M}}\sqrt{t_{mid}}=6k_1(\sqrt{n}+\sqrt{2\log (1/\delta)}+\frac{1}{50})\sqrt{\frac{15nt_0}{14d'}}\leq 1,$$
and that
$$\lambda_{\mathcal{M}}-\frac{r}{\sqrt{t}}\geq \sqrt{d'}+\sqrt{2\log(1/\delta)}, K\lambda_{\mathcal{M}}-\frac{r}{\sqrt{t}}\geq\sqrt{n}+\sqrt{2\log (1/\delta)}$$
so Theorem 2 applies to all $\{x=x_{sig}+\sqrt{t}z|(x_{sig},z)\in E\}$
on the entire interval $[t_0,t_{mid}]$, and by the union bound, $\mathbb{P}(E)\geq 1-3\delta$.

We use the following lemma to construct a covering $\{B(y_i,r)\}$ of $\mathrm{supp}(p_0)$ such that $\forall i, y_i\in\mathcal{M}$.

\begin{lemma}
Let $\mathcal{M}$ be a $C^2$ manifold in $\mathbb{R}^n$ with reach $1/k_1$. If a subset $V\subseteq \mathcal{M}$ admits a covering \footnote{In this paper, all balls $B(x_i,r)$ are considered closed balls and all norms $\left\|x\right\|$ are considered 2-norms unless otherwise specified.} $\{B(y_i,r)\}_{i=1}^P$, then there exists $\{z_i\}_{i=1}^P\subseteq\mathcal{M}$ such that $\{B(z_i,r+2k_1r^2)\}_{i=1}^P$ also form a covering of $\mathcal{M}$.
\end{lemma}
\begin{proof}
If $r>\frac{1}{2k_1}$, the claim trivially holds considering that $B(y_i,r)\subseteq B(y,2r)$ for all $y\in B(y_i,r)\cap\mathcal{M}$ (the balls that do not intersect $\mathcal{M}$ can be directly discarded). 

If $r<\frac{1}{2k_1}$, then for each $y_i$ such that $B(y_i,r)\cap\mathcal{M}\neq\emptyset$, there exists a unique point $g(y_i)\in\mathcal{M}$ nearest to $y_i$. In this case, we have:
\begin{itemize}
\item $y_i-g(y_i)=\lambda_y\vec{v}_y, \vec{v}_y\perp T_{g(y_i)}\mathcal{M},\left\|\vec{v}_y\right\|=1$
\item $\forall z\in B(y_i,r)\cap \mathcal{M}$, $z-g(y_i)$ can be decomposed as 
$$z-g(y_i)=\vec{v}_1+\vec{v}_2, \vec{v}_1\perp T_{g(y_i)}\mathcal{M},\vec{v}_2\in T_{g(y_i)}\mathcal{M}, \left\|v_1\right\|\leq2k_1r^2.$$
\end{itemize}
If $\lambda_y\leq2k_1r^2$, then $B(y_i,r)\subseteq B(g(y_i),r+2k_1r^2)$. If $\lambda_y>2k_1r^2$, then considering the fact that $\left\|v_1\right\|\leq2k_1r^2,$
$$\left\|y-z\right\|^2=\left\|y_i-g(y_i)-\vec{v_1}\right\|^2+\left\|\vec{v}_2\right\|^2\geq\left\|(2k_1r^2)\vec{v}_y-\vec{v}_1\right\|^2+\left\|\vec{v}_2\right\|^2=\left\|g(y_i)+(2k_1r^2)\vec{v}_y-z\right\|^2,$$
hence
$$B(y_i,r)\cap \mathcal{M} \subseteq B(g(y_i)+2k_1r^2\vec{v}_y,r)\cap\mathcal{M}\subseteq B(g(y_i),r+2k_1r^2)$$
meaning that the balls $\{B(g(y_i),r+2k_1r^2)\}_{i=1}^P$ also form a covering of $V$.
\end{proof}
This lemma tells us that, as
$$r\leq\frac{\sqrt{t_0}}{50\sqrt{d'}}; 2k_1r\leq \frac{1}{150n}\leq\frac{1}{150d}$$
any covering 
$$\{B(z_i,\frac{150n}{150n+1}r)\}_{i=1}^{P'}, P'\leq C_P(\frac{150n}{150n+1}r)^{-d}\leq C_Pr^{-d}\cdot (1+1/75)$$
can be modified into a covering $\{B(y_i,r)\}_{i=1}^{P'}$ with $y_i\in \mathcal{M}$ for all $i$. As the covering number $P$ only enters the complexity bound  via linear and log terms, the additional $1+1/75$ factor is trivially discarded in the big $\mathcal{O}$-notation.

Thus, we can use Theorem 4 to construct a network that approximates (\ref{discreteapprox}) for all $(x,t)$ in the domain \footnote{note that this network will be used for both $[t_0,t_{mid}]$ and $[t_{mid},T]$, with the same network approximation error. Only the discretization error is analyzed differently.}
$$x\in \mathrm{supp}(p_0)+B\Big(0,R+\sqrt{Tn}+\sqrt{2T\log(1/\delta)}\Big), t\in[t_0,T]$$
and the mean squared error for score approximation on $[t_0,t_{mid}]$ is upper bounded by
$$(2\epsilon)^2\mathbb{P}(E)+(\frac{2R}{t})^2\mathbb{P}(E^c)\leq 4\epsilon^2+3\delta(\frac{2R}{t})^2\leq 7\epsilon^2, \forall t\in [t_0,t_{mid}].$$

For $t\in[t_{mid},T]$, we revert to the general, non-manifold bound. Set 
$$\lambda=\max\left\{\sqrt{\frac{15}{14}n+\frac{1}{7}\log C_P+\frac{3}{7}\log \frac{1}{\epsilon}+\frac{2}{7}\log (2R)+\frac{\log n}{14}+\frac{d+6}{14}\log \frac{1}{{t_0}}}+2,\sqrt{n}+2\sqrt{\log \frac{2R}{\epsilon t_0}}+\frac{1}{50}\right\},$$
we know that $\frac{\lambda_{\mathcal{M}}}{\sqrt{t_0}}\geq\frac{\lambda}{\sqrt{t_{mid}}}$.
Therefore, the previous choice of $r$ still satisfies 
$$r\leq\min\{\frac{\sqrt{t_{mid}}}{50\lambda},\frac{\sqrt{t_{mid}}}{11\lambda}\cdot (\frac{\epsilon\sqrt{t_{mid}}}{36\lambda})^{\frac{1}{M+1}}\cdot{M}^{\frac{M}{4(M+1)}}\},$$
so Theorem 2's bounds on discretization error still applies. Re-using the proof of Part 1 and the already constructed score network, which works for both $[t_0,t_{mid}]$ and $[t_{mid},T]$, finishes the error analysis argument.
\section{Proof of Theorem 2}
\begin{theorem2}
Let $p_0$ be a compact distribution supported on a $C^2$, $d'$-dimensional manifold $\mathcal{M}\subseteq \mathbb{R}^n$ with positive reach $\frac{1}{k_1}$ (For truly arbitrary distributions, $\mathbb{R}^n$ can be viewed as its own submanifold with infinite reach), such that for all $r<1$, $\mathrm{supp}(p_0)$ can be covered by $C_Pr^{-d}$ balls of radius $r$ ($d<d'$). $\forall M\geq 1$, define the function $\mathrm{Clamp}_B$ as  $$\mathrm{Clamp}_B(x)=\max\{-B,\min\{B,x\}\},$$ 
acting position-wise on vectors, and define the functions \footnote{In the definition below, $H_j$ denotes the Hermite tensor polynomial, and we allow ourselves a slight abuse of notation to define the operation $:$ between a $j$-covariant tensor $P$ and a $(j+1)$-contravariant tensor $Q$ as producing a $1-$contravariant tensor $v$ such that $f(v)=(P\otimes f):Q$ for any 1-covariant tensor $f$.} $\tilde{R}_{1,i};\tilde{R}_{2,i}$ as
$$\tilde{R}_{1,i}=\sum_{j=1}^M\frac{1}{j!}\cdot t^{-\frac{j}{2}}\cdot H_j(\frac{x-y_i}{\sqrt{t}}):\big(\int_{B(y_i,r)}(y-y_i)^{\otimes j}dp_i(y)\big)$$
$$\tilde{R}_{2,i}=\int_{B(y_i,r)}(y-y_i)dp_i(y)+\sum_{j=1}^{M-1}\frac{1}{j!}\cdot t^{-\frac{j}{2}}\cdot H_j(\frac{x-y_i}{\sqrt{t}}):\big(\int_{B(y_i,r)}({y-y_i})^{\otimes j+1}dp_i(y)\big).$$
Take an arbitrary covering $\{B(y_i,r)\}_{i=1}^P$ of $\mathrm{supp}(p_0)$ such that $y_i\in\mathcal{M}, \forall i$, and an arbitrary decomposition $p_0=\sum_{i=1}^Pw_ip_i$ such that $\mathrm{supp}(p_i)\subseteq B(y_i,r),\ \forall i$. Under the VE scheme and the assumptions that $\epsilon\leq 1/4,t\leq 1/k_1^2,$ by choosing
$$\lambda(\epsilon)\geq \sqrt{\frac{9}{7}C_2+\frac{1}{7}\log \frac{C_1C_p}{\epsilon}+\frac{\log n}{14}+\frac{d+2}{14}\log \frac{1}{{t_c}}}+2; K\leq\frac{1}{6k_1\lambda(\epsilon)\sqrt{t}};$$
$$r\leq\min\big\{\frac{\sqrt{t_c}}{50\lambda(\epsilon)},\frac{\sqrt{t_c}}{11\lambda(\epsilon)}\cdot (\frac{\epsilon\sqrt{t_c}}{36\lambda(\epsilon)})^{\frac{1}{M+1}}\cdot{M}^{\frac{M}{4(M+1)}}\big\}, t_c=\min\{t,1/4\}$$
and defining 
$$\tilde{p}_{i,t}(x)=e^{-\frac{\left\|x-y_i\right\|^2}{2t}}\cdot \big(1+\mathrm{Clamp_{\frac{1}{3}}(\tilde{R}_{1,i}})\big), \tilde{q}_{i,t}(x)=y_i+\frac{\mathrm{Clamp}_{2r}(\tilde{R}_{2,i})}{1+\mathrm{Clamp_{\frac{1}{3}}(\tilde{R}_{1,i}})}$$
the discrete approximation (equivalent to the Softmax expression (\ref{scoredecomposition}))
$$
\nabla\log p_{t}(x)\approx \frac{1}{t}\frac{\sum_{i=1}^Pw_i\tilde{p}_{i,t}(x)\tilde{q}_{i,t}(x)}{\sum_{i=1}^Pw_i\tilde{p}_{i,t}(x)}-\frac{x}{t}
$$
has error bounded by $\epsilon$, if $x\in B(z, K\lambda(\epsilon)\sqrt{t})$
holds for some $(C_1, C_2, \lambda(\epsilon)\sqrt{t})$-EA regular point $z$, and the projection of $x-z$ onto $z$'s tangent space $T_z\mathcal{M}$ has norm at most $\lambda(\epsilon)\sqrt{t}$.
\end{theorem2}
\subsection{Proof Step 1: Taylor Expansion}
Denoting 
$$p_{i,t}(x)=\int_{B(y_i,r)}e^{-\frac{\left\|x-y\right\|^2}{2t}}dp_i(y); \nabla\log p_{i,t}(x)=\frac{\int_{B(y_i,r)}e^{-\frac{\left\|x-y\right\|^2}{2t}}\cdot \frac{y-y_i}{t}dp_i(y)}{\int_{B(y_i,r)}e^{-\frac{\left\|x-y\right\|^2}{2t}}dp_i(y)}+\frac{y_i-x}{t},$$
Taylor expanding $e^{-\frac{\left\|x-y\right\|^2}{2t}}$ at reference point $y_i$ proves $p_{i,t}(x)$ to equal
$$\int_{B(y_i,r)}\Big[\sum_{j=0}^M\frac{1}{j!}(\nabla^j_ye^{-\frac{\left\|x-y\right\|^2}{2t}}(y_i)):(y-y_i)^{\otimes j}+\frac{1}{(M+1)!}(\nabla_{y}^{M+1}e^{-\frac{\left\|x-y\right\|^2}{2t}}(y')):(y-y_i)^{\otimes M+1}\Big]dp_i(y).$$
Therefore, noting that that 
$$\nabla_y^je^{-\frac{\left\|x-y\right\|^2}{2t}}(y’)=t^{-\frac{j}{2}}e^{-\frac{\left\|x-y'\right\|^2}{2t}}H_j(\frac{x-y'}{\sqrt{t}}),$$ 
where $H_j(\lambda)$, the Hermite tensor polynomial\citep{hermite}, is a tensor with $n^j$ entries, all of which are polynomial in the $n$ components of $\lambda$ and fully independent of $t$, $p_{i,t}(x)$ further equals $e^{-\frac{\left\|x-y_i\right\|^2}{2t}}$ times
$$1+\sum_{j=1}^M\frac{1}{j!}\cdot t^{-\frac{j}{2}}\cdot H_j(\frac{x-y_i}{\sqrt{t}}):\big(\int_{B(y_i,r)}(y-y_i)^{\otimes j}dp_i(y)\big)+\Delta_{M+1},$$
where $\Delta_{M+1}$ equals 
$$\frac{1}{(M+1)!}\cdot t^{-\frac{M+1}{2}}\int_{B(y_i,r)}e^{\frac{\|x-y_i\|^2-\left\|x-y'\right\|^2}{2t}}H_{M+1}(\frac{x-y'}{\sqrt{t}}):(y-y_i)^{\otimes M+1}dp_i(y).$$
Similarly, Taylor expanding $e^{-\frac{\left\|x-y\right\|^2}{2t}}$ inside $\int_{B(y_i,r)} e^{-\frac{\left\|x-y\right\|^2}{2t}}\cdot (y-y_i)dp_i(y)$ (note that the $y-y_i$ term is kept intact from Taylor expansion) at reference point $y_i$ to the $M-1$th order, intentionally one order lower to prevent the need of additionally storing a higher order tensor 
$$\int_{B(y_i,r)}({y-y_i})^{\otimes M+1}dp_i(y),$$
we deduce that this term is equal to 
$$\int_{B(y_i,r)}(y-y_i)\cdot\Big[\sum_{j=1}^{M-1}\frac{1}{j!}(\nabla^j_ye^{-\frac{\left\|x-y\right\|^2}{2t}}(y_i)):({y-y_i})^{\otimes j}+\frac{1}{M!}(\nabla_{y}^{M}e^{-\frac{\left\|x-y\right\|^2}{2t}}(y')):(y-y_i)^{\otimes M}\Big]dp_i(y)$$
and hence $e^{-\frac{\left\|x-y_i\right\|^2}{2t}}$ times
$$\int_{B(y_i,r)}(y-y_i)dp_i(y)+\sum_{j=1}^{M-1}\frac{1}{j!}\cdot t^{-\frac{j}{2}}\cdot H_j(\frac{x-y_i}{\sqrt{t}}):\Big(\int_{B(y_i,r)}({y-y_i})^{\otimes j+1}dp_i(y)\Big)+\tilde{\Delta}_{M},$$
where $\tilde{\Delta}_{M}$ equals 
$$\frac{1}{M!}\cdot t^{-\frac{M}{2}}\int_{B(y_i,r)}e^{\frac{\left\|x-y_i\right\|^2-\left\|x-y'\right\|^2}{2t}}H_{M}(\frac{x-y'}{\sqrt{t}}):(y-y_i)^{\otimes M+1}dp_i(y).$$
These error formulas allow us to conduct a more accurate single-kernel analysis using the property of Hermite tensor polynomials.
\subsection{Proof Step 2: Single Kernel Error Analysis}
The properties of manifold reach \citep{reachproperty} guarantee that $\forall y,y'\in \mathcal{M}, \forall \vec{v}\perp T_{y'}\mathcal{M},$
\begin{align}
\label{smoothnesscond}
 \|P_{(T_y\mathcal{M})^\perp}(y'-y)\|\leq \frac{1}{2}k_1\left\|y'-y\right\|^2,\left\|P_{T_y\mathcal{M}}(\vec{v})\right\|\leq k_1\left\|y'-y\right\|\|\vec{v}\|.
\end{align}
Thus, $\forall y_i\in\mathcal{M}$, $\forall y\in \mathrm{Conv}(\mathcal{M}\cap B(y_i,r))$, $y-{y}_i$ can be decomposed as $h_{1,i}\vec{\omega}_{1,i}+h_{2,i}\vec{\omega}_{2,i}$, $h_{1,i}\leq \frac{k_1}{2}r^2$; $h_{2,i}\leq r$; and $\vec{\omega}_{1,i}, \vec{\omega}_{2,i}$ are unit vectors such that $\vec{\omega}_{1,i}\perp T_{{y}_i}\mathcal{M}$; $\vec{\omega}_{2,i}\in T_{{y}_i}\mathcal{M}$. In addition, assuming that 
\begin{align}
\label{decompx}
x-y_i=\lambda_{1,i}\vec{v}_{1,i}+\lambda_{2,i}\vec{v}_{2,i}, \lambda_{1,i}>0, \lambda_{2,i}>0, \vec{v}_{1,i}\perp T_{{y}_i}\mathcal{M}, \vec{v}_{2,i}\in T_{{y}_i}\mathcal{M},
\end{align}
the inequality
$$\left\|x-y_1\right\|_2^2-\left\|x-y_2\right\|_2^2\leq(\lambda_{1,i}+\frac{1}{2}k_1r^2)^2+(\lambda_{2,i}+r)^2-(\lambda_{1,i}-\frac{1}{2}k_1r^2)^2-(\lambda_{2,i}-r)^2$$
holds for all $y_1,y_2\in \mathrm{Conv}(\mathcal{M}\cap B(y_i,r))$. 

Using these basic facts, we analyze the error of our approximation formula for a single kernel. We begin by analyzing the residual terms $\Delta_{M+1}$ and $\tilde{\Delta}_{M}$.

As $y'\in \mathrm{Conv}(\mathcal{M}\cap B(y_i,r))$ holds for all $y\in B(y_i,r)$, we know that 
$$e^{{\frac{\left\|x-y_i\right\|^2}{2t}-\frac{\left\|x-y'\right\|^2}{2t}}}\leq e^{F(r,t,\lambda_{1,i},\lambda_{2,i})}$$
hence
\begin{align}
\label{simpleboundforp}
\log p_{i,t}(x)+\frac{\left\|x-y_i\right\|^2}{2t}\leq F(r,t,\lambda_{1,i},\lambda_{2,i}),
\end{align}
where we denote 
$$F(r,t,\lambda_{1,i},\lambda_{2,i}):=\frac{\frac{1}{2}k_1r^2(\lambda_{1,i}+r)+r(\lambda_{2,i}+r)}{t}.$$
Meanwhile, a known property of Hermite tensor polynomials shows that for any unit vector $v$,
$$H_k(u):v^{\otimes k}=He_k(\langle u,v\rangle)$$
where 
$$He_{k}(\lambda)=(-1)^{k}e^{\frac{\lambda^2}{2}}\frac{d^{k}}{d\lambda^{k}}(e^{-\frac{\lambda^2}{2}})$$
is the scalar Hermite polynomial which admits the expression
$$He_{k}(\lambda)=\sum_{m=0}^{\lfloor k/2\rfloor}(-1)^m\frac{k!}{2^m m!(k-2m)!}\lambda^{k-2m},$$
an upper bound for which is provided by the following lemma.
\begin{lemma}
$\forall k, \forall \lambda_{max}\geq2, \forall \lambda\in[-\lambda_{max},\lambda_{max}],|He_k(\lambda)|\leq \sqrt{k!}(\lambda_{max})^k$.
\end{lemma}
\begin{proof}
As $He_k(\lambda)$ is given by the recursive formula
$$He_k(\lambda)=\lambda\cdot He_{k-1}(\lambda)-(k-1)\cdot He_{k-2}(\lambda),$$
we know that 
$$|He_k(\lambda)|\leq\lambda_{max}|He_{k-1}(\lambda)|+(k-1)|He_{k-2}(\lambda)|,He_0(\lambda)=1, |He_1(\lambda)|\leq\lambda_{max}$$
Meanwhile we also have that
$$\sqrt{(k)!}(\lambda_{max})^k\geq \lambda_{max}\sqrt{(k-1)!}(\lambda_{max})^{k-1}+(k-1)\sqrt{(k-2)!}(\lambda_{max})^{k-2},\forall k \geq 2, \lambda_{max}\geq 2,$$
so the stated property follows from mathematical induction.
\end{proof}
Combining the above properties, $\Delta _{M+1}$ is upper bounded by
\begin{align}
\label{valbound}
e^{F(r,t,\lambda_{1,i},\lambda_{2,i})}\cdot\frac{1}{\sqrt{(M+1)!}}\tilde{F}(r,t,\lambda_{1,i},\lambda_{2,i})^{M+1},
\end{align}
where $\tilde{F}(r,t,\lambda_{1,i},\lambda_{2,i})=\max\{F(r,t,\lambda_{1,i},\lambda_{2,i}),\frac{2r}{\sqrt{t}}\}$. Similarly, noting that under our definition, $\forall$ 1-covariant tensor $f$,
$$f:\big(H_{M}(\frac{x-y'}{\sqrt{t}}):(y-y_i)^{\otimes M+1}\big)=\big(H_{M}(\frac{x-y'}{\sqrt{t}}):(y-y_i)^{\otimes M}\big)\big(f(y-y_i)\big),$$
hence
$$\left\|H_{M}(\frac{x-y'}{\sqrt{t}}):(y-y_i)^{\otimes M+1}\right\|\leq \left|H_{M}(\frac{x-y'}{\sqrt{t}}):(y-y_i)^{\otimes M}\right|\cdot \left\|y-y_i\right\|,$$ 
$\tilde{\Delta}_{M}$ is upper bounded by
\begin{align}
\label{gradbound}
e^{F(r,t,\lambda_{1,i},\lambda_{2,i})}\frac{1}{\sqrt{M!}}\tilde{F}(r,t,\lambda_{1,i},\lambda_{2,i})^M\cdot r
\end{align}
We then consider necessary conditions under which the clamping activates.
From the triangle inequality, we know that 
$$|\tilde{R}_{1,i}(x)|\leq(|p_{i,t}(x)-e^{-\frac{\left\|x-y_i\right\|^2}{2t}}|)\cdot e^{\frac{\left\|x-y_i\right\|^2}{2t}}+|\Delta_{M+1}|,$$
which is upper bounded by
$$e^{F(r,t,\lambda_{1,i},\lambda_{2,i})}\cdot\Big(1+\frac{1}{\sqrt{(M+1)!}}\tilde{F}(r,t,\lambda_{1,i},\lambda_{2,i})^{M+1}\Big)-1.$$
Similarly, $\|\tilde{R}_{2,i}(x)\|$ is upper bounded by 
$$\left\|\int_{B(y_i,r)}e^{\frac{\left\|x-y_i\right\|^2-\left\|x-y\right\|^2}{2t}}\cdot (y-y_i)dp_i(y)\right\|+\left\|\tilde{\Delta}_M\right\|,$$
hence
$$re^{F(r,t,\lambda_{1,i},\lambda_{2,i})}\cdot \Big(1+\frac{1}{\sqrt{M!}}\tilde{F}(r,t,\lambda_{1,i},\lambda_{2,i})^M\Big)$$
Thus, we can ensure that the clamping never activates upon calculating $\tilde{R}_{1,i}$ and $\tilde{R}_{2,i}$ when
\begin{align}
\label{noclamp}
r\leq\frac{\sqrt{t}}{20}; e^{F(r,t,\lambda_{1,i},\lambda_{2,i})}\leq\frac{5}{4}.
\end{align}
Finally, as $\left\|y_i-(t\nabla\log p_{i,t}(x)+x)\right\|\leq r$, our clamping criterion guarantees
$$|\log p_{i,t}(x)-\log \tilde{p}_{i,t}(x)|\leq \log \frac{3}{2}+F(r,t,\lambda_{1,i},\lambda_{2,i})$$
$$\|\tilde{q}_{i,t}(x)-(t\nabla\log p_{i,t}(x)+x)\|\leq(3\sqrt{n}+1)r$$
as upper bounds for the extreme-case approximation error.

\subsection{Proof Step 3: Decomposition into Near and Far Groups}
Consider a point $x$ such that $\lambda_{1,i};\lambda_{2,i}$ as defined of (\ref{decompx}) satisfies $\sqrt{\lambda_{1,i}^2+\lambda_{2,i}^2}=\rho_1,\lambda_{2,i}=\rho_2$ for some $y_i\in\mathcal{M}$, and consider a new point ${y}_j\in\mathcal{M}$ some distance away from ${y}_i$. Here, $x-{y}_j$ can be decomposed as 
$$\lambda_{1,j}\vec{v}_{1,j}+\lambda_{2,j}\vec{v}_{2,j},\vec{v}_{1,j}\perp T_{{y}_j}\mathcal{M},\vec{v}_{2,j}\in T_{{y}_j}\mathcal{M},$$
where we have 
$$\lambda_{1,j}\leq \rho_1+\left\|{y}_j-{y}_i\right\|;\lambda_{2,j}\leq \rho_2+\left\|y_j-y_i\right\|+(k_1\left\|{y}_j-{y}_i\right\|)\cdot \rho_1.$$

We fix a $\lambda(\epsilon)$ large, a $K$ large, and an $r$ small, and impose the initial constraints 
\begin{align}
\label{baseconstraint}
t\leq\frac{1}{k_1^2}; K\leq\frac{1}{6k_1\lambda(\epsilon)\sqrt{t}} ; r\leq\min\{\frac{\sqrt{t}}{20},\frac{1}{40}\}; \lambda(\epsilon)>2.
\end{align}

Now, consider an $x$ for which there exists a $(C_1, C_2,\lambda(\epsilon)\sqrt{t})$-EA regular point $z\in\mathcal{M}$ and unit vectors $\vec{v}_{1,z};\vec{v}_{2,z}$, $\vec{v}_{1,z}\perp T_z\mathcal{M},\vec{v}_{2,z}\in T_z\mathcal{M}$, such that
\begin{align}
\label{distancecond}
x-z=\lambda_{1,z}\vec{v}_{1,z}+\lambda_{2,z}\vec{v}_{2,z}; \sqrt{\lambda_{1,z}^2+\lambda_{2,z}^2}\leq K\lambda(\epsilon)\sqrt{t}-r, \lambda_{2,z}\leq \lambda(\epsilon)\sqrt{t}-r.
\end{align}
According to the definition of $(C_1,C_2,r)$-EA regularity, there exists a ball $B(z_0,\lambda(\epsilon)\sqrt{t})$ satisfying the regularity inequality such that $\left\|z-z_0\right\|\leq\lambda(\epsilon)\sqrt{t}$. In addition, $p_0$ admits a covering of balls 
$$\{B(z,8\lambda(\epsilon)\sqrt{t}-r)\}\cup\{B(z_i,\lambda(\epsilon)\sqrt{t}-r)\}_{i=1}^N,\quad N\leq C_P(\lambda(\epsilon)\sqrt{t}-r)^{-d},$$
where, noting that \footnote{The argument goes as follows: consider a point $z$ in $B(\lambda R\vec{v},R)\backslash B(0,8R)$ where $\lambda<\sqrt{63}$. Decomposing $z=z_1+z_2$ where $z_1\parallel \vec{v}$ and $z_2\perp \vec{v}$, we know that $\left\|z_1\right\|>\sqrt{63}R$ as $\left\|z_2\right\|<R$, therefore, $$\left\|z-\sqrt{63}R\vec{v}\right\|^2=(\left\|z_1\right\|-\sqrt{63}R)^2+\left\|z_2\right\|^2\leq(\left\|z_1\right\|-\lambda R)^2+\left\|z_2\right\|^2=\left\|z-\lambda R\vec{v}\right\|_2^2$$}
$$B(0,8R)\cup B(\lambda R\vec{v},R)\subseteq B(0,8R)\cup B(\sqrt{63}R\vec{v},R),\forall R,\left\|\vec{v}\right\|=1,\lambda<\sqrt{63},$$
we assume 
$$\left\|z_i-z\right\|\geq\sqrt{63}(\lambda(\epsilon)\sqrt{t}-r)$$ 
without loss of generality.

we decompose all kernels into near groups and far groups in the following way: for a fixed $r$-covering $\{B(y_i,r)\}_{i=1}^{P(r)}$ of $p_0$, we choose a decomposition $A_0,A_1,...,A_N$ of $\{1,2,...,P(r)\}$ such that
$$y_i\in B(z,8\lambda(\epsilon)\sqrt{t}-r),\forall i \in A_0$$
$$y_i\in B(z_j,\lambda(\epsilon)\sqrt{t}-r)\backslash B(z,8\lambda(\epsilon)\sqrt{t}-r),\forall i \in A_j$$
and denote $\tilde{A}_0:=\{i|y_i\in B(z,2\lambda(\epsilon)\sqrt{t}+r)\}$.

We now consider bounding the far groups $\{A_j\}_{j\geq1}$. As $\left\|{y}_i-z\right\|\leq\left\|z_j-z\right\|+\lambda(\epsilon)\sqrt{t}-r$ holds for every kernel $B(y_i,r)$ in $A_j$, condition (\ref{distancecond}) and the fact that $\mathcal{M}$ is $C^2$ with reach $1/k_1$ ensure that in the decomposition
$$x-{y}_i=\lambda_{1,i}\vec{v}_{1,i}+\lambda_{2,i}\vec{v}_{2,i}, \vec{v}_{1,i}\perp T_{y_i}\mathcal{M}, \vec{v}_{2,i}\in T_{y_i}\mathcal{M},\left\|\vec{v}_{1,i}\right\|=\left\|\vec{v}_{2,i}\right\|=1,$$
we have
$$\lambda_{1,i}\leq (K+1)\lambda(\epsilon)\sqrt{t}+\left\|z-z_j\right\|-r;\lambda_{2,i}\leq2\lambda(\epsilon)\sqrt{t}+\left\|z-z_j\right\|+(\lambda(\epsilon)\sqrt{t}+\left\|z-z_j\right\|)k_1K\lambda(\epsilon)\sqrt{t}-r$$
As a result, the
difference between $\sum_{i\in A_j}w_ip_{i,t}(x)$ and $\sum_{i\in A_j}w_i\tilde{p}_{i,t}(x)$ is upper bounded by 
$$\min\{\sum_{i\in A_j}w_i\tilde{p}_{i,t}(x),\sum_{i\in A_j}w_ip_{i,t}(x)\}\cdot\big(\frac{3}{2}e^{\phi(r,t,\left\|z-z_j\right\|)}-1\big),$$
where $\phi(r,t,\left\|z-z_j\right\|)$ denotes the value of $F(r,t,\lambda_{1,i},\lambda_{2,i})$ when
$$\lambda_{1,i}= (K+1)\lambda(\epsilon)\sqrt{t}+\left\|z-z_j\right\|-r;\lambda_{2,i}=2\lambda(\epsilon)\sqrt{t}+\left\|z-z_j\right\|+(\lambda(\epsilon)\sqrt{t}+\left\|z-z_j\right\|)k_1K\lambda(\epsilon)\sqrt{t}-r,$$
and hence
$$\frac{\partial \phi}{\partial \left\|z-z_j\right\|}=\frac{\frac{1}{2}k_1r^2+r(1+k_1K\lambda(\epsilon)\sqrt{t})}{t}\leq\frac{2r}{t}$$

What's more, as $\left\|y_i-z\right\|\leq2\lambda(\epsilon)\sqrt{t}+r$ holds for all $B(y_i,r)$ that intersects $B(z_0,\lambda(\epsilon)\sqrt{t})$, while $B(y_i,r)\subseteq B(z_j,(\lambda(\epsilon)\sqrt{t}-r)+r)$ holds for all $i\in A_j$, we also have that
\begin{align}
\label{weightbound}
\mu\big(B(z_0,\lambda(\epsilon)\sqrt{t}\big)\leq\sum_{i\in\tilde{A}_0}w_i; \mu(B(z_j,\lambda(\epsilon)\sqrt{t}))\geq \sum_{i\in A_j}w_i.
\end{align}

\subsection{Proof Step 4: Bounding Far Group Contributions}
$\forall y \in \mathcal{M}$, the conditions (\ref{distancecond}) and (\ref{smoothnesscond}) guarantee that 

$$|\langle x-z,z-y\rangle|\leq \lambda(\epsilon)\sqrt{t}\left\|z-y\right\|+\frac{1}{2}k_1K\lambda(\epsilon)\sqrt{t}\left\|z-y\right\|^2$$

Thus, $\forall i \in \tilde{A}_0, y\in \mathrm{Conv}(B(y_i,r)\cap\mathcal{M}),$ the fact that $r<\frac{1}{40}\lambda(\epsilon)\sqrt{t}$, hence $2\lambda(\epsilon)\sqrt{t}+2r\leq\frac{41}{20}\lambda(\epsilon)\sqrt{t}$, guarantees that 
$$\left\|x-y\right\|^2=\left\|x-z\right\|^2+\left\|z-y\right\|_2^2+2\langle x-z,z-y\rangle\leq\left\|x-z\right\|_2^2+9\lambda^2(\epsilon)t+\frac{9}{2}k_1K\lambda^3(\epsilon)t\sqrt{t}$$
while  $\forall i\in {A}_j, j\geq 1, y\in \mathrm{Conv}(B(y_i,r)\cap\mathcal{M})$, $\left\|x-y\right\|^2$ is no smaller than $\left\|x-z\right\|^2$ plus
$$\left\|z-y\right\|^2-2\lambda(\epsilon)\sqrt{t}\left\|z-y\right\|-k_1K\lambda(\epsilon)\sqrt{t}\left\|z-y\right\|^2.$$
Here, (\ref{baseconstraint}) ensures that $k_1K\lambda(\epsilon)\sqrt{t}<1/6$, so the fact $\left\|z-y\right\|-\left\|z-z_j\right\|\in(-\lambda(\epsilon)\sqrt{t},\lambda(\epsilon)\sqrt{t})$ ensures that whenever $\left\|z-z_j\right\|>6\lambda(\epsilon)\sqrt{t}$, $\left\|x-y\right\|^2$ is no smaller than $\left\|x-z\right\|_2^2$ plus
$$G(\left\|z-z_j\right\|)=(\left\|z-z_j\right\|-\lambda(\epsilon)\sqrt{t})^2-2\lambda(\epsilon)\sqrt{t}(\left\|z-z_j\right\|-\lambda(\epsilon)\sqrt{t})-k_1K\lambda(\epsilon)\sqrt{t}(\left\|z-z_j\right\|-\lambda(\epsilon)\sqrt{t})^2,$$
where we have 
$$\frac{\partial G}{\partial \left\|z-z_j\right\|}=2(\left\|z-z_j\right\|-\lambda(\epsilon)\sqrt{t})-2\lambda(\epsilon)\sqrt{t}-2k_1K\lambda(\epsilon)\sqrt{t}(\left\|z-z_j\right\|-\lambda(\epsilon)\sqrt{t})\geq\frac{5}{3}\left\|z-z_j\right\|-4\lambda(\epsilon)\sqrt{t}.$$

Thus,
$\forall \alpha \in \tilde{A}_0, \beta\in A_j$, as $y_\alpha\in\mathcal{M}, y_\beta\in\mathcal{M}$, $\mathrm{supp}(p_\alpha)\subseteq B(y_\alpha,r)\cap\mathcal{M}$, $\mathrm{supp}(p_\beta)\subseteq B(y_\beta,r)\cap\mathcal{M}$, and $$p_{\alpha,t}(x)=\int_{\mathrm{supp}(p_\alpha)}e^{-\frac{\|x-y\|^2}{2t}}dp_\alpha(y), p_{\beta,t}(x)=\int_{\mathrm{supp}(p_\beta)}e^{-\frac{\|x-y\|^2}{2t}}dp_\beta(y),$$
we have
$$\frac{\min\{p_{\alpha,t}(x),e^{-\frac{\left\|x-y_\alpha\right\|^2}{2t}}\}}{\max\{p_{\beta,t}(x),e^{-\frac{\|x-y_\beta\|^2}{2t}}\}}\geq \exp\big(\frac{G(\left\|z-z_j\right\|)-9\lambda^2(\epsilon)t-\frac{9}{2}k_1K\lambda^3(\epsilon)t\sqrt{t}}{2t}\big)$$
Therefore, again using formula (\ref{weightbound}) and the definition of $(C_1,C_2,r)$-EA regularity , 
$$\big|\sum_{i\in A_j}w_ip_{i,t}(x)-\sum_{i\in A_j}w_i\tilde{p}_{i,t}(x)\big|$$
is upper bounded by $\frac{3}{2}C_1 H_1(x,t)e^{\phi(r,t,\left\|z-z_j\right\|)}E(t,\left\|z-z_j\right\|)$ (with $\frac{3}{2}e^{\phi(x,t)}-1$ simplified to $\frac{3}{2}e^{\phi(x,t)}$ for convenience), where 
$$H_1(x,t)=\min\Big\{\frac{3}{2}\sum_{i\in \tilde{A}_0}w_i \tilde{p}_{i,t}(x),\sum_{i\in\tilde{A}_0}w_i p_{i,t}(x)\Big\}$$
and $E(r,t,\left\|z-z_j\right\|)$ is such that $\log E(r,t,\left\|z-z_j\right\|)$ equals
$$C_2\left({\frac{\left\|z-z_0\right\|+\left\|z_j-z\right\|}{\lambda(\epsilon)\sqrt{t}}}\right)+\frac{1}{2}({9\lambda^2(\epsilon)+\frac{9}{2}k_1K\lambda^3(\epsilon)\sqrt{t}}-\frac{G(\left\|z-z_j\right\|)}{t}).$$
Taking derivatives with respect to $\left\|z-z_j\right\|$ gives $\frac{3}{2}C_1 H_1(x,t)e^{\phi(r,t,\left\|z-z_j\right\|)}E(t,\left\|z-z_j\right\|)$ times
$$\frac{\partial \phi}{\partial \left\|z-z_j\right\|}+\frac{C_2}{\lambda(\epsilon)\sqrt{t}}-\frac{1}{2t}\frac{\partial G}{\partial\left\|z-z_j\right\|},$$
which is consistently smaller than zero on the entire $[7\lambda(\epsilon)\sqrt{t},+\infty)$ whenever $\lambda(\epsilon)>\sqrt{C_2}+2$.

As $\sqrt{63}(\lambda(\epsilon)\sqrt{t}-r)\geq 7.7\lambda(\epsilon)\sqrt{t}$, the term $\frac{3}{2}C_1e^{\phi(r,t,\left\|z-z_j\right\|)}E(t,\left\|z-z_j\right\|)$ is upper bounded by  
$$\frac{3}{2}C_1\exp\Big(\phi(r,t,7.7\lambda(\epsilon)\sqrt{t})+C_2(\frac{\lambda(\epsilon)\sqrt{t}+7.7\lambda(\epsilon)\sqrt{t}}{\lambda(\epsilon)\sqrt{t}})+\frac{1}{2}(9\lambda^2(\epsilon)+\frac{9}{2}k_1K\lambda^3(\epsilon)\sqrt{t}-\frac{G(7.7\lambda(\epsilon)\sqrt{t})}{t})\Big),$$
and hence
$$\frac{3}{2}C_1\cdot \exp\Big(\frac{\frac{9}{2}k_1\lambda(\epsilon)\sqrt{t}r^2+\frac{1}{2}k_1K\lambda(\epsilon)\sqrt{t}r^2+12\lambda(\epsilon)\sqrt{t}r}{t}+9C_2+25k_1K\lambda^3(\epsilon)\sqrt{t}-11.2\lambda^2(\epsilon)\Big),$$
and hence 
$$D(\lambda(\epsilon))=\frac{3}{2}C_1\cdot \exp \Big(\frac{13\lambda(\epsilon)r}{\sqrt{t}}+9C_2-7\lambda^2(\epsilon)\Big).$$
Similarly, for the $z$ satisfying condition (\ref{distancecond}), 
$$\left\|\sum_{i\in A_j}w_i p_{i,t}(x)(\nabla\log p_{i,t}(x)-\frac{z-x}{t})-\sum_{i\in A_j}w_i\tilde{p}_{i,t}(x)\frac{\tilde{q}_{i,t}(x)-z}{t}\right\|$$
is upper bounded by 
$$\sum_{i\in A_j}w_i\tilde{p}_{i,t}(x)\cdot\frac{(3\sqrt{n}+1)r}{t}+\big|\sum_{i\in A_j}w_i\big(p_{i,t}(x)-\tilde{p}_{i,t}(x)\big)\big|\cdot\frac{\left\|z-z_j\right\|+\lambda(\epsilon)\sqrt{t}}{t},$$
and, noting that $\tilde{p}_{i,t}(x)\leq\frac{4}{3}e^{-\frac{\left\|x-y_i\right\|^2}{2t}},$ is further upper bounded by the sum of 
$$\frac{4}{3}C_1H_1(x,t)E(t,\left\|z-z_j\right\|)\frac{(3\sqrt{n}+1)r}{t}$$
and
$$\frac{3}{2}C_1H_1(x,t)e^{\phi(r,t,\left\|z-z_j\right\|)}E(t,\left\|z-z_j\right\|)\frac{\left\|z-z_j\right\|+\lambda(\epsilon)\sqrt{t}}{t}$$
As before, taking derivatives of the first term with respect to $\left\|z-z_j\right\|$ gives the term itself times 
$$\frac{C_2}{\lambda(\epsilon)\sqrt{t}}-\frac{1}{2t}\frac{\partial G}{\partial \left\|z-z_j\right\|}$$
while taking derivatives of the second term with respect to $\left\|z-z_j\right\|$ produces the term itself times
$$\frac{\partial \phi}{\partial \left\|z-z_j\right\|}+\frac{C_2}{\lambda(\epsilon)\sqrt{t}}-\frac{1}{2t}\frac{\partial G}{\partial\left\|z-z_j\right\|}+\frac{1}{\left\|z-z_j\right\|+\lambda(\epsilon)\sqrt{t}}$$
when $\lambda(\epsilon)>\sqrt{C_2}+2$, both derivatives are smaller than zero on the entire $[\sqrt{63}(\lambda(\epsilon)\sqrt{t}-r),+\infty).$ As a result, 
$$\left\|\sum_{i\in A_j}w_i p_{i,t}(x)(\nabla\log p_{i,t}(x)-\frac{z-x}{t})-\sum_{i\in A_j}w_i\tilde{p}_{i,t}(x)\frac{\tilde{q}_{i,t}(x)-z}{t}\right\|$$
is upper bounded by $H_1(x,t)$ times
$$\tilde{D}(\lambda(\epsilon))=\frac{4}{3}C_1\cdot \frac{(3\sqrt{n}+1)r}{t}\cdot \exp(9C_2-7\lambda^2(\epsilon))+D(\lambda(\epsilon))\cdot\frac{9\lambda(\epsilon)}{\sqrt{t}}$$
\subsection{Proof Step 5: Multi-Kernel Error Analysis}
Note that $\forall z\in\mathbb{R}^n$, the approximation formula can be equivalently written as
$$\nabla\log p_t(x)\approx\frac{1}{t}\frac{\sum_{i=1}^Pw_i\tilde{p}_{i,t}(x)(\tilde{q}_{i,t}(x)-z)}{\sum_{i=1}^Pw_i\tilde{p}_{i,t}(x)}-\frac{x-z}{t}$$
by choosing $z$ to be the $(C_1,C_2,r)$-EA regular point that satisfies condition (\ref{distancecond}), the total approximation error of our formula can be upper bounded by the sum of
\begin{align}
\label{termone}
\left\|\frac{\sum_{i=1}^{P(r)} w_i\cdot\frac{1}{t}\big[\tilde{p}_{i,t}(x)(\tilde{q}_{i,t}(x)-z)-p_{i,t}(x)(t\nabla\log p_{i,t}(x)-z+x)\big]}{\sum_{i=1}^{P(r)} w_i \tilde{p}_{i,t}(x)}\right\|
\end{align}
and
\begin{align}
\label{termtwo}
\left|\frac{\sum_{i=1}^{P(r)} w_i(p_{i,t}(x)-\tilde{p}_{i,t}(x))}{\sum_{i=1}^{P(r)} w_i p_{i,t}(x)}\right|\cdot\left\|\frac{\sum_{i=1}^{P(r)} w_i p_{i,t}(x)(\nabla\log p_{i,t}(x)-\frac{z-x}{t})}{\sum_{i=1}^{P(r)} w_i \tilde{p}_{i,t}(x)}\right\|.
\end{align}
Here, \ref{termone} is bounded by
$$\frac{\sum_{i\in A_0} w_i\cdot\frac{1}{t}\left\|\tilde{p}_{i,t}(x)(\tilde{q}_{i,t}(x)-z)-p_{i,t}(x)(t\nabla\log p_{i,t}(x)-z+x)\right\|}{\sum_{i\in A_0} w_i \tilde{p}_{i,t}(x)}+\frac{3}{2}C_P(\lambda(\epsilon)\sqrt{t}-r)^{-d}\tilde{D}(\lambda(\epsilon)).$$
Furthermore, the first component above can be decomposed into
\begin{align}
\label{termonesubone}
\frac{\sum_{i\in A_0} w_i\cdot\frac{1}{t}\left\|\tilde{p}_{i,t}(x)\bigl[(\tilde{q}_{i,t}(x)-z)-(t\nabla\log p_{i,t}(x)-z+x)\bigr]\right\|}{\sum_{i\in A_0} w_i \tilde{p}_{i,t}(x)}
\end{align}
plus
\begin{align}
\label{termonesubtwo}
\frac{\sum_{i\in A_0} w_i\left\|(\nabla\log p_{i,t}(x)-\frac{z-x}{t})\bigl(\tilde{p}_{i,t}(x)-p_{i,t}(x)\bigr)\right\|}{\sum_{i\in A_0} w_i \tilde{p}_{i,t}(x)}
\end{align}
This time, recalling that $\left\| y_i-z\right\|\leq 8\lambda(\epsilon)\sqrt{t}-r$ holds for all $i\in A_0$ (hence $\left\|y-z\right\|\leq8\lambda(\epsilon)\sqrt{t}$ holds for all $y\in B(y_i,r),i\in A_0$), and that 
$$x-z=\lambda_{1,z}\vec{v}_{1,z}+\lambda_{2,z}\vec{v}_{2,z}; \sqrt{\lambda_{1,z}^2+\lambda_{2,z}^2}\leq K\lambda(\epsilon)\sqrt{t}-r, \lambda_{2,z}\leq \lambda(\epsilon)\sqrt{t}-r, \vec{v}_{1,z}\perp T_z\mathcal{M},\vec{v}_{2,z}\in T_z\mathcal{M},$$
we know that the decomposition 
$$x-{y}_i=\lambda_{1,i}\vec{v}_{1,i}+\lambda_{2,i}\vec{v}_{2,i}, \vec{v}_{1,i}\perp T_{y_i}\mathcal{M}, \vec{v}_{2,i}\in T_{y_i}\mathcal{M},\left\|\vec{v}_{1,i}\right\|=\left\|\vec{v}_{2,i}\right\|=1$$
satisfies 
$$\lambda_{1,i}\leq (K+8)\lambda(\epsilon)\sqrt{t}-r; \lambda_{2,i}\leq 9\lambda(\epsilon)\sqrt{t}+8k_1K\lambda^2(\epsilon)t-r.$$
Thus, decomposing $\psi(r,t)$ to be the value of $F(r,t,\lambda_{1,i},\lambda_{2,i})$ when $\lambda_{1,i}$ and $\lambda_{2,i}$ take exactly the above values, in which case we clearly have $\psi(r,t)>2r/\sqrt{t},$ the formulas (\ref{valbound}), (\ref{gradbound}), (\ref{noclamp}) ensures that whenever $\psi(r,t)\leq\frac{2}{9}$, (\ref{termonesubone}) is upper bounded by
$$\max_{i\in A_0}\left\|\frac{\tilde{q}_{i,t}(x)-x}{t}-\nabla\log p_{i,t}(x)\right\|\leq\max_{i\in A_0}\frac{1}{t}\left\|\frac{\tilde{R}_{2,i}}{1+\tilde{R}_{1,i}}-\frac{\tilde{R}_{2,i}+\tilde{\Delta}_M}{1+\tilde{R}_{1,i}+\Delta_{M+1}}\right\|,$$
which, noting that $\left\|\tilde{R}_{2,i}+\tilde{\Delta}_M\right\|$ can be rewritten as 
$$\left\|\int_{B(y_i,r)}(y-y_i)e^{\frac{\left\|x-y_i\right\|^2-\left\|x-y\right\|^2}{2t}}dp_i(y)\right\|\leq r\int_{B(y_i,r)}e^{\frac{\left\|x-y_i\right\|^2-\left\|x-y\right\|^2}{2t}}dp_i(y)\leq r\cdot p_{i,t}(x)\cdot e^{\frac{\left\|x-y_i\right\|^2}{2t}},$$
can be further upper bounded via (\ref{simpleboundforp}) by  
$$\frac{1}{t}\max_{i\in A_0} \Big\{e^{\psi(r,t)}r\cdot(\frac{\Delta_{M+1}}{(1+\tilde{R}_{1,i})(1+\tilde{R}_{1,i}+\Delta_{M+1})})+\frac{\tilde{\Delta}_M}{1+\tilde{R}_{1,i}}\Big\},$$
and hence 
$$\frac{9r}{4t}\Big(e^{2\psi(r,t)}\frac{1}{\sqrt{(M+1)!}}\psi(r,t)^{M+1}\Big)+\frac{3r}{2t}\Big(e^{\psi(r,t)}\frac{1}{\sqrt{M!}}\psi(r,t)^M\Big).$$
For (\ref{termonesubtwo}), noting that $\|\nabla\log p_{i,t}(x)-\frac{z-x}{t}\|\leq \frac{8\lambda(\epsilon)}{\sqrt{t}}$, we can upper bound it via
$$\frac{8\lambda(\epsilon)}{\sqrt{t}}\cdot \max_{i\in A_0}\big\{\frac{|\tilde{p}_{i,t}(x)-p_{i,t}(x)|}{\tilde{p}_{i,t}(x)}\big\},$$
and hence 
$$\frac{12\lambda(\epsilon)}{\sqrt{t}}\cdot \max_{i\in A_0}\big\{|\tilde{p}_{i,t}(x)-p_{i,t}(x)|e^{\frac{\left\|x-y_i\right\|^2}{2t}}\big\},$$
and hence, whenever $\psi(r,t)\leq\frac{2}{9}$,
$$\frac{12\lambda(\epsilon)}{\sqrt{t}}\cdot \Big(e^{\psi(r,t)}\frac{1}{\sqrt{(M+1)!}}\psi(r,t)^{M+1}\Big).$$
For (\ref{termtwo}), whenever $\psi(r,t)\leq\frac{2}{9}$, the factor
$$\left|\frac{\sum_{i=1}^{P(r)} w_i(p_{i,t}(x)-\tilde{p}_{i,t}(x))}{\sum_{i=1}^{P(r)} w_i p_{i,t}(x)}\right|$$
can be upper bounded via
$$\left|\frac{\sum_{i\in A_0}w_i(p_{i,t}(x)-\tilde{p}_{i,t}(x))}{\sum_{i\in A_0} w_i p_{i,t}(x)}\right|+C_P(\lambda(\epsilon)\sqrt{t}-r)^{-d}D(\lambda(\epsilon))$$
and hence, noting that (\ref{simpleboundforp}) gives $p_{i,t}(x)\geq e^{-\frac{\left\|x-y_i\right\|^2}{2t}}e^{-\psi(r,t)}$ for all $i\in A_0$, by 
$$e^{2\psi(r,t)}\frac{1}{\sqrt{(M+1)!}}\psi(r,t)^{M+1}+C_P(\lambda(\epsilon)\sqrt{t}-r)^{-d}D(\lambda(\epsilon))$$
while the factor
$$\left\|\frac{\sum_{i=1}^{P(r)} w_i p_{i,t}(x)(\nabla\log p_{i,t}(x)-\frac{z-x}{t})}{\sum_{i=1}^{P(r)} w_i \tilde{p}_{i,t}(x)}\right\|$$
can be upper bounded by the sum of
\begin{align}
\label{termtwosubone}
\left\|\frac{\sum_{i\in A_0} w_i p_{i,t}(x)(\nabla\log p_{i,t}(x)-\frac{z-x}{t})}{\sum_{i\in A_0} w_i \tilde{p}_{i,t}(x)}\right\|
\end{align}
and
\begin{align}
\label{termtwosubtwo}
\left\|\frac{\sum_{j}\sum_{i\in A_j} w_i p_{i,t}(x)(\nabla\log p_{i,t}(x)-\frac{z-x}{t})}{\sum_{i\in \tilde{A}_0} w_i \tilde{p}_{i,t}(x)}\right\|
\end{align}
Here, (\ref{noclamp}) ensures that whenever $\psi (r,t)\leq\frac{2}{9}$, 
$$\max_{i\in A_0}\{\frac{p_{i,t}(x)}{\tilde{p}_{i,t}(x)}\}=\max_{i\in A_0}\frac{1+\tilde{R}_{1,i}+\Delta_{M+1}}{1+\tilde{R}_{1,i}}\leq\max_{i\in A_0}\{1+\frac{3}{2}(\Delta_{M+1})\},$$
bounding (\ref{termtwosubone}) by
$$\frac{8\lambda(\epsilon)}{\sqrt{t}}\cdot \Big(1+\frac{3}{2}\big(e^{\psi(r,t)}\frac{1}{\sqrt{(M+1)!}}\psi(r,t)^{M+1}\big)\Big).$$
Finally, (\ref{termtwosubtwo}) is upper bounded by $C_P(\lambda(\epsilon)\sqrt{t}-r)^{-d}$ times
$$\frac{3}{2}C_1E(t,\left\|z-z_j\right\|)\cdot\frac{\left\|z-z_j\right\|+\lambda(\epsilon)\sqrt{t}}{t},$$
taking derivative of which with respect to $\left\|z-z_j\right\|$ gives the original term times
$$\frac{1}{\left\|z-z_j\right\|+\lambda(\epsilon)\sqrt{t}}+\frac{C_2}{\lambda(\epsilon)\sqrt{t}}-\frac{1}{2t}\frac{\partial G}{\partial\left\|z-z_j\right\|}$$
which, via the previous derivative calculations, also decays uniformly on $[\sqrt{63}(\lambda(\epsilon)\sqrt{t}-r),+\infty)$, bounding (\ref{termtwosubtwo}) by
$$\frac{15\lambda(\epsilon)}{\sqrt{t}}C_1C_P(\lambda(\epsilon)\sqrt{t}-r)^{-d}\cdot e^{9C_2-7\lambda^2(\epsilon)}$$
\subsection{Proof Step 6: Choice for $\lambda(\epsilon)$ and $r$}
Combining the above error analysis, to ensure that the entire approximation pipeline produces total error $\leq \epsilon$ for all $x$ satisfying  (\ref{distancecond}), it suffices to check that our parameter choice satisfies the basic constraints 
$$
t\leq\frac{1}{k_1^2}; K\leq\frac{1}{6k_1\lambda(\epsilon)\sqrt{t}} ; r\leq\min\{\frac{\sqrt{t}}{20},\frac{1}{40}\}; \lambda(\epsilon)>2
$$
and the error analysis constraints (note: we denote $t_c=\min\{t,\frac{1}{4}\}$ for simplicity and omit the $e^{\psi(r,t)}$ or $e^{2\psi(r,t)}$ factors present in several terms, as we have already separately bounded $e^{\psi(r,t)}$ as our very first requirement).
\begin{itemize}
\item $\psi(r,t)\leq\frac{2}{9};e^{\psi(r,t)}\leq \frac{5}{4}$
\item $\frac{3}{2}C_P(\lambda(\epsilon)\sqrt{t}-r)^{-d}\tilde{D}(\lambda(\epsilon))\leq\epsilon/100$
\item $\frac{8\lambda(\epsilon)}{\sqrt{t_c}}C_P(\lambda(\epsilon)\sqrt{t}-r)^{-d}D(\lambda(\epsilon))\leq\epsilon/100$
\item $\frac{15\lambda(\epsilon)}{\sqrt{t_c}}C_1C_P(\lambda(\epsilon)\sqrt{t}-r)^{-d}\cdot e^{9C_2-7\lambda^2(\epsilon)}\leq\epsilon/100$
\item $\frac{12\lambda(\epsilon)}{\sqrt{t_c}}\cdot \frac{1}{\sqrt{(M+1)!}}\psi(r,t)^{M+1}\leq{\epsilon}/{3}$
\item $\frac{r}{t_c}\cdot\frac{1}{\sqrt{M!}}\psi(r,t)^M\leq{\epsilon}/{30}$
\end{itemize}
Reviewing the formulas for $\tilde{D}(\lambda(\epsilon)), D(\lambda(\epsilon))$ while noting that 
$$\lambda(\epsilon)e^{\frac{13\lambda(\epsilon)r}{\sqrt{t}}-7\lambda^2(\epsilon)}\leq e^{\log\lambda(\epsilon)+\lambda(\epsilon)-7\lambda^2(\epsilon)}\leq e^{-7(\lambda(\epsilon)-\frac{1}{7})^2};e^{-7(2-\frac{1}{7})^2}\leq 10^{-7},$$
it thus suffices to have $\psi(r,t)\leq\frac{2}{9}$ and
$$\lambda(\epsilon)\geq \sqrt{\frac{9}{7}C_2+\frac{1}{7}\log \frac{C_1C_p}{\epsilon}+\frac{\log n}{14}+\frac{d+2}{14}\log \frac{1}{{t_c}}}+2;\psi(r,t)^{M+1}\leq\frac{\epsilon}{3}\cdot \frac{\sqrt{t_c}}{12\lambda(\epsilon)}\cdot\sqrt{M!},$$
Finally, noting that $M!\geq M^{M/2}$ and that
$$\psi(r,t)=\frac{\frac{1}{2}k_1r^2((K+8)\lambda(\epsilon)\sqrt{t})+r(9\lambda(\epsilon)\sqrt{t}+8k_1K\lambda(\epsilon)^2t)}{t}\leq\frac{11\lambda(\epsilon)r}{\sqrt{t}},$$
it suffices to set 
$$r\leq\min\{\frac{\sqrt{t_c}}{50\lambda(\epsilon)},\frac{\sqrt{t_c}}{11\lambda(\epsilon)}\cdot (\frac{\epsilon\sqrt{t_c}}{36\lambda(\epsilon)})^{\frac{1}{M+1}}\cdot{M}^{\frac{M}{4(M+1)}}\}$$
\section{Proof of Theorem 3}
\begin{theorem3}
(1) Let $\mu$ be an arbitrary probability distribution in $\mathbb{R}^n$ with compact support. For any fixed radius $r$, define $U_\delta$ to be the set of points that are $(e^{3n}\frac{1}{\delta},\frac{n}{2},r)$-EA regular under $\mu$, then
$\mu(U_\delta)>1-\delta.$

(2) Let $\mu$ be a distribution concentrated on a $C^2$, $d$-dimensional manifold $\mathcal{M}\subseteq \mathbb{R}^n$. Assume that for all $r>0$, $\mathcal{M}$ admits a covering $\{B(x_i,r)\}_{i=1}^N$ such that $\forall y_i\in\mathcal{M}$, $B(y_i,\rho)$ contains the centers of at most $C_\mathcal{M}(1+\rho/r)^d$ balls inside $\{B(x_i,r)\}_{i=1}^N$. For any fixed radius $r$, let $U_\delta$ be the set of points that are $(C_{\mathcal{M}}e^{2.5d}\frac{6}{\delta},\frac{1}{2}d,r)$-EA regular under $\mu$, then $\mu(U_\delta)>1-\delta$.
\end{theorem3}
\subsection{Proof of Part 1 (General Case)}
Given a distribution $\mu$ and constants $C_1,C_2,r$, we define a ball $B(x,r)$ to be "bad" if there exists a ball $B(y,r)$ such that
$$C_1e^{C_2\frac{\left\|x-y\right\|_2}{r}}\cdot \mu( B(x,r))<\mu(B(y,r)).$$
Via the axiom of choice, there exists a mapping $g: \{x|B(x,r) \ \mathrm{bad}\}\to\mathbb{R}^n$ taking the center of each "bad" ball to a $y$ satisfying the above criterion. What's more, as $\mu$ has compact support, $\mathrm{Im}(g)\subseteq\{y|\mu(B(y,r))>0\}$ is also a bounded set.

We construct $\tilde{g}: \{x|B(x,r) \ \mathrm{bad}\}\to\mathbb{R}^n$ s.t. $d(y_1,y_2)>2r, \forall y \in \mathrm{Im}(\tilde{g})$, and
$$C_1\cdot(1-\tau)\cdot e^{C_2(\frac{\left\|x-y\right\|_2}{r}-2)}\cdot \mu( B(x,r))< \mu(B(\tilde{g}(x),r)), \forall B(x,r) \ \mathrm{bad},$$
with $\tau$ an arbitrarily small constant, in the following manner. First, pick an arbitrary $y_1\in \mathrm{Im}(g)$ such that 
$$\mu(B(y_1,r))>(1-\tau)\sup_{\mathrm{Im(g)}}(\mu(B(y,r))),$$
and define 
$$g_1(x)=g(x),\forall \left\|g(x)-y_1\right\|>2r;g_1(x)=y_1,\forall \left\|g_1(x)-y_1\right\|\leq 2r.$$
Now suppose we have defined $g_1,...,g_j$ and $y_1,...,y_j$. We then pick an arbitrary $y_{j+1}\in K_j:= \mathrm{Im}(g_j)\backslash\{y_1,...,y_j\}$ such that 
$$\mu(B(y_{j+1},r))>(1-\tau)\sup_{K_j}(\mu(B(y,r))),$$
and define 
$$g_{j+1}(x)=g_j(x),\forall\left\|g_j(x)-y_{j+1}\right\|>2r;g_{j+1}(x)=y_{j+1},\forall \left\|g_j(x)-y_{j+1}\right\|<2r.$$
As $\mathrm{Im}(g)$ is bounded, our construction guarantees that after a finite number of iterations, $\mathrm{Im}(g_j)$ becomes a discrete set such that the distance between any two points is greater than $2r$. Take $\tilde{g}$ to be the endpoint of this finite iteration. As the image of each point is relocated at most once throughout the entire sequence, we have 
$$\mu(B(\tilde{g}(x),r))>(1-\tau)\mu(B(g(x),r));\left\|\tilde{g}(x)-x\right\|\leq \left\|g(x)-x\right\|+2r,$$
hence
$$C_1\cdot(1-\tau)\cdot e^{C_2(\frac{\left\|x-y\right\|_2}{r}-2)}\cdot \mu( B(x,r))< \mu(B(\tilde{g}(x),r)), \forall B(x,r) \ \mathrm{bad}.$$
Denote $\mathrm{Im}(\tilde{g}):=\{y_1,...,y_N\},$ where 
$$\sum_{i=1}^N\mu(B(y_i,r))<1$$
as the $N$ balls do not overlap. We now consider the decomposition of $\mathbb{R}^n$ into countable cubes of edge length $\frac{2r}{\sqrt{n}}$, i.e. $\{\prod_{i=1}^n[\frac{2j_ir}{\sqrt{n}},\frac{2(j_i+1)r}{\sqrt{n}})|j_1,...,j_n\in\mathbb{Z}\}$.

If a cube $V_j$ centered at $z_j$ contains a point that is not $(C_1,C_2,r)$-EA regular, we know from definition that $B(z_j,r)$ must be a "bad" ball, and hence 
$$\mu(V_j)\leq\mu(B(z_j,r))\leq\max_{i\leq N}\{\frac{\mu(B(y_i,r))}{C_1\cdot(1-\tau)\cdot e^{C_2(\frac{\left\|z_j-y_i\right\|}{r}-2)}}\}\leq\sum_{i=1}^N\frac{\mu(B(y_i,r))}{C_1\cdot(1-\tau)\cdot e^{C_2(\frac{\left\|z_j-y_i\right\|}{r}-2)}}$$
Thus the total measure of cubes that contain non-$(C_1,C_2,r)$-EA regular points must be no greater than 
$$\sum_{i=1}^N\sum_{j\in \mathbb{Z}^n}\frac{\mu(B(y_i,r))}{C_1\cdot(1-\tau)\cdot e^{C_2(\frac{\left\|z_j-y_i\right\|}{r}-2)}},$$
which, via translation invariance with respect to integral grids, can be upper bounded by
$$\max_{\left\|\vec{v}\right\|_\infty<\frac{1}{2}}\left[\sum_{j\in \mathbb{Z}^n}\frac{1}{C_1\cdot(1-\tau)\cdot e^{C_2(\frac{2\left\|j+\vec{v}\right\|}{\sqrt{n}}-2)}}\right].$$
As $\sqrt{n}\left\|x\right\|_2\geq\left\|x\right\|_1$ holds for all $ x\in\mathbb{R}^n$, this can be upper bounded by
$$\frac{e^{2C_2}}{C_1(1-\tau)}\prod_{l=1}^n\left[\max_{|v|<\frac{1}{2}}\sum_{j\in \mathbb{Z}}e^{-C_2(\frac{2|j+v|}{n})}\right]$$
and hence
$$\frac{e^{2C_2}}{C_1(1-\tau)}\left[\frac{2}{1-e^{-\frac{2C_2}{n}}}\right]^n.$$
Plugging $C_2=\frac{n}{2}$ and $C_1=e^{3n}\cdot \frac{1}{\delta}$ into the formula and letting $\tau\to 0$ immediately verifies our claim.
\subsection{Proof of part 2 (Manifold Case)}
Given $C_1, C_2, r$, we still define a ball $B(x,r)$ as "bad" if there exists a witness ball $B(y,r)$ such that
$$\mu(B(x,r))\leq \frac{\mu(B(y,r))}{C_1\cdot e^{C_2(\frac{\left\|x-y\right\|}{r})}},$$
under this definition, a point $x$ is not $(C_1,C_2,r)$-EA regular if and only if any ball containing it is a bad ball. Using the previous proof techniques, $\forall \tau>0$, we can construct $P_0$ non-overlapping balls $\{B(y_j,r)\}_{j=1}^{P_0}$ such that for each ``bad" ball $B(x,r)$,
$$\mu(B(x,r))\leq \frac{\mu(B(y_i,r))}{C_1\cdot(1-\tau)\cdot e^{C_2(\frac{\left\|x-y\right\|}{r}-2)}}$$
for some $i\leq P_0$, where we choose $\tau=e^{-C_2}$ for convenience of canceling terms.

Construct a covering $\{B(x_i,r)\}_{i=1}^N$ of $\mathcal{M}$ satisfying the covering condition stated in part (2) of the present theorem. The total measure of non $(C_1, C_2,r)$-EA regular points is upper bounded by the total measure of bad balls inside $\{B(x_i,r)\}_{i=1}^N$, which is in turn upper bounded by
$$\sum_{i=1}^N \max_{j\leq P_0}\frac{\mu(B(y_j,r))}{C_1\cdot(1-\tau)\cdot e^{C_2(\frac{\left\|x_i-y_j\right\|}{r}-2)}},$$
and hence
$$\sum_{j=1}^{P_0}\Big(\sum_{i=1}^N\frac{\mu(B(y_j,r))}{C_1\cdot(1-\tau)\cdot e^{C_2(\frac{\left\|x_i-y_j\right\|}{r}-2)}}\Big).$$
Noting that for all $x$ such that $\left\|x-y_j\right\|\in[mr,mr+r),$
$$\frac{\mu(B(y_j,r))}{C_1\cdot(1-\tau)\cdot e^{C_2(\frac{\left\|x-y_j\right\|}{r}-2)}}\leq\frac{e^{2C_2}\mu(B(y_j,r))}{C_1(1-\tau)\cdot e^{mC_2}}=\sum_{k=m}^\infty\frac{e^{2C_2}\mu(B(y_j,r))}{C_1(1-\tau)\cdot e^{kC_2}}\cdot(1-e^{-C_2}),$$
denoting $A_k=\{i\leq N|\left\|x_i-y_j\right\|\leq (k+1)r\},$
$$\sum_{i=1}^N\frac{\mu(B(y_j,r))}{C_1\cdot(1-\tau)\cdot e^{C_2(\frac{\left\|x_i-y_j\right\|}{r}-2)}}$$
can be upper bounded by
$$\sum_{k=0}^\infty\Big(|A_k|\cdot(1-e^{-C_2})\cdot \frac{e^{2C_2}\mu(B(y_j,r))}{C_1\cdot(1-\tau)\cdot e^{k\cdot C_2}}\Big)$$
considering that $\mathrm{supp}(\mu)\subseteq\mathcal{M}$, we can assume without loss of generality that $\forall y_i, \left\|\tilde{y}_i-y_i\right\|\leq r$ for some $\tilde{y}_i\in \mathcal{M}$, hence $B(y_j,(k+1)r)\subseteq B(\tilde{y}_j,(k+2)r)$, meaning that $|A_k|\leq C_\mathcal{M}(k+3)^d$. Therefore, the above expression is upper bounded by
$$C_\mathcal{M}\mu(B(y_j,r))\sum_{k=0}^\infty(k+3)^d\frac{1}{C_1e^{(k-2)C_2}}$$
as $\frac{(k+4)^d}{(k+3)^d}$ is upper bounded by $(4/3)^d$, hence $(k+3)^d\leq3^d\cdot(\frac{4}{3})^{kd}$, this term is upper bounded by 
$$\frac{C_\mathcal{M}e^{2C_2}}{C_1}\mu(B(y_j,r))\cdot3^d\cdot\Big(\sum_{k=0}^\infty e^{k\cdot(d\log\frac{4}{3}-C_2)}\Big)$$

Therefore, noting that $\frac{1}{2}-\log \frac{4}{3}>\frac{1}{6}$ and that $\sum_{j=1}^{P_0}\mu(B(y_j,r))\leq1$, taking $C_2=\frac{d}{2}$ and taking $C_1=C_{\mathcal{M}}e^{2.5d}\frac{6}{\delta}$ bounds the total measure of non $(C_1,C_2,r)$-EA regular points by $\delta$.
\section{Proof of Lemma 1 and Corollary 1}
\begin{lemma1}
Let $h(x,t)$ be a function \footnote{can be vector or tensor-valued} with $\|h(x,t)\|\leq (\max\{1,{\|x\|}/{\sqrt{t}}\})^M$ for some $M\in \mathbb{N}$, and define the operator $H_h$ such that
$$(H_h\circ \mu)(x,t)=\frac{\int_{\mathbb{R}^n}e^{-\frac{\left\|x-y\right\|^2}{2t}}h(y-x,t)d\mu}{\int_{\mathbb{R}^n}e^{-\frac{\left\|x-y\right\|^2}{2t}}d\mu}$$
for each probability measure $\mu$. Then, for each $x$ at most distance $\lambda\sqrt{t}$ from a $(e^{3n}\frac{1}{\delta}, \frac{n}{2}, \lambda\sqrt{t})$-EA regular point $z$, $\lambda\geq \max\{\sqrt{n},M,\sqrt{\log (1/\delta)}\}$, 
$$\|(H_h\circ\mu)(x,t)\|\leq(15\lambda)^M+1.$$
\end{lemma1}
\begin{proof}
With a trivial change of coordinates, we assume that $x=0$. Decomposing $\mathbb{R}^n$ into countable cubes of edge length $\frac{2\lambda\sqrt{t}}{\sqrt{n}}$, 
$$\{V_{j}=\prod_{i=1}^n[\frac{(2j_i-1)\lambda\sqrt{t}}{\sqrt{n}},\frac{(2j_i+1)\lambda\sqrt{t}}{\sqrt{n}})|j=(j_1,...,j_n)\in\mathbb{Z}^n\},$$
the measure of each cube is clearly upper bounded by that of a closed ball with radius $\lambda\sqrt{t}$, centered at 
$$\frac{\lambda\sqrt{t}}{\sqrt{n}}\cdot(2j_1,2j_2,...,2j_n).$$
We now define the set 
$$\mathcal{J}:=\{(j_1,j_2,...,j_n)|\sum_{i=1}^n(j_i)^2\leq 49n\}.$$
Clearly, denoting $z_0$ to be the ball such that
$$z\in B(z_0,\lambda\sqrt{t}); \frac{1}{\delta}e^{\frac{n}{2}(\frac{\|y-z_0\|}{\lambda\sqrt{t}}+6)}\mu(B(z_0,\lambda\sqrt{t}))\geq\mu (B(y,\lambda\sqrt{t})), \forall y\in\mathbb{R}^n,$$
$\|(H_h\circ \mu)(x,t)\|$ can be upper bounded by
$$\frac{\sum_{j\in \mathcal{J}}\int_{V_j}e^{-\frac{\left\|y\right\|^2}{2t}}\|h(\frac{y}{\sqrt{t}})\|d\mu}{\sum_{j\in \mathcal{J}}\int_{V_j}e^{-\frac{\left\|y\right\|^2}{2t}}d\mu}+\frac{\sum_{j\in \mathcal{J}^c}\int_{V_j}e^{-\frac{\left\|y\right\|^2}{2t}}\|h(\frac{y}{\sqrt{t}})\|d\mu}{\int_{B(z_0,\lambda\sqrt{t})}e^{-\frac{\left\|y\right\|^2}{2t}}d\mu}$$
where the first term is upper bounded by $$\sup\big\{\|h(\frac{y}{\sqrt{t}})\||y\in \bigcup_{j\in\mathcal{J}}V_j\big\}\leq(15\lambda)^M.$$
For the second term, we begin by analyzing a single $j\in\mathcal{J}^c$, where we note that
$$\frac{\int_{V_j}e^{-\frac{\left\|y\right\|^2}{2t}}\|h(\frac{y}{\sqrt{t}})\|d\mu}{\int_{B(z_0,\lambda\sqrt{t})}e^{-\frac{\left\|y\right\|^2}{2t}}d\mu}\leq\frac{\int_{V_j}d\mu}{\int_{B(z_0,\lambda\sqrt{t})}d\mu}\cdot e^{\frac{(\|z_0\|+\lambda\sqrt{t})^2}{2t}-\frac{\lambda^2(2\|j\|-\sqrt{n})^2}{2n}}\cdot(\frac{2\|j\|}{\sqrt{n}}+1)^M\lambda^M$$
which, further noting that
$$\frac{\int_{V_j}d\mu}{\int_{B(z_0,\lambda\sqrt{t})}d\mu}\leq\frac{1}{\delta}\cdot e^{3n}\cdot e^{\frac{n}{2\lambda\sqrt{t}}(2\lambda\sqrt{t}+\frac{2\lambda\sqrt{t}}{\sqrt{n}}\|j\|)}\leq\frac{1}{\delta}e^{4n}\cdot e^{\sqrt{n}\|j\|}$$
is upper bounded by
$$\frac{1}{\delta}e^{4n}e^{\sqrt{n}\|j\|}e^{\frac{9}{2}\lambda^2-\frac{\lambda^2(2\|j\|-\sqrt{n})^2}{2n}}\cdot (\frac{2\|j\|}{\sqrt{n}}+1)^M\lambda^M.$$
Denoting $\tilde{V}_j$ to be the cube $\prod_{i=1}^n[2j_i-1,2j_i+1]$, whenever $\|j\|\geq2\sqrt{n}$, the above expression is further upper bounded by 
$$\int_{\tilde{V}_j}\frac{1}{2^n\delta}e^{4n}e^{\frac{\sqrt{n}(\|x\|+\sqrt{n})}{2}}e^{\frac{9}{2}\lambda^2-\frac{\lambda^2(\|x\|-2\sqrt{n})^2}{2n}}\cdot (\frac{\|x\|}{\sqrt{n}}+2)^M\lambda^Mdx,$$
$dx$ denoting the normal $n$-dimensional Lebesgue measure. As a result, noting that
$$\bigcup_{j\in\mathcal{J}^c}\tilde{V}_j\subseteq \mathbb{R}^n\backslash B\big(0,12\sqrt{n}\big),$$
the second term is upper bounded by
$$\frac{\lambda^M}{2^n\delta}e^{\frac{9}{2}(\lambda^2+n)}\int_{\|x\|\geq12\sqrt{n}}e^{\frac{\sqrt{n}\|x\|}{2}}e^{-\frac{\lambda^2(\|x\|-2\sqrt{n})^2}{2n}}\cdot (\frac{\|x\|}{\sqrt{n}}+2)^Mdx,$$
and hence, noting that
$$\frac{\lambda^2(\|x\|-2\sqrt{n})^2}{2n}-\frac{\sqrt{n}\|x\|}{2}\geq\frac{\lambda^2(\|x\|-3\sqrt{n})^2}{2n}\geq\frac{\lambda^2\|\frac{3x}{4}\|^2}{2n}, \forall \lambda\geq\sqrt{n},\|x\|\geq12\sqrt{n}$$
is upper bounded by 
$$\frac{\lambda^M}{2^n\delta}e^{\frac{9}{2}(\lambda^2+n)}\int_{\|x\|\geq12\sqrt{n}}e^{-\frac{\lambda^2\|\frac{3x}{4}\|^2}{2n}}\cdot (\frac{3\|x\|}{2\sqrt{n}})^Mdx=\frac{2^{M}}{\delta}\cdot\left(\frac{2}{3}\right)^ne^{\frac{9}{2}(\lambda^2+n)}({\sqrt{n}}/{\lambda})^{n}\int_{\|u\|\geq9\lambda}e^{-\frac{\|u\|^2}{2}}\|u\|^Mdu.$$
As a result, noting that $e^{\frac{\|u\|^2}{8}}\geq \|u\|^M,\forall \|u\|\geq8M$, and hence
$$\int_{\|u\|\geq9\lambda}e^{-\frac{\|u\|^2}{2}}\|u\|^Mdu\leq\int_{\|u\|\geq9\lambda}e^{-\frac{3\|u\|^2}{8}}du\leq\int_{\|u\|\geq9\lambda}e^{-\frac{49\|u\|^2}{162}}du=(9/7)^n\int_{\|v\|\geq7\lambda}e^{-\frac{\|v\|^2}{2}}dv,$$
the final term being upper bounded by $e^{-18\lambda^2}\cdot\left(\frac{9\sqrt{2\pi}}{7}\right)^n\leq(3.6)^ne^{-18\lambda^2}$ via the Gaussian concentration inequality since $7\lambda\geq\sqrt{n}+\sqrt{2\cdot(18\lambda^2)}$, we have
$$\frac{\sum_{j\in \mathcal{J^c}}\int_{V_j}e^{-\frac{\left\|y\right\|^2}{2t}}\|h(\frac{y}{\sqrt{t}})\|dp(y)}{\int_{B(z_0,\lambda\sqrt{t})}e^{-\frac{\left\|y\right\|^2}{2t}}dp(y)}\leq(2.4)^ne^{-18\lambda^2}\cdot\frac{2^{M}}{\delta}e^{\frac{9}{2}(\lambda^2+n)}({\sqrt{n}}/{\lambda})^{n}\leq\frac{2^{M}\cdot(2.4)^n}{\delta}e^{-9\lambda^2}\leq1,$$
and summing the terms thus finishes the proof.
\end{proof}
\begin{corollary1}
Let $\{f_i\}_{i=1}^S$ \footnote{can be vector or tensor valued} satisfy $\|f_i(x,t)\|\leq C_i\max\{1,\|x\|/\sqrt{t}\}^{M_i}$, $\forall i\leq S$, and let multiplication in $\mathrm{Im}(f_i)$ satisfy $\|f(x)\cdot g(y)\|\leq\|f(x)\|\|g(y)\|$. Under the VE scheme, for any compact distribution $p_0$ with probability measure $\mu$,  $\mathrm{E}_{p_t(x)}\|\prod_{i=1}^S(H_{f_i}\circ\mu)(x,t)\|^j$ is bounded by a constant depending on $j, n,$ and $\{f_i\}$, but independent of $p_0$ or $t$.
\end{corollary1}
\begin{proof}
Set $\lambda=\max\{\sqrt{n}+\tau\cdot\sqrt{2},M\}; \delta=e^{-\tau^2}$. We note that $\lambda\geq1$, for any $x$ at most distance $\lambda\sqrt{t}$ away from a $(e^{3n}\frac{1}{\delta},\frac{n}{2},\lambda\sqrt{t})$-EA regular point $z$, $$\|\prod_{i=1}^S(H_{f_i}\circ\mu)(x,t)\|\leq (\prod_{i=1}^S C_i)\cdot(16\lambda)^{\sum_{i=1}^SM_i}$$
Meanwhile, using the same argument from Theorem 1, $x\sim p_t(x)$ is one such point with probability at least $1-2e^{-\tau^2}$. That is, probability of the event
$$\|\prod_{i=1}^S(H_{f_i}\circ\mu)(x,t)\|\geq(\prod_{i=1}^S C_i)\cdot(16(\sqrt{n}+\tau\cdot\sqrt{2}))^{\sum_{i=1}^SM_i}$$
is upper bounded by $2e^{-\tau^2}.$
Therefore, the statement is proven using the formula 
$$\int_{\Omega}\|f\|^jd\mu=\int_0^\infty ju^{j-1}\mu\big(\{x|\|f(x)\|\geq u\}\big)du$$
and the fact that super-exponential decay overwhelms polynomial growth. 
\end{proof}
\section{Proof of Theorem 4}
\begin{theorem4}
Defining 
$$R=\max_{x,y\in\mathrm{supp}(p_0)}\left\|x-y\right\|, V_J=\{z|\inf_{x\in\mathrm{supp}(p_0)}\left\|x-j\right\|_\infty\leq J\sqrt{T}\},$$
under the assumption that $R>\max\{3,3\sqrt{n}r\}$, $n\geq 25$, $r\leq \min\{\sqrt{t_0},\frac{1}{25}\}$, $M\leq \sqrt{n}$, $M\leq\frac{1}{r}$, the expression (equivalent to the Softmax expression \ref{discreteapprox})
$$\frac{\sum_{i=1}^Pw_i\tilde{p}_{i,t}(x)\tilde{q}_{i,t}(x)}{\sum_{i=1}^Pw_i\tilde{p}_{i,t}(x)},$$
with $\tilde{p}_{i,t}$ and $\tilde{q}_{i,t}$ defined in (\ref{Thm:Discretizationerror}), can be approximated with error $\leq t_0\epsilon\leq\frac{1}{4}$ for all $(t,x)\in[t_0,T]\times V_J$, using a ReLU network whose output remain strictly within $\mathrm{Conv(\mathrm{supp}(p_0))}+B(0,t_0\epsilon+3\sqrt{n}r)$ across the entire $\mathbb{R}^n$. The depth and width of the network are bounded by 
$$D\leq\mathcal{O}\left(M\log \frac{M}{\epsilon}+M^2\log n\omega+\log \frac{\omega P}{\epsilon}\log \log \frac{\omega P}{\epsilon}\right)$$ 
and
$$W\leq \mathcal{O}\left(P\left(\frac{(n+M)^M}{(M-1)!}\log\frac{n\omega M}{\epsilon }+\log^2\frac{\omega P}{\epsilon}\right)\right),$$
respectively, where we define
$$\omega=\frac{(R+J\sqrt{T})}{t_0}.$$
\end{theorem4}
\subsection{Proof Step 1: Outline of Approximation Method}
we rewrite the discretization formula above as
\begin{align}
\label{expression}
\Big\langle(\tilde{q}_{1,t}-y_0,...,\tilde{q}_{P,t}-y_0),\mathrm{Softmax}\Big(\log w_1+\log \tilde{p}_{1,t},...,\log w_P+\log \tilde{p}_{P,t}\Big)\Big\rangle,
\end{align}
with $y_0$ an arbitrary point inside $\mathrm{supp}(p_0)$, introduced to upper bound $\left\|\tilde{q}_{i,t}-y_0\right\|$ by $R+3\sqrt{n}r$.

Noting that for all $x_1,...,x_P,$
$$\left\|\mathrm{Softmax}(x_1+\Delta_1,x_2+\Delta_2,...,x_P+\Delta_P)-\mathrm{Softmax}(x_1,...,x_P)\right\|_1$$
is upper bounded by 
$$\sum_{i=1}^P\left|\frac{e^{x_i+\Delta_i}-e^{x_i}}{\sum_{j=1}^Pe^{x_j}}\right|+\sum_{i=1}^P\left|\frac{e^{x_i+\Delta_i}}{\sum_{j=1}^Pe^{x_j}}-\frac{e^{x_i+\Delta_i}}{\sum_{j=1}^Pe^{x_j+\Delta_j}}\right|$$
and hence, noting that $e^{r}-1>1-e^{-r}$ for all $r>0$, is further upper bounded by
$$\sum_{i=1}^P\left|\frac{e^{x_i+|\Delta_i|}-e^{x_i}}{\sum_{j=1}^Pe^{x_j}}\right|+\frac{(\sum_{i=1}^Pe^{x_i+\Delta_i})\sum_{j=1}^P|e^{x_j+|\Delta_j|}-e^{x_j}|}{(\sum_{j=1}^Pe^{x_j})(\sum_{j=1}^Pe^{x_j+\Delta_j})}\leq2\big(\exp\small(\max_{i\leq P}|\Delta_i|\small)-1\big)$$
to accurately approximate the above expression with $l^2$ error no greater than $t_0\epsilon$ for every possible $x\in V_J$, it suffices to:
\begin{enumerate}
\item Calculate $\{\tilde{q}_{i,t}(x)\}_{i=1}^P$, each with $l^2$ error $\leq \frac{t_0\epsilon}{12}$
\item Calculate $\{\log \tilde{p}_{i,t}\}_{i=1}^P$, each with error $\leq \frac{t_0\epsilon}{3(R+3\sqrt{n}r)}$ 
\item Calculate the Softmax function with $l^1$ error $\leq \frac{t_0\epsilon}{20(R+3\sqrt{n}r)}$

Here, as $\frac{t_0\epsilon}{3(R+3\sqrt{n}r)}\leq1/30$, and $(e^x-1)/x\leq 21/20$ holds for all $x\leq1/30$, error caused by inaccurate input of Softmax is upper bounded by $\frac{2t_0\epsilon}{3(R+3\sqrt{n}r)}\cdot\frac{21}{20}\leq\frac{7t_0\epsilon}{10(R+3\sqrt{n}r)}$ due to the previous error analysis, so the $l^1$ difference between the real and approximated outputs is at most $\frac{3t_0\epsilon}{4(R+3\sqrt{n}r)}.$

\item Calculate the final weighted average ($P\cdot n$ multiplications), with total error $\leq \frac{t_0\epsilon}{12}$.

Here, as $\mathrm{Softmax}$ always outputs a vector with $l^1$ norm 1, and $\left\|\tilde{q}_{i,t}(x)-y_0\right\|_2\leq (R+3\sqrt{n}r)$ due to the positionwise clamping on $\tilde{q}$, the total $l^2$ error caused by inaccurate inputs is upper bounded by
$$(R+3\sqrt{n}r)\cdot\frac{3t_0\epsilon}{4(R+3\sqrt{n}r)}+\frac{t_0\epsilon}{12}\cdot 1+\frac{3t_0\epsilon}{4(R+3\sqrt{n}r)}\cdot \frac{t_0\epsilon}{12}\leq \frac{11t_0\epsilon}{12},$$
thus bounding the total approximation error by $t_0\epsilon$.
\end{enumerate}
\subsection{Proof Step 2: Approximating Softmax}
For Softmax approximation, we provide an independently interesting lemma.
\begin{lemma}
The softmax function with $P>20$ variables, $\mathrm{Softmax}(x_1,...,x_P)$, can be approximated with $l^1$ error $\leq \delta$ on the entire $\mathbb{R}^P$, using a $\mathrm{ReLU}$ neural network of depth $D$ and width $W$ that consistently produces non-negative outputs. Here,
$$W\sim P\cdot\mathcal{O}\Big(\log ^2\frac{P}{\delta}\Big) ; D\sim \mathcal{O}(\log \frac{P}{\delta}\log \log \frac{P}{\delta})$$
\end{lemma}
\begin{proof}
Noting that 
$$\mathrm{Softmax}(x_1,...x_P)=\mathrm{Softmax}(x_1-\lambda,...,x_P-\lambda), \forall \lambda \in \mathbb{R},$$
The Softmax function can be estimated with error $\leq \delta$ via the following procedure.
\begin{enumerate}
\item Calculate $x_{max}=\max\{x_1,...,x_P\}$ with zero error using a network of width $\mathcal{O}(P)$ and depth $\mathcal{O}(\log_2P)$ by iteratively calculating 
$$\max\{x_1,x_2\}=x_1+\mathrm{ReLU}(x_2-x_1)$$ 
\item Deduct $x_{max}$ from each of the $P$ entries via translation invariance, truncating the input range to $(-\infty,0]^P$ with at least 1 input value exactly equaling 0.  
\item Construct a network $\varphi(x)$  that calculates $e^{x}$ on the interval $[\log\frac{\delta}{4P^2},0]$ with error no more than $\frac{\delta}{4P^2}$. Calling $$\varphi\Big(\mathrm{ReLU}\big(x-\log\frac{\delta}{4P^2}\big)+\log\frac{\delta}{4P^2}\Big)$$
$P$ times calculates $(e^{x_1-x_{max}},...,e^{x_P-x_{max}})$ on $(-\infty,0]^P$ with $L^\infty$ error no more than $\frac{\delta}{4P^2}$.
\item Calculate $f(\sum_i(e^{x_i-x_{max}}))$, where $f(t)=\frac{1}{t}$, on the domain $[1,P]$ once, with error no more than $\frac{\delta}{3P}$. 

Here, the input value $\sum_i(e^{x_i-x_{max}})$ is at most $\frac{\delta}{4P}$ away from its true value, hence whenever $\frac{\delta}{4P}\leq\frac{1}{10}$, the total error of calculating $f(\sum_i(e^{x_i-x_{max}}))$ is at most $\frac{\delta}{4P}\cdot f'(1-\frac{1}{10})+\frac{\delta}{3P}\leq\frac{2\delta}{3P}$.
\item Calculate $h(x,y)=xy$ ($x=e^{x_i-x_{max}},y=f(\sum_i(e^{x_i-x_{max}}))$) on the interval $[0,1.1]^2$ (the extra 0.1 constituting ``elbow room" for error terms) $P$ times, each with error no more than $\frac{\delta}{6P}$.

Here, maximum approximation error for each $(x,y)$ pair is $\frac{\delta}{4P^2}*1.1+\frac{2\delta}{3P}*1.1+\frac{\delta}{4P^2}\frac{2\delta}{3P}+\frac{\delta}{6P}\leq\frac{\delta}{P}$, hence the summed $l^1$ approximation error is upper bounded by $\delta$.
\item Applying a final $\mathrm{ReLU}$ clamping to ensure output weights are non-negative, which never increases the $l^1$ error as the real weights are always positive.
\end{enumerate}
Citing standard results, approximating $g(x)=x^2$ on $[0,U]$  with error $\delta$ (\citep{squarerelu}) can be done with a network of width $\mathcal{O}(\log(\frac{U}{\delta}))$ and depth $\mathcal{O}(\log(\frac{U}{\delta}))$; while approximating $e^{x}$ on $[-U,0]$ with error $\delta$ and $\frac{1}{t}$ on $[1,U]$ with error $\delta$ (\citep{exponentialrelu}, Theorem 3.7) can both be done with a network of width $\mathcal{O}(\log^2(\frac{U}{\delta}))$ and depth $\mathcal{O}(\log(\frac{U}{\delta})\log(\log\frac{U}{\delta}))$ (with different constant terms hidden by the big $\mathcal{O}$-notation). Therefore, noting that $$4xy=(x+y)^2-(x-y)^2; \lim_{x\to\infty}\frac{\log (x\log x)}{\log x}=1,$$
the above steps can be performed via a neural network with total width and depth bounded by
$$W\sim P\cdot\mathcal{O}\Big(\log ^2\frac{P}{\delta}\Big) ; D\sim \mathcal{O}(\log \frac{P}{\delta}\log \log \frac{P}{\delta})$$
finishing the proof.
\end{proof}
\subsection{Proof Step 3: Approximating $\log \tilde{p}_{i,t}$ and $\tilde{q}_{i,t}$}
The estimation of $\log \tilde{p}_{i,t}(x)$ and $\tilde{q}_{i,t}(x)$ mainly boils down to estimating $\tilde{R}_{1,i}$, $\tilde{R}_{2,i}$, and $-\frac{\left\|x-y_i\right\|^2}{2t}$, for this purpose, we propose calculating the exact value of 
$$\big\{\frac{1}{j!}\int_{B(y_i,r)}(y-y_i)^{\otimes j}d p_i(y)\big\}_{j=1}^M$$
in advance and baking them into the corresponding weights of a linear layer, in which case $\tilde{p}_{i,t}(x)$ and $\tilde{q}_{i,t}(x)$ can be approximated via the following procedure.
\begin{enumerate}
\item Calculate $\{t^{-M'/2}\}_{M'=1}^{M}$ using a ReLU network for future use.
\item Calculate $\frac{\left\|x-y_i\right\|^2}{t}$ and the terms 
$$\bigcup_{M'\leq M}\{\prod_{j=1}^{M'}p_{a_j}(\frac{x-y_i}{\sqrt{t}})|(a_1,...,a_j)\in S_{n,j}\}$$
using a ReLU network, where $p_{a_j}$ denotes projection onto a vector's $a_j$th component and $S_{n,j}$ denotes the set of ordered $j$-tuples $(a_1,...,a_j)$ where $a_k\in\{1,2,...,n\},\forall k.$
\item Propagate the calculated terms through a single linear layer to construct the tensors $\{H_{M'}(\frac{x-y_i}{\sqrt{t}})\}_{M'=1}^M$
\item Calculate
$$H_{M'}(\frac{x-y_i}{\sqrt{t}}):\int_{B(y_i,r)}(y-y_i)^{\otimes {M'}}dp_i(y); H_{M'-1}(\frac{x-y_i}{\sqrt{t}}): \int_{B(y_i,r)}(y-y_i)^{\otimes ({M'}+1)}dp_i(y)$$
using the linear layer storing $\big\{\int_{B(y_i,r)}(y-y_i)^{\otimes M'}d p_i(y)\big\}_{M'=1}^M$.
\item Multiply results by the corresponding $\{t^{-j/2}\}$ via ReLU approximated multiplication, then sum terms via a linear layer to get $\tilde{R}_{1,i}$, $\tilde{R}_{2,i}$.
\item Use ReLU networks to approximate $\log (1+\mathrm{Clamp}_{\frac{1}{3}}(\tilde{R}_{1,i}))$ and $\frac{\mathrm{Clamp}_{2r}(\tilde{R}_{2,i})}{1+\mathrm{Clamp}_{{1}/{3}}(\tilde{R}_{1,i})}$. 
\end{enumerate}
In the above procedure, Step 2 is the only step that possibly evokes complexity roughly $\mathcal{O}(n^M)$, so we begin by estimating its total complexity and error tolerance.

\textbf{Error Tolerance of Step 2}: note the fact that 
$$H_{M'}(u):v^{\otimes {M'}}=H{e_{M'}}(\langle u,\frac{v}{\|v\|}\rangle)\cdot \|v\|^{M'},He_{k}(\lambda)=\sum_{m=0}^{\lfloor k/2\rfloor}(-1)^m\frac{k!}{2^m m!(k-2m)!}\lambda^{k-2m}$$
we know that $\forall v \in \mathbb{R}^n$, $H_{M'}(u):v^{\otimes {M'}}$ admits the expression
$$\sum_{j=0}^{M'}\sum_{S_{n,j}}C_{M',v,(a_1,...,a_j)}\prod_{k=1}^jp_{a_k}(u),$$
where $C_{M',v,(a_1,...,a_j)}=0$ when $M'-j$ is odd and 
$$C_{M',v,a_1,...,a_{j}}=(-1)^{(M'-j)/2}\frac{M'!\left\|v\right\|^{M'}}{2^{(M'-j)/2} j!((M'-j)/2)!}\prod_{k=1}^jp_{a_k}(\frac{v}{\left\|v\right\|})$$
when $M'-j$ is even. As a result,
$$\sum_{S_{n,j}}|C_{M',v,a_1,...,a_j}|=\frac{M'!\left\|v\right\|^{M'}}{2^{(M'-j)/2} j!\big((M'-j)/2\big)!}\cdot\Big(\sum_{l=1}^n\left|p_l\big(\frac{v}{\left\|v\right\|}\big)\right|\Big)^j$$
and is hence upper bounded by
$$\frac{M'!\left\|v\right\|^{M'}}{2^{(M'-j)/2} j!\big((M'-j)/2\big)!}\cdot n^{j/2},$$
noting that $\left\|v\right\|_1\leq\sqrt{n}\left\|v\right\|_2$ holds for all $v\in\mathbb{R}^n$. 
We thus conclude that, the terms
$$H_{M'}(\frac{x-y_i}{\sqrt{t}}):\int_{B(y_i,r)}(y-y_i)^{\otimes M'}dp_i(y); H_{M'}(\frac{x-y_i}{\sqrt{t}}):\int_{B(y_i,r)}(y-y_i)^{\otimes M'+1}dp_i(y)$$
equal
$$\sum_{j=0}^{M'}\sum _{S_{n,j}}\Big(\int_{v\in B(0,r)}C_{M',v,(a_1,...,a_j)}dp_i(v+y_i)\Big)\prod_{k=1}^j p_{a_k}(\frac{x-y_i}{\sqrt{t}})$$
and 
$$\sum_{j=0}^{M'}\sum _{S_{n,j}}\Big(\int_{v\in B(0,r)}(C_{M',v,(a_1,...,a_j)}\cdot v)dp_i(v+y_i)\Big)\prod_{k=1}^j p_{a_k}(\frac{x-y_i}{\sqrt{t}}),$$
respectively, where  
$$\sum_{S_{n,j}}\Big|\int_{v\in B(0,r)}C_{M',v,(a_1,...,a_j)}dp_i(v+y_i)\Big|; \sum_{S_{n,j}}\Big\|\int_{v\in B(0,r)}(C_{M',v,(a_1,...,a_j)}\cdot v)dp_i(v+y_i)\Big\|$$
also equals zero when $M'-j$ is odd and is upper bounded by
$$\frac{M'!r^{M'}}{2^{(M'-j)/2} j!((M'-j)/2)!}\cdot n^{j/2};\frac{M'!r^{M'+1}}{2^{(M'-j)/2} j!((M'-j)/2)!}\cdot n^{j/2}$$
when $M'-j$ is even. Thus, if we were able to calculate, for each $j\leq M$, every element in the set $\{\prod_{k=1}^jp_{\alpha_k}(\frac{x-y_i}{\sqrt{t}})|(\alpha_1,...,\alpha_j)\in S_{n,j}\}$ each with error no more than (note that $r\leq \min\{\sqrt{t_0},\frac{1}{25}\}$):
$$\min_{j+2l\leq M}\left\{\frac{2^lj!l!}{M\cdot (j+2l)\cdot r^{j+2l}}\cdot n^{-j/2}t_0^{(j+2l)/2} \delta\right\}\geq\frac{j!}{M^2\cdot r^j}\cdot n^{-j/2}t_0^{j/2}\delta\geq\frac{j!}{M^2}\cdot n^{-j/2}\delta,$$
then for all $M'\leq M$ and $j\leq M'$, 
$$\frac{1}{M'!}\sum _{S_{n,j}}\Big(\int_{v\in B(0,r)}C_{M',v,(a_1,...,a_j)}dp_i(v+y_i)\Big)\prod_{k=1}^j p_{a_k}(\frac{x-y_i}{\sqrt{t}})$$
can be calculated with error $\leq \frac{t_0^{M'/2}\delta}{MM'}$, hence
$$\frac{1}{M'!}H_{M'}(\frac{x-y_i}{\sqrt{t}}):\int_{B(y_i,r)}(y-y_i)^{\otimes {M'}}dp_i(y); \frac{1}{M'!}H_{M'-1}(\frac{x-y_i}{\sqrt{t}}): \int_{B(y_i,r)}(y-y_i)^{\otimes ({M'}+1)}dp_i(y)$$
can be calculated with error no more than $\frac{t_0^{M'/2}\delta}{M}$ and $\frac{t_0^{M'/2}\delta}{M}r$, respectively.

\textbf{Computational Complexity}: Note the aforementioned facts that $xy=(\frac{x+y}{2})^2-(\frac{x-y}{2})^2$ and that approximating $g(x)=x^2$ on $[0,U]$ with error $\delta$ can be done with a network of width $\tilde{C}(\log(\frac{U}{\delta}))$ and depth $\tilde{C}(\log(\frac{U}{\delta}))$ for some constant $\tilde{C}$ \citep{squarerelu}, values of elements in the set 
$$Q_{n,j}:=\{\prod_{k=1}^jp_{\alpha_k}(\frac{x-y_i}{\sqrt{t}})|(\alpha_1,...,\alpha_j)\in S_{n,j}\}$$
can be inductively calculated with error $\leq\delta_1$ by first calculating the values of all elements in $Q_{n,j-1}$ with error $\frac{\sqrt{t_0}\delta_1}{2(R+J\sqrt{T})}$, then multiplying $p_{a_j}(\frac{x-y_i}{\sqrt{t}})$ with the already calculated $\prod_{k=1}^{j-1}p_{a_k}(\frac{x-y_i}{\sqrt{t}})$. In particular, the $n$ elements in the set $S_{n,1}$ are calculated by directly multiplying the $n$ coordinates of $x-y_i$ with $t^{-1/2}$ (computed once alongside the powers $\{t^{-M'/2}|1\leq M'\leq M\}$).

In the process mentioned above, monomials of different order are calculated sequentially (i.e., the total network width equals that of the widest layer, while the total network depth equals the summed depth of all layers); and lower order monomials must be preserved rather than discarded after calculation (Preserving an element down a ReLU network takes two additional neurons per layer, one storing its positive part, the other storing its negative part). Therefore, as the set $\{\prod _{k=1}^ja_k|(a_1,...,a_j)\in S_{n,j}\}$ contains $\binom{n+j-1}{j}$ distinct values, mathematical induction proves that this can be done for all $j\leq M$ using a single neural network of width $W_{M,mul}(\delta_1)$ and depth $D_{M,mul}(\delta_1)$, such that
$$W_{k+1,mul}(\delta_1)=\max\Big\{W_{k,mul}(\frac{\sqrt{t_0}\delta_1}{2(R+J\sqrt{T})}),\binom{n+k}{k+1}\tilde{C}\log \big(\frac{1}{\delta_1} (\max\{\frac{\sqrt{t_0}}{R+J\sqrt{T}},1\})^k\big)+2\sum_{l< k}\binom{n+l}{l+1}\Big\}$$
$$D_{k+1,mul}(\delta_1)=D_{k,mul}(\frac{\sqrt{t_0}\delta_1}{2(R+J\sqrt{T})})+\tilde{C}\log \big(\frac{1}{\delta_1} (\max\{\frac{\sqrt{t_0}}{R+J\sqrt{T}},1\})^k\big)$$
$$D_{1,mul}(\delta_1)=\tilde{C}\cdot \log(\frac{\max\{R+J\sqrt{T},t_0^{-1/2}\}}{\delta_1});W_{1,mul}(\delta_1)=nD_{1,mul}(\delta_1)$$
thus, noting that 
$$\binom{n+k-1}{k}\geq\binom{n+k-2}{k-1}\cdot \sqrt{n}, \forall 1\leq k\leq M\leq \sqrt{n},$$
we can inductively prove that under our $\sqrt{n}>M$ criterion, $W_M$ can be upper bounded by (the 2 $\tilde{C}$ relaxation is intentionally loose)
$$ 2\tilde{C}\frac{(n+M)^M}{M!}\big(M\log \small(\frac{R+J\sqrt{T}}{\sqrt{t_0}}\small)+\log \frac{1}{\delta_1}\big)$$
while $D_M$ can be upper bounded by
$$2\tilde{C}(M\log \frac{1}{\delta_1}+M^2\log \frac{2(R+J\sqrt{T})}{\sqrt{t_0}}).$$
Repeating the procedure for $P$ kernels, therefore, require a network of the same depth and $P$ times the width.

With this in mind, we list out the exact error tolerance of steps 1 to 6. As steps 3 and 4 only involve one linear layer with exact weights and introduce no additional error, to approximate $\{\tilde{q}_{i,t}(x)\}_{i=1}^P$ each with $l^2$ error $\leq\frac{t_0\epsilon}{12}$ and $\{\log\tilde{p}_{i,t}(x)\}_{i=1}^P$ each with error $\leq\frac{t_0\epsilon}{3(R+3\sqrt{n}r)}$, it suffices to:
\begin{itemize}
\item Calculate $\{t^{-M'/2}\}_{M'=1}^M$  (results can be calculated once and used for all kernels, hence cost independent of $P$) each with error no greater than
$$\left\{\frac{M'!t_0\epsilon}{36M(R+3\sqrt{n}r)}\cdot \frac{1}{\sqrt{M'!}}\left(\frac{r(R+J\sqrt{T})}{\sqrt{t_0}}\right)^{-M'}\right\}_{M'=1}^M$$
an error tolerance that can be proven to suffice, noting that
$$H_{M'}(\frac{x-y_i}{\sqrt{t}}):\int_{B(y_i,r)}(y-y_i)^{\otimes M'}dp_i(y)\leq\sqrt{M'!}\cdot\left( \frac{r(R+J\sqrt{T})}{\sqrt{t_0}}\right)^{M'}$$
\item Calculate $\frac{\left\|x-y_i\right\|^2}{t}$ each with error no greater than
$$\frac{t_0\epsilon}{36M(R+3\sqrt{n}r)}$$
\item Calculate, for all $M'\leq M$, all elements in the set
$$\left\{\prod_{j=1}^{M'}p_{a_j}(\frac{x-y_i}{\sqrt{t}})|(a_1,...,a_{M'})\in S_{n,M'}\right\}$$
each with error no greater than 
$$\frac{M'!}{M^2}\cdot n^{-M'/2}\cdot \frac{t_0\epsilon}{9M(R+3\sqrt{n}r)}$$
\item Calculate 
$$\frac{t^{-M'/2}}{M'!}H_{M'}(\frac{x-y_i}{\sqrt{t}}):\int_{B(y_i,r)}(y-y_i)^{\otimes {M'}}dp_i(y)$$
and
$$\frac{t^{-M'/2}}{M'!}H_{M'-1}(\frac{x-y_i}{\sqrt{t}}): \int_{B(y_i,r)}(y-y_i)^{\otimes ({M'}+1)}dp_i(y)$$
for all $M'\leq M$ by multiplying the two already-calculated components, each with error no greater than 
$$\frac{t_0\epsilon}{36M(R+3\sqrt{n}r)}$$
\item Calculate $\log (1+\mathrm{Clamp}_{\frac{1}{3}}(\tilde{R}_{1,i}))$ and $\frac{\mathrm{Clamp}_{2r}(\tilde{R}_{2,i})}{1+\mathrm{Clamp}_{{1}/{3}}(\tilde{R}_{1,i})}$, each with error no greater than 
$$\frac{t_0\epsilon}{20(R+3\sqrt{n}r)}$$
\end{itemize}
Re-using previous methods of product calculation, we know that the first three items above can be done for all $P$ kernels using a ReLU network of total depth $D$ and width $W$ of order at most (ignoring lower order terms and factors that can be upper bounded by constants, which include the $3\sqrt{n}r$ term that only appear alongside, and is upper bounded by, $R$):
$$D\leq\mathcal{O}\left(M\log \frac{M}{\epsilon}+M^2\log n\omega\right); W\leq \mathcal{O}\left(P\frac{(n+M)^M}{(M-1)!}\log\frac{n\omega M}{\epsilon }\right),\omega=\frac{(R+J\sqrt{T})}{t_0}$$
while Theorem 3.7 of \citep{exponentialrelu} show that the calculation of $\log$ and inverse on $[\frac{2}{3},\frac{4}{3}]$, needed for the final item requires a network complexity of order
$$D\leq\mathcal{O}\left(\log \frac{\omega}{\epsilon}\log\log \frac{\omega}{\epsilon}\right); W\leq \mathcal{O}\left(P\log^2\frac{\omega}{\epsilon}\right)$$
combining this with the complexity of Softmax calculation, which requires a network complexity of order 
$$D\leq\mathcal{O}\left(\log \frac{\omega P}{\epsilon}\log\log \frac{\omega P}{\epsilon}\right); W\leq \mathcal{O}\left(P\log^2\frac{\omega P}{\epsilon}\right)$$
we acquire the overall complexity bound
$$D\leq\mathcal{O}\left(M\log \frac{M}{\epsilon}+M^2\log n\omega+\log \frac{\omega P}{\epsilon}\log \log \frac{\omega P}{\epsilon}\right);$$ $$W\leq \mathcal{O}\left(P\left(\frac{(n+M)^M}{(M-1)!}\log\frac{n\omega M}{\epsilon }+\log^2\frac{\omega P}{\epsilon}\right)\right)$$
Finally, noting the fact that for any arbitrary points $\{z_j\}_{j=1}^P\subseteq \mathrm{supp}(p_0)$, scalars $\{V_{1,j}\}_{j=1}^P$, vectors $\{V_{2,j}\}_{j=1}^P$, and positive weight inputs $\{w_j\}_{j=1}^P$ summing up to 1, we have
$$\left\|z_j-y_0+\frac{\mathrm{Clamp}_{2r}(V_{2,j})}{1+\mathrm{Clamp}_{1/3}(V_{1,j})}\right\|\leq R+3\sqrt{n}r$$
and
$$\sum_{j=1}^Pw_j(z_j+\frac{\mathrm{Clamp}_{2r}(V_{2,j})}{1+\mathrm{Clamp}_{1/3}(V_{1,j})})\in \mathrm{Conv}(\mathrm{supp}(p_0))+B(0,3\sqrt{n}r),$$
the network's output always remains within $\mathrm{Conv}(p_0)+B(0,3\sqrt{n}r+t_0\epsilon),$ as our approximated Softmax weights are no smaller than zero and sum up to the range $[1-\frac{t_0\epsilon}{20(R+3\sqrt{n}r)},1+\frac{t_0\epsilon}{20(R+3\sqrt{n}r)}]$ for any possible inputs $(x_1,...,x_P)\in\mathbb{R}^P$.
\section{Proof of $C_\mathcal{M}$'s existence}
\begin{lemma}
Let $\mathcal{M}$ be a compact, $C^2$ manifold with positive reach $1/k_1$. Then, there exists a constant $C_\mathcal{M}$ satisfying the following condition: $\forall r$, there exists a covering $\{B(x_i,r)\}$ of $\mathcal{M}$ such that $\forall R, \forall y\in\mathcal{M}$, $B(y,R)$ contains the centers of at most $C_\mathcal{M}(\frac{R}{r}+1)^d$ balls in $\{B(x_i,r)\}$.   
\end{lemma}
\begin{proof}
Without loss of generality, we assume $r\leq\frac{1}{6k_1}$ (the compact manifold $\mathcal{M}$ admits a covering $\{B(x_i,\frac{1}{6k_1})\}_{i=1}^Q$, setting $C_\mathcal{M}=Q$ clearly suffices for all $r>\frac{1}{6k_1}$). 

For each point $y\in\mathcal{M}$, we define the function $f(y,R), R\geq\frac{1}{3k_1}$ to be the smallest number of closed balls centered on $\mathcal{M}$ with radius $\frac{1}{6k_1}$ that is required to cover $B(y_i,R)\cap\mathcal{M}$. Further, we define 
$$g(y,R)=\left(\frac{1}{3k_1R}\right)^d{f(y,R)}$$
As $\mathcal{M}$ is compact (hence can be covered by finite balls of radius $\frac{1}{6k_1}$), $\{g(y,R)|y\in\mathcal{M},R\geq\frac{1}{3k_1}\}$ is obviously upper bounded. Thus, we take $$D_{\mathcal{M}}=\{\sup\{g(y,R)|y\in\mathcal{M},R\geq\frac{1}{3k_1}\}\}.$$

Now, suppose that for each $r\leq\frac{1}{6k_1}$, we can construct a covering $\{B(x_i,r)\}$ of $\mathcal{M}$ such that, for each $y\in\mathcal{M}$ and each $R\leq\frac{1}{3k_1}$, $B(y,R)$ contains the centers of at most $\tilde{C}_d(\frac{R}{r}+1)^d$ balls in $\{B(x_i,r)\}$. If this is so, then for each $R>\frac{1}{3k_1}$ and each $y\in \mathcal{M}$, $B(y,R)$ can be covered by $f(y,R)$ balls of radius $\frac{1}{6k_1}$, every one of which would contain the centers of at most $\tilde{C}_d(\frac{1}{6k_1r}+1)^d$ balls in $\{B(x_i,r)\}$, summing up to a total of $\tilde{C}_d(\frac{1}{6k_1r}+1)^df(y,R)$, and hence
$$\tilde{C}_d\left(\frac{\frac{1}{6k_1r}+1}{\frac{R}{r}+1}\right)^df(y,R)\cdot (\frac{R}{r}+1)^d\leq\tilde{C}_d\cdot\left(\frac{1}{3k_1R}\right)^df(y,R)\cdot(\frac{R}{r}+1)^d\leq\tilde{C}_d{D}_\mathcal{M}\cdot(\frac{R}{r}+1)^d$$
It only remains for us to compute $\tilde{C}_d$. For each $r\leq\frac{1}{6k_1}$, we begin by constructing a maximal set $\{x_i\}_{i=1}^W$ such that the pairwise distance between points is at least $r$ (i.e. no bigger set containing $\{x_i\}_{i=1}^W$ would have this property). The closed balls $\{B(x_i,r)\}_{i=1}^W$ must form a covering of $\mathcal{M}$, or else we would be able to insert another point into $\{x_i\}_{i=1}^W$ without violating the $r$-separation property, which contradicts the maximality of $\{x_i\}_{i=1}^W$.

For any $y\in\mathcal{M}$, we assume, under a trivial change of coordinates, that $y=0$ and $T_y\mathcal{M}=\mathbb{R}^d\times \{0\}$. Then, there exists a Lipschitz continuous function $f$ with Lipschitz constant $L=\frac{1}{2}$ \citep{reachproperty} such that, for all $R\leq\frac{1}{3k_1}$, 
$$\mathcal{M}\cap B(y,R)\subseteq\{(x,f(x))|x\in\mathbb{R}^d,\|x\|\leq R\}.$$
As a result, for each point $(z,f(z))$ in $\mathcal{M}\cap B(y,R)$, the ball $B((z,f(z)),r)$ is guaranteed to cover the set 
$$\{(x,f(x))|\|x\|\leq R; \|z-x\|\leq\frac{3}{4}r\},$$
meaning that for each center pair $(x_1,f(x_1))$ and $(x_2,f(x_2))$ contained by $\mathcal{M}\cap B(y,R)$, the fact that $\|x_1-x_2\|\geq r$ (the separation property) must imply $\|x_1-x_2\|\geq\frac{3}{4}r.$

Finally, in the Euclidean space $\mathbb{R}^d$, a subset of $B(0,R)$ with pairwise distance between points at least $\frac{3}{4}r$ must have cardinality upper bounded by $\tilde{C}_d(\frac{R}{r}+1)^d$ for some $\tilde{C}_d$, as a cubic grid of edge length $\frac{3r}{4\sqrt{d}}$ may contain at most one such point and $B(0,R)$ can be covered by at most $$(\frac{8\sqrt{d}R}{3r}+1)^d\leq(\frac{8\sqrt{d}}{3})^d(\frac{R}{r}+1)^d$$
such grids, finishing the proof.
\end{proof}
\section{Design of Manifold for Experiments}
Let $T_{m,q,K,\omega}$ denote a tree with depth $K+1$ where each parent node has $m$ child nodes. The number 0 is assigned to the root node, and a random integer chosen uniformly from $\{0,1,2,...,q-1\}$ is assigned to each leaf node, with $\omega$ recording the set of random integers chosen for this purpose.

Let $L_{m,q,K,\omega}$ denote the set of leaf nodes inside the tree $T_{m,q,K,\omega}$. For each $l\in L$, let the sequence $s_l=\{0a_{1l}a_{2l}...a_{Kl}\}$ denote the line of ancestor nodes tracing downwards to $l$. Under such a definition, a one dimensional distribution $p_{T_{m,q,k,w_1}}$ is given by
$$p_{T_{m,q,k,w_1}}=\frac{1}{m^K}\sum_{l\in L_{m,q,K,\omega}}U\left(\left[\sum_{j=1}^K\frac{a_{jl}}{q^j},\sum_{j=1}^K\frac{a_{jl}}{q^j}+\frac{1}{q^K}\right]\right).$$
As $k\to\infty$, the support of $p_{T_{m,q,k,w_1}}$ converges to a set of upper Minkowski dimension $\frac{\log m}{\log q}$. Thus, for finite $K$ such that $q^{-K}<<\sqrt{t_0}$, $\mathrm{supp}(p_{T_{m,q,k,w_1}})$ can be viewed as a fractal of upper Minkowski dimension $\frac{\log m}{\log q}$ for the purpose of score approximation on $[t_0,T]$.

For our experiment, we thus sample three random seeds $w_0, w_1, w_2$, and construct the random variables $x\sim \tilde{p}_x=p_{T_{4,5,k,w_0}}$, $y_1\sim p_{T_{2,5,k,w_1}}$, $y_2\sim p_{T_{2,5,k,w_2}}$, and $y=(y_1,y_2)\sim \tilde{p}_y$, setting $k=6$ with $5^{-6}=1/15625$. We twist the support of $\tilde{p}_x$ and $\tilde{p}_y$ to become manifolds in $\mathbb{R}^{32}$ by defining
$$X=(x,\frac{\sin({\pi x})}{20},\frac{1-\cos(\pi x)}{20})\cdot V_1; Y=(y_1,y_2,\frac{\sin({\pi y_1})}{20},\frac{\sin({\pi y_2})}{20})\cdot V_2,$$
where $V_1\in \mathbb{R}^{3\times32}; V_2\in\mathbb{R}^{4\times 32}$ both has orthonormal rows. As such, the underlying distribution $p_X$ of $X$ is contained in a subspace of dimension 3; a submanifold of dimension 1; and has a support of upper Minkowski dimension $\frac{\log 4}{\log5}$; while the underlying distribution $p_Y$ of $Y$ is contained in a subspace of dimension $4$, a submanifold of dimension 2, and has a support of upper Minkowski dimension $\frac{\log 4}{\log 5}$. Thus, comparing the scaling behavior upon approximating $\nabla\log p_{X,t}(x)$ to the scaling behavior upon approximating $\nabla\log p_{Y,t}(x)$ creates a decent test case as to which of the three dimensionalities dominate network complexity.
\section{Experiment Protocol and Raw Experimental Data}

All code, every result file and every training curve are provided in the supplementary
material; this appendix records the protocol in enough detail to re-run any single
number in Figure 2.

\subsection{Data and Exact Score}
The two test distributions $p_X$ and $p_Y$ are the ones constructed in
Appendix G with tree depth $K=6$. For $p_X$ we place two quadrature
points in each of the $4^6$ leaf intervals, and for $p_Y$ one point at the centre of
each of the $2^6\times2^6$ leaf squares, so that each distribution is represented by
$N=16{,}384$ atoms $\{a_i\}$ before the embedding into $\mathbb{R}^n$ ($n=32$ in the
main experiments, $n=64$ in the ablation); $p_0$ is \emph{defined} as the uniform
measure on these atoms. The finest structure of the support is therefore at scale
$5^{-6}\approx 6.4\times10^{-5}$, well below $\sqrt{t_0}\ge 0.0125$ for every $t_0$
used, so on $[t_0,T]$ the support behaves as a fractal of upper Minkowski dimension
$\log 4/\log 5$. Because $p_0$ is a finite mixture, the noised density
$p_t=p_0*\varphi_{t}$ is a Gaussian mixture and its score is available in closed form,
\[
\nabla\log p_t(x)=\frac{\mathbb{E}[x_0\mid x_t=x]-x}{t},\qquad
\mathbb{E}[x_0\mid x_t=x]=\sum_i \operatorname{softmax}_i\!\Big(-\tfrac{\|x-a_i\|^2}{2t}\Big)\,a_i ,
\]
which we evaluate in float64. Every error reported in the paper is measured against
this exact score; no approximation enters the metric.

\subsection{Network}
$s_\theta(x,t)$ is a fully connected ReLU network with $D=8$ hidden layers of equal
width $W$ (the swept variable), input $[\,x,\ u,\ \sin\pi u,\ \cos\pi u,\ \sin 2\pi u,\
\sqrt{t}\,]$ with $u=\log(t/t_0)/\log(T/t_0)\in[0,1]$, and an $n$-dimensional output.
The output is the estimate of $\mathbb{E}[x_0\mid x_t]$ and the score estimate is
$\nabla\log p_{t,appr}(x)=(s_\theta(x,t)-x)/t$, the data-prediction formulation~(\ref{dataperdiction}).
To maintain consistency with theorem statements, no skip connections, normalisation layers or input/output preconditioning are used.
All arithmetic is IEEE float32 with TF32 tensor-core matrix multiplication disabled
(the metric requires roughly $10^{-4}$ absolute accuracy of $s_\theta$ at the smallest
$t_0$, which TF32 rounding would not support).

\subsection{Training}
\paragraph{Objective.} Since the posterior mean is available exactly, the network is
fitted by regression on the oracle rather than by denoising score matching:
\[
\mathcal{L}(\theta)=\mathbb{E}_{t}\,\mathbb{E}_{x_t}\Big[\Big(\tfrac{t_{\mathrm{ref}}}{t}\Big)^{2}
\big\|s_\theta(x_t,t)-\mathbb{E}[x_0\mid x_t]\big\|^2\Big],\qquad t_{\mathrm{ref}}=\sqrt{t_0T},
\]
which has the same minimiser as the denoising loss but no target noise, and whose
weight makes it equal (up to the constant $t_{\mathrm{ref}}^2$) to the score mean squared
error(\ref{MSE}) that the theorems bound. Training points are drawn as
$x_t=a+\sqrt{t}\,z$ with $a$ uniform over the atoms and $z\sim\mathcal{N}(0,I_n)$; $t$ is
log-uniform on $[t_0,T]$ for half of every batch and log-uniform on $[t_0,10t_0]$ for the
other half, since the smallest noise levels dominate the max-over-$t$ metric. Because
the oracle costs about 5\,ms per batch of 8192, targets are precomputed into a pool of
$2^{24}\approx1.68\times10^{7}$ triples $(x_t,\mathbb{E}[x_0\mid x_t],t)$ kept on the
GPU, from which minibatches are sampled with replacement; the whole pool is regenerated
from fresh random draws every $10^5$ steps (about 10\,s), so each sample is reused
roughly 50 times and there is no fixed training set. A pilot comparison at $W=128$
(supplementary material, \texttt{logs/control\_dsm\_vs\_exact\_pilot}) reached
$\epsilon=0.53$ after $3\times10^5$ steps with the oracle target versus $1.35$ with the
standard noisy denoising target under an identical schedule.

\paragraph{Optimisation.} Adam ($\beta=(0.9,0.999)$, $\varepsilon=10^{-8}$), batch size
8192, base learning rate $\eta_0=3\times10^{-4}\cdot\min\{1,\sqrt{64/W}\}$, linear warm-up
over the first 1000 steps, then cosine decay over a budget of $S_{\max}=10^{7}$ steps
down to $10^{-3}\eta_0$. In addition the learning rate is halved whenever the evaluation
metric plateaus: every 2000 steps the EMA copy of the weights (decay 0.9995) is
evaluated; once a window of $\max\{10^5,\,0.1\cdot\text{step}\}$ steps has elapsed, the
mean evaluation over the recent half of the window is compared with the mean over the
earlier half, and if it is not at least 1\% lower the learning rate is multiplied by
$1/2$ and a new window starts. A run stops early (``converged'') when such a plateau
is detected with the learning rate already below $10^{-3}\eta_0$, and otherwise at
$S_{\max}$ (``max steps''). Narrow networks therefore finish in $1.4$--$9\times10^{6}$
steps while wide ones use most of the budget; Tables 3--5
list the number of steps and halvings for every run. Seeds are fixed (network
initialisation and sample stream: 100; evaluation set: 1234; fractal trees: 0, 1, 2)
and one seed is used per configuration; re-running a configuration on the same
hardware reproduces its number to all printed digits.

\subsection{Evaluation}
For each configuration a fixed evaluation set is drawn once: at each of eight
geometrically spaced times $t_1=t_0<\dots<t_8=T=1$, $65{,}536$ samples
$x_t=a+\sqrt{t}z$ together with the exact score. The reported quantity is
\[
\epsilon=\max_{k\le 8}\ \sqrt{\tfrac16\,\mathbb{E}_{x\sim p_{t_k}}\big\|\nabla\log p_{t_k}(x)-\nabla\log p_{t_k,appr}(x)\big\|^2}\,,
\]
i.e.\ the normalisation and the ``for all $t\in[t_0,T]$'' quantifier of
Theorem~\ref{thm:main}; a configuration achieves error $\epsilon$ in the sense of the
theorem exactly when this quantity is at most $\epsilon$. The value plotted in
Figure 2 is that of the final raw (non-averaged) weights. The EMA
copy is uniformly worse at the end of training (by a factor 1.2--3, Tables 3--5); this gap is an artefact of accumulating the
average in float32 (the update $\theta_{\rm ema}\leftarrow 0.9995\,\theta_{\rm ema}+0.0005\,\theta$
cannot track $\theta$ closer than about $4\times10^{-5}$ relative), and disappears in
the float64 control run below, where the two agree to $0.4\%$. With $65{,}536$
evaluation samples per time the statistical uncertainty of $\epsilon$ is below 1\%,
far smaller than the spacing of the width ladder.

\subsection{Controls and Caveats}
\begin{itemize}
\item \textbf{Optimisation outliers.} The $W=16$ runs on $p_X$ (both $t_0$) stopped
after $1.4$--$1.5\times10^6$ steps at $\epsilon=4.87$ and $1.57$ respectively: the loss
stalled on a plateau early, the plateau rule halved the learning rate ten times and the
run terminated. They are shown in Figure 2(a) but excluded from
the slope fits; the corresponding $W=16$ run on $p_Y$ ($\epsilon=1.67$) trained normally.
\item \textbf{Error floor at large width.} At $t_0=1.5625\times10^{-4}$ the widths
96, 128 and 192 all reach $\epsilon\approx0.05$--$0.06$ on $p_X$, and at fixed width the
product $\epsilon\cdot t_0$ is nearly constant across all six $t_0$ (Table~\ref{tab:raw-x}),
i.e.\ the absolute error of the $x_0$-prediction saturates at about $2\times10^{-5}$
independently of width and $t_0$. To test whether this is a float32 artefact, the
$W=128$, $t_0=1.5625\times10^{-4}$ run was repeated with the network, the sample pool
and the optimiser entirely in float64 (\texttt{code/control\_fp64\_unit.py}); it
reached $\epsilon=0.0443$ versus $0.0509$ in float32 after the same $10^{7}$ steps, so
machine precision accounts for at most 13\% of the floor. The plateau is therefore an
optimisation/representation limit of the fixed-depth network at the smallest noise
levels rather than a numerical one, and the slope fits in
Figure 2(a) use $20\le W\le192$ as stated.
\item \textbf{Ambient dimension.} Table~\ref{tab:raw-n64} repeats three widths of both
distributions with $n=64$ instead of $32$; the embedding uses a fresh random orthonormal
frame of the larger dimension and everything else is unchanged.
\item \textbf{Compute.} One configuration takes $0.7$--$5.3$ GPU-hours (about 1.8\,ms per step at $W\le128$); the complete set of runs reported here used about 210 GPU-hours.
\end{itemize}
\section{Raw Data}
\begin{table}[h]
\centering
\small
\caption{Least-squares slopes of $\log\epsilon$ against $\log W$ (widths $20\le W\le 192$, $W=16$ excluded as optimisation outliers) and against $\log t_0$ (all six $t_0$).}
\label{tab:slopes}
\begin{tabular}{llr}
\toprule
series & fit & slope \\
\midrule
$p_X$, $t_0=0.00015625$ & $\log\epsilon$ vs $\log W$ & -1.165 \\
$p_Y$, $t_0=0.00015625$ & $\log\epsilon$ vs $\log W$ & -1.150 \\
$p_X$, $t_0=0.000625$ & $\log\epsilon$ vs $\log W$ & -0.982 \\
$p_X$, $W=64$ & $\log\epsilon$ vs $\log t_0$ & -0.928 \\
$p_Y$, $W=64$ & $\log\epsilon$ vs $\log t_0$ & -0.839 \\
$p_X$, $W=128$ & $\log\epsilon$ vs $\log t_0$ & -0.925 \\
\bottomrule
\end{tabular}
\end{table}
\begin{table}[h]
\centering
\small
\caption{Raw results, distribution $p_Y$ ($d'=2$), $n=32$.}
\label{tab:raw-y}
\begin{tabular}{llrrrrrrlr}
\toprule
dist & $n$ & $t_0$ & $W$ & $\epsilon$ (raw) & $\epsilon$ (EMA) & steps & lr drops & stop & GPU-h \\
\midrule
y & 32 & 0.005 & 64 & 0.0098 & 0.0122 & 6.05M & 9 & converged & 2.9 \\
y & 32 & 0.0025 & 64 & 0.0185 & 0.0238 & 4.41M & 10 & converged & 2.1 \\
y & 32 & 0.00125 & 64 & 0.0283 & 0.0412 & 7.47M & 8 & converged & 3.7 \\
y & 32 & 0.000625 & 64 & 0.0393 & 0.0536 & 6.05M & 9 & converged & 3.0 \\
y & 32 & 0.0003125 & 64 & 0.1181 & 0.1359 & 4.41M & 10 & converged & 2.2 \\
y & 32 & 0.00015625 & 16 & 1.6717 & 1.7396 & 6.05M & 9 & converged & 3.1 \\
y & 32 & 0.00015625 & 20 & 0.9738 & 1.0807 & 10.00M & 7 & max steps & 4.8 \\
y & 32 & 0.00015625 & 24 & 0.6888 & 0.7743 & 7.47M & 8 & converged & 3.6 \\
y & 32 & 0.00015625 & 28 & 0.5431 & 0.6392 & 10.00M & 4 & max steps & 4.9 \\
y & 32 & 0.00015625 & 32 & 0.4022 & 0.4662 & 10.00M & 2 & max steps & 4.9 \\
y & 32 & 0.00015625 & 40 & 0.3196 & 0.4242 & 9.23M & 6 & converged & 4.5 \\
y & 32 & 0.00015625 & 48 & 0.2080 & 0.3150 & 9.23M & 4 & converged & 4.5 \\
y & 32 & 0.00015625 & 56 & 0.2113 & 0.3106 & 9.23M & 5 & converged & 4.5 \\
y & 32 & 0.00015625 & 64 & 0.1768 & 0.2442 & 8.31M & 8 & converged & 4.0 \\
y & 32 & 0.00015625 & 96 & 0.1097 & 0.1754 & 10.00M & 8 & max steps & 5.0 \\
y & 32 & 0.00015625 & 128 & 0.1316 & 0.1990 & 6.05M & 10 & converged & 3.2 \\
y & 32 & 0.00015625 & 192 & 0.0580 & 0.1708 & 9.23M & 7 & converged & 5.3 \\
\bottomrule
\end{tabular}
\end{table}

\begin{table}[h]
\centering
\small
\caption{Raw results, distribution $p_X$ ($d'=1$), $n=32$. $\epsilon=\sqrt{\mathrm{MSE}_{\max}/6}$ of the final raw model (used in all figures) and of its EMA copy.}
\label{tab:raw-x}
\begin{tabular}{llrrrrrrlr}
\toprule
dist & $n$ & $t_0$ & $W$ & $\epsilon$ (raw) & $\epsilon$ (EMA) & steps & lr drops & stop & GPU-h \\
\midrule
x & 32 & 0.005 & 64 & 0.0029 & 0.0067 & 8.31M & 7 & converged & 4.2 \\
x & 32 & 0.005 & 128 & 0.0022 & 0.0061 & 6.05M & 9 & converged & 3.1 \\
x & 32 & 0.0025 & 64 & 0.0065 & 0.0115 & 8.31M & 9 & converged & 4.2 \\
x & 32 & 0.0025 & 128 & 0.0038 & 0.0102 & 6.73M & 9 & converged & 3.5 \\
x & 32 & 0.00125 & 64 & 0.0150 & 0.0276 & 6.05M & 10 & converged & 3.0 \\
x & 32 & 0.00125 & 128 & 0.0081 & 0.0212 & 5.45M & 10 & converged & 2.8 \\
x & 32 & 0.000625 & 16 & 1.5686 & 1.5945 & 1.38M & 10 & converged & 0.7 \\
x & 32 & 0.000625 & 20 & 0.1027 & 0.1290 & 9.23M & 5 & converged & 4.8 \\
x & 32 & 0.000625 & 24 & 0.0943 & 0.1187 & 8.31M & 8 & converged & 4.2 \\
x & 32 & 0.000625 & 28 & 0.0777 & 0.1206 & 6.73M & 9 & converged & 3.4 \\
x & 32 & 0.000625 & 32 & 0.0472 & 0.0777 & 9.23M & 7 & converged & 4.7 \\
x & 32 & 0.000625 & 40 & 0.0349 & 0.0592 & 8.31M & 7 & converged & 4.2 \\
x & 32 & 0.000625 & 48 & 0.0405 & 0.0693 & 5.45M & 10 & converged & 2.9 \\
x & 32 & 0.000625 & 56 & 0.0335 & 0.0610 & 5.45M & 9 & converged & 2.8 \\
x & 32 & 0.000625 & 64 & 0.0262 & 0.0463 & 4.41M & 10 & converged & 2.3 \\
x & 32 & 0.000625 & 96 & 0.0155 & 0.0405 & 6.05M & 9 & converged & 3.1 \\
x & 32 & 0.000625 & 128 & 0.0203 & 0.0438 & 3.97M & 10 & converged & 2.1 \\
x & 32 & 0.000625 & 192 & 0.0106 & 0.0364 & 7.47M & 9 & converged & 3.7 \\
x & 32 & 0.0003125 & 64 & 0.0450 & 0.1011 & 6.05M & 9 & converged & 3.1 \\
x & 32 & 0.0003125 & 128 & 0.0259 & 0.0622 & 5.45M & 9 & converged & 2.8 \\
x & 32 & 0.00015625 & 16 & 4.8735 & 4.8983 & 1.53M & 10 & converged & 0.7 \\
x & 32 & 0.00015625 & 20 & 1.0260 & 1.2177 & 6.05M & 9 & converged & 3.1 \\
x & 32 & 0.00015625 & 24 & 0.3070 & 0.4009 & 9.23M & 5 & converged & 4.5 \\
x & 32 & 0.00015625 & 28 & 0.2476 & 0.3268 & 9.23M & 6 & converged & 4.8 \\
x & 32 & 0.00015625 & 32 & 0.2565 & 0.4177 & 8.31M & 8 & converged & 4.2 \\
x & 32 & 0.00015625 & 40 & 0.1168 & 0.2069 & 9.23M & 7 & converged & 4.7 \\
x & 32 & 0.00015625 & 48 & 0.1875 & 0.2811 & 6.05M & 9 & converged & 3.1 \\
x & 32 & 0.00015625 & 56 & 0.1464 & 0.2255 & 3.97M & 10 & converged & 2.0 \\
x & 32 & 0.00015625 & 64 & 0.0726 & 0.1411 & 6.73M & 8 & converged & 3.4 \\
x & 32 & 0.00015625 & 96 & 0.0585 & 0.1520 & 6.05M & 9 & converged & 3.2 \\
x & 32 & 0.00015625 & 128 & 0.0509 & 0.1223 & 10.00M & 8 & max steps & 5.1 \\
x & 32 & 0.00015625 & 192 & 0.0522 & 0.1177 & 5.45M & 10 & converged & 3.1 \\
\bottomrule
\end{tabular}
\end{table}

\begin{table}[h]
\centering
\small
\caption{Ambient-dimension ablation, $n=64$, $t_0=1.5625\times10^{-4}$.}
\label{tab:raw-n64}
\begin{tabular}{llrrrrrrlr}
\toprule
dist & $n$ & $t_0$ & $W$ & $\epsilon$ (raw) & $\epsilon$ (EMA) & steps & lr drops & stop & GPU-h \\
\midrule
x & 64 & 0.00015625 & 32 & 0.1632 & 0.2897 & 8.31M & 8 & converged & 4.2 \\
x & 64 & 0.00015625 & 64 & 0.0835 & 0.2199 & 3.97M & 10 & converged & 2.0 \\
y & 64 & 0.00015625 & 32 & 0.3965 & 0.4844 & 10.00M & 4 & max steps & 5.1 \\
y & 64 & 0.00015625 & 64 & 0.1452 & 0.2635 & 8.31M & 8 & converged & 4.2 \\
\bottomrule
\end{tabular}
\end{table}

\end{document}